\documentclass[preprint]{elsarticle}

\usepackage[applemac]{inputenc}
\usepackage{multirow}

\graphicspath{{Graphics/}}

\DeclareGraphicsExtensions{.png,.pdf,.jpg}

\usepackage{epsfig}
\usepackage{amsmath}
\usepackage{amssymb}

\usepackage{amssymb}
\usepackage[dvipsnames,usenames]{color}
\usepackage{soul}
\usepackage{amsthm}

\theoremstyle{definition}

\theoremstyle{remark}

\newcommand{\tab}[1][5]{\hspace{#1mm}}

\begin{document}

\begin{frontmatter}

\title{Approximate Robotic Mapping from sonar data by modeling Perceptions with Antonyms\tnoteref{TIN}}

\tnotetext[TIN]{This work has been supported by the Spanish Department of
Science and Innovation (MICINN) under program Juan de la Cierva JCI-2008-3531,
and the European Social Fund.}


\author{Sergio Guadarrama\corref{cor1}}
\ead{sergio.guadarrama@softcomputing.es} \cortext[cor1]{Corresponding author}
\address{Fundamentals of Soft Computing Unit\\European Centre for Soft Computing\\ Mieres, Asturias (Spain)}
\author{Antonio Ruiz-Mayor}
\ead{aruiz@fi.upm.es}
\address{Departamento de Tecnolog\'{\i}a Fot\'{o}nica\\ Universidad Polit\'{e}cnica de Madrid\\
Campus de Montegancedo, Boadilla del Monte (Spain)}

\begin{abstract}

This work, inspired by the idea of ``Computing with Words and Perceptions'' proposed by Zadeh in \cite{Zadeh:2001:From_computing_with_numbers_to_computing_with_words,Zadeh:2006:A_new_frontier_in_computation}, focuses on how to transform measurements into perceptions \cite{Gua:2007:A_Contribution_to_Computing_with_Words_Perceptions}
for the problem of map building by Autonomous Mobile Robots. We propose to
model the perceptions obtained from sonar-sensors as two grid maps: one for
obstacles and another for empty spaces. The rules used to build and integrate
these maps are expressed by linguistic descriptions and modeled by fuzzy
rules. The main difference of this approach from other studies reported in the
literature is that the method presented here is based on the hypothesis that
the concepts ``occupied'' and ``empty'' are antonyms rather than complementary
(as it happens in probabilistic approaches), or independent (as it happens in
the previous fuzzy models).

Controlled experimentation with a real robot in three representative indoor
environments has been performed and the results presented. We offer a
qualitative and quantitative comparison of the estimated maps obtained by the
probabilistic approach, the previous fuzzy method and the new antonyms-based
fuzzy approach. It is shown that the maps obtained with the antonyms-based
approach are better defined, capture better the shape of the walls and of the
empty-spaces, and contain less errors due to rebounds and short-echoes.
Furthermore, in spite of noise and low resolution inherent to the
sonar-sensors used, the maps obtained are accurate and tolerant to
imprecision.
\end{abstract}


\end{frontmatter}

\section{Introduction}

A commonly required capability of Autonomous Mobile Robots (AMR) is map
building. Without a given map, the robot has to navigate, perceive the
environment (exteroception), integrate each actual perception with the
previous ones, and maintain a coherent and sufficiently accurate
representation of the environment. On the one hand, maps can be used as
references for navigation, in particular for robot localization and path
planning. On the other hand, they can be used as themselves for mapping
purposes.

Mapping is an active research area, and there is no final taxonomy of map
types. There are several ways to represent a map, for instance, a classical
differentiation is between metric and topological maps; metric maps are based
on Cartesian reference systems; topological maps emphasize the relations
between environmental elements, typically rooms and corridors. This work
focuses on metric maps, in particular occupation grid maps, in which the
information is represented by bidimensional grids. Habitually, each grid's
cell corresponds to a squared space region parallel to the floor and at the
height of the robot sensors, and it contains the available knowledge about the
cell.

Habitual sensors used in indoor AMRs in order to build maps include: odometric
sensors (that measure the relative position of the robot with respect to
previous ones), and range sensors (that measure the distance to obstacles).
Main range sensors are ultrasonic, or sonar, and radial laser sensors.
Although laser range sensors offer a greater angular resolution, sonar sensors
have reduced cost, are present on almost every robot platform, and require the
computation of a smaller raw data volume.

This work focuses only on sonar exteroception because it suffers from a higher
imprecision that other kinds of exteroceptions
\cite{CaoBorenstein:2002:Experimental_Characterization_of_Polaroid_Ultrasonic_Sensors},
and therefore an accurate mapping is more difficult to achieve, and because
the results could be later extend to other kinds of exteroceptions. An
objective of this work is to study how fuzzy logic can help us manage this
imprecision
\cite{zadeh:2005:Toward_a_generalized_theory_of_uncertainty_GTU_an_outline,diaz2009fuzzy}.

Odometric sensors suffer from some problems
\cite{BorensteinEverett:96:Navigating_Mobile_Robots}; they have a limited
resolution and offer incorrect readings when the wheel slips. Consequently,
imprecision of odometric estimation grows with distance and number of
maneuvers.

Sonar sensing also suffers from several problems
\cite{LeeChung:2009:Effective_Maximum_Likelihood_Grid_Map}; the measure of
``time of flight'' (TOF) has imprecision inherent in the measuring instrument,
it has a poor angular resolution due to the transducers aperture, and the
signal emitted forms an open solid angle which does not permit us to know
exactly in which part of the wavefront the obstacle is located
\cite{Nehmzov:2000:Mobile_Robotics}. Additionally, if the angle of incidence
of the beam in respect to the surface is greater than half of the sensor
aperture, the echo may not return, or may return after being reflected on
other surfaces. This effect is more likely when the surface is planar and
smooth. These kinds of surfaces, such as glass, marble, polished wood,
plastic, etc. are commonly found in indoor environments. In general, it can be
said that a model of the AMR position based on the distances obtained from
ultrasonic sensors is not continuous and not linear.

The specific problem we focus on is how to build an accurate and robust map of
the environment by integrating the actual perception with past perceptions
\cite{durrant2006simultaneous,bailey2006simultaneous,Cohen:2006:Statistical_evaluation_method}.
This problem includes the sub-problem of how to handle sensor noise and the
contradictions that arise during the process.

\begin{quotation}
{\it``A fundamental difference between measurements and perceptions is that,
in general, measurements are crisp numbers whereas perceptions are fuzzy
numbers or, more generally, fuzzy granules, that is, clumps of objects in
which the transition from membership to nonmembership is gradual rather than
abrupt''.} L. Zadeh in
\cite{Zadeh:2001:From_computing_with_numbers_to_computing_with_words}
\end{quotation}

The approach taken in this paper is based on the assumption that a perception
can be represented by means of linguistic descriptions
\cite{Zadeh:2006:A_new_frontier_in_computation,Gua:2007:A_Contribution_to_Computing_with_Words_Perceptions},
which express imprecise constraints and therefore are gradually satisfied
\cite{Zadeh:2008:Is_there_a_need_for_fuzzy_logic,GuaMunozVaucheret:2004:Fuzzy_Prolog:_a_new_approach_using_soft_constraints_propagation},
and on the hypothesis that occupied and empty are antonyms instead of
complementary. In consequence, we propose to build a fuzzy model of obstacles
and another of empty-spaces that verify the properties of antonyms.

The main differences of this work with previous ones are:
\begin{itemize}
 \item It is not assumed that obstacles and empty-spaces are
complementary, and that one is the complement of other, as happens in probabilistic models.
 \item It is not assumed that obstacles and empty-spaces are independent as happens in  previous fuzzy models.
 \item We assume that obstacles and empty-spaces form a pair of antonyms and should be modeled as such.
 \item It is not assumed that observations are independent as happens in
 probabilistic models, in fact some observations are used to correct others.
 \item It is not assumed that the exact position of the robot is know, so
 there is imprecision about it. So the model should tolerate that imprecision and still produce robust maps.
 \item We dealt with rebounds, short echoes, and other noises by defining a set
of fuzzy rules that capture our knowledge about the problem.
 \item The way of building the aggregated maps by means of linguistic quantifiers differs
greatly from previous approaches, and it allowed us to handle the partial
contribution of each sonar reading to the aggregated map.
\end{itemize}

The main contributions of this work are:
\begin{itemize}
 \item A robust model based on antonyms that can properly
handle the imprecision and contradictions that arise in the process of
building navigation maps.
 \item The antonyms-based model allows to discard rebounds and
short-echoes and reduce greatly the contradictions and errors. Also, by
dealing explicitly with contradictions, this model is able to distinguish
between two kinds of unknown cells, the ones that are unknown due to
contradictions and the ones that are unknown because are unexplored. This
allowed to the robot to recognize which zones needed to be navigated with care
and which ones needed to be explored later on.
\end{itemize}

The maps obtained by the antonyms-based method are better defined, capture better the shape of the walls and of the empty-spaces, and contain less errors due to rebounds and short-echoes. The use of approximate maps allowed us to synthesize the accumulated
information from samples in a way that kept the data structure constant while
the accuracy of the representation increased with the number of
samples. The proposed method obtains better maps with higher confidences, and therefore is more robust to noise and to imprecision of sonar-sensors. Based on the qualitative and quantitative comparison performed, we can conclude that the antonyms-based method performs better than the probabilistic method and the previous fuzzy method, obtaining a better recall of obstacles and empty-spaces, a good balance between precision and recall, a higher $TCR$ and a smaller $MAE$ in the three experiments performed.

The rest of the paper is organized as follows: Section \ref{sec:state}
presents related works and highlight their differences with the current work.
Section \ref{sec:foundations} explains the theoretical foundations that
support the contribution. Section \ref{sec:approximate_maps} details the main
contribution, antonyms-based fuzzy maps. Section \ref{sec:experiments}
presents the experiments and results. Finally, conclusions are covered in
Section \ref{sec:conclusion}. Additionally, an appendix containing all the
maps for an easy comparison has been included.

\section{Related Work}
\label{sec:state}

In the last decade some approximations of the AMR mapping and localization
problems have been published
\cite{DudekJenkin:2000:Computational_Principles_of_Mobile_Robotics,BorensteinEverett:96:Navigating_Mobile_Robots,
GutmannSchlegel:96:AMOS,ChowRadIP:2002:Enhancement_of_Probabilistic_Grid-based_Map_for_Mobile_Robot_Applications},
but definitive solutions have not been found. Depending on the kind of sensors
used, different approaches are taken, for example: in the case of sonar
sensors \cite{fazli2006simultaneous,yan2006building}, in the case of laser
rangefinder sensors \cite{nguyen2005comparison,nguyen2007comparison}, in the
case of video sensors \cite{se2005vision,Eustice:2006:Visually_mapping}, or
combinations of them as in \cite{Li:2006:Robot_map_building,chen2009ep}.

The three main approaches to robot map building are: topological
\cite{kortenkamp:1994:Topological_Mapping_for_Mobile_Robots,Thrun:1998:Learning_metric-topological_maps},
feature-based \cite{Leonardetal:1992:Dynamic_map_building_for_AMR} and
grid-based \cite{elfes:1989:Using_Occupancy_Grids_for_Mobile_Robot}. Recently other approaches have appeared using sparse matrices
\cite{Yguel-RSS-07} or graph-based maps \cite{Grisetti-RSS-07}.

In the case of grid-based maps, a typical way of representing the robot
observations is by means of an occupancy grid, in which each cell can have two
states, empty or occupied, and the grid contains the probability of the cell
being occupied, as it was proposed by Elfes in
\cite{Elfes:1987:Sonar_based_real_world_mapping_and_navigation,elfes:1989:Using_Occupancy_Grids_for_Mobile_Robot},
and recently used in
\cite{Thrun:2005:Probabilistic_Robotics,LeeChung:2009:Effective_Maximum_Likelihood_Grid_Map}.
To have a feasible algorithm to build this occupancy grid using a
probabilistic approach it is necessary to assume that the current observation
is independent from the previous ones, and that the probability of being
occupied for each cell is independent from the others, but in most practical
cases these assumptions fail, and the maps based on them contain severe
errors, (see
\cite{RiboPinz:2001:A_comparison_of_three_uncertainty_calculi_for_building_sonar-based_occupancy_grids,zhang2007comparative}
for a comparison). A difference between the work presented here and probabilistic approaches is that our model does not need to assume that observations are independent,
in fact, the opposite is assumed when some observations are used to correct
others (see Section \ref{sec:con_int} for details).

Several sensor models have been developed by different authors
\cite{Thrun:2005:Probabilistic_Robotics}. These models provide different
description levels of factors that produce uncertainty. Probabilistic models
usually use  stochastic techniques
\cite{Elfes:1987:Sonar_based_real_world_mapping_and_navigation,Foxetal:99:Montecarlo_Localization,
GutmannSchlegel:96:AMOS} (see in \cite{Thrun:2005:Probabilistic_Robotics} a
classification of probabilistic sensor models) in order to build an occupancy
grid (where cells contain the probability of being occupied), but in
\cite{Saffiotti:1997:The_uses_of_fuzzy_logic_in_autonomous_robot_navigation,RiboPinz:2001:A_comparison_of_three_uncertainty_calculi_for_building_sonar-based_occupancy_grids}
it was proved that fuzzy models are more robust (have a higher recall of
obstacles) and better suited for managing this imprecision and for building
approximate maps (where cells contain the degree of being an obstacle, or of
being an empty space).

The model presented here is a fuzzy model, because the approximate
maps built contain the degree of being an obstacle, or of being an empty
space. However, the main difference of this model with previous fuzzy
models
\cite{Ruspini:1990:Fuzzy_Logic_in_the_Flakey,Saffiotti:1997:The_uses_of_fuzzy_logic_in_autonomous_robot_navigation,RiboPinz:2001:A_comparison_of_three_uncertainty_calculi_for_building_sonar-based_occupancy_grids,OrioloUliviVendittelli:1997:Fuzzy_maps:_A_new_tool_for_mobile_robot_perception_and_planning,GasosMartin:1997:Mobile_robot_localization_using_fuzzy_maps,AguirreGonzalez:2003:A_Fuzzy_Perceptual_Model_for_Ultra_sound_Sensors_Applied_to_Intelligent_Navigation_of_Mobile_Robots,karaman2005navigation,yan2006building}
is that in our model obstacles and empty-space form a pair of antonyms. And
that impose certain constraints (see Section \ref{sec:antonyms}) over the
obstacle and empty-spaces maps that allows us to handle explicitly
contradictions that arouse in the process, and to warranty that the integrated map
will contain much less contradictions.

Recently some works have focused on the problem of dealing with the
contradictions that arise during the mapping and navigation process, for
example Lee and Chung following a probabilistic approach proposed in
\cite{LeeChung:2009:Effective_Maximum_Likelihood_Grid_Map} to use a conflict
evaluated maximum approximated likelihood (CEMAL) approach to deal with that
contradictions, although they assumed that observations are independent and
that the exact position of the robot is known in advance. As it has been said
before in our model it is not assumed the independency of observations nor
that the exact robot position is know in advance (see Section
\ref{sec:experiments} for details).

\section{Fundamentals}
\label{sec:foundations}

Occupancy grid maps are built using a grid of cells with a certain size. Each
cell is associated with a state and a confidence degree, that in the case of
probability maps
\cite{Elfes:1987:Sonar_based_real_world_mapping_and_navigation} is the
probability mass function, or in the case of fuzzy maps
\cite{OrioloUliviVendittelli:1997:Fuzzy_maps:_A_new_tool_for_mobile_robot_perception_and_planning}
is the possibility degree. Before introducing the foundations of antonyms,
probabilistic maps and fuzzy maps are presented.

\subsection{Probabilistic Maps}

In the probabilistic approach, to be able to tackle grids of a size big enough
(e.g 100x100) it is assumed that cells states are independent and complete.
That is, each cell has a state $s$, occupied or empty $s(C_{ij}) \in \{E,O\}$,
with a certain probability defined by the function
$P:\{E,O\}\rightarrow[0,1]$, that can be represented by a pair of matrices of
size $(n\times m)$, with the probability associated with each state:
\begin{equation}\begin{array}{c}
 Map_{Occup} = \left[\begin{array}{ccc}
  P[s(C_{11})=O] &  ... &  P[s(C_{1m})=O]\\
   \vdots &  P[s(C_{ij})=O] & \vdots\\
    P[s(C_{n1})=O] & ... &   P[s(C_{nm})=O]
  \end{array}\right]\\
  \\
  Map_{Empty} = \left[\begin{array}{ccc}
  P[s(C_{11})=E] &  ... &  P[s(C_{1m})=E]\\
   \vdots &   P[s(C_{ij})=E] & \vdots\\
    P[s(C_{n1})=E] & ... &   P[s(C_{nm})=E]
  \end{array}\right]
   \end{array}
 \end{equation}
 In this case, and since both Maps are complementary
  \begin{equation}P[s(C_{ij}=E)]+P[s(C_{ij})=O] = 1,\end{equation}
    then maintaining one map is enough.

Each cell is associated with a probability mass function, which is estimated
from a sensor model that measures confidence, and an integration process done
by the Bayes-Rule. For example, in
\cite{RiboPinz:2001:A_comparison_of_three_uncertainty_calculi_for_building_sonar-based_occupancy_grids}
Ribo and Pinz proposed a sensor model detailed below\footnote{Small mistakes
presented in the original formula have been corrected}, although other models
of the sensor measurements $p[r|s(C_{ij})=X]$ have been defined (see
\cite{Thrun:2002:Robotic_Mapping_A_Survey} for details).

\begin{align}
& p[r|s(C_{ij})=O] = p_1[r|s(\theta,\rho)=O] + p_2[\theta,\rho|r] \label{eq:sensor_model}\\
p_1[r|s(\theta,\rho)] =& \left\{\begin{array}{ll}
 \left(1-\lambda\right)0.5+\lambda p_E & 0\leq \rho < r- 2\delta r, \\
 (0.5-p_E)\left(1-\lambda\left(\frac{r - \rho-\delta r}{\delta r}\right)^2\right) & r-2\delta r \leq \rho < r- \delta r,\\
 \lambda\left(p_O-0.5\right)\left(1-\left(\frac{r-\rho}{\delta r}\right)^2\right) & r-\delta r \leq \rho < r + \delta r, \\
 0.5 & \rho \geq r + \delta r
 \end{array}\right. \label{eq:sensor_reading}\\
 p_2[\theta,\rho|r] =& \left\{\begin{array}{lrl}
 p_E, & 0\leq & \rho < r- \delta r, \\
 0.5, & r - \delta r \leq & \rho
 \end{array}\right.\label{eq:cell_matching}
\end{align}
Where $r$ is a given range reading, $\rho$ is the distance from the sensor to
$C_{ij}$, $\theta$ is the angular distance between the beam axis and
$C_{ij}$, $2\delta r$ is the width considered proximal to the reading, or
the error in measure, and $\lambda = \Gamma(\rho)\cdot\Delta(\theta)$ is the
confidence in the reading expressed by the confidence in the distance (expressed in
meters) and in the angle (expressed in radians), as follows:
\begin{align}
\Gamma(\rho) &= 1 + \frac{1+tanh(2(\rho - \rho_v))}{2} \\
\Delta(\theta) &= \left\{\begin{array}{ll}
 1 - \left(\frac{\theta}{0.2182}\right)^2, & 0 \leq |\theta| \leq 0.2182 \\
 0, & \mbox{ other}
\end{array}\right.
\end{align}

These values come from the analysis of the response of the sensors (see Figure
\ref{fig:polaroid}), and from the increase of the width of the cone with the
distance (see \cite{noykov2007calibration}). Then using the sensor
measurements model and the upcoming sensors readings, the updating process of
the probabilistic occupancy grid is done using the Bayes rule:
\begin{align} P[s(C_{ij})=O|r] = \frac{p[r|s(C_{ij})=O]\cdot
P[s(C_{ij})=O]}
 {\sum_{X\in\{E,O\}} p[r|s(C_{ij})=X]\cdot P[s(C_{ij})=X]}\end{align}

\subsection{Fuzzy Maps}

In this case the empty and occupied maps are defined by two fuzzy sets $\mu_E$
and $\mu_O$
\cite{OrioloUliviVendittelli:1997:Fuzzy_maps:_A_new_tool_for_mobile_robot_perception_and_planning,RiboPinz:2001:A_comparison_of_three_uncertainty_calculi_for_building_sonar-based_occupancy_grids};
and two different sensor models are used, one for building the occupied map
$f_O(\rho,r)$, and another for building the empty map $f_E(\rho,r)$.

\begin{align}
 f_O(\rho,r) = & \left\{\begin{array}{ll}
  0, & 0 \leq \rho < r-\delta_r,\\
  k_O\left(1-(\frac{r-\rho}{\delta_r})^2\right), & r-\delta_r \leq \rho <
  r+\delta_r,\\
  0, & \rho \geq r + \delta_r,\end{array}\right.\\
  f_E(\rho,r) = & \left\{\begin{array}{ll}
  k_E, & 0 \leq \rho < r-\delta_r,\\
  k_E(\frac{r-\rho}{\delta_r})^2, & r-\delta_r \leq \rho <
  r+\delta_r,\\
  0, & \rho \geq r, \end{array}\right.
 \end{align}
 $k_O$ and $k_E$ are defined such that $k_O\leq 1$ and $k_E \leq 1$

Then for each reading $r$ the partial maps are built using the sensor model
and the confidence in the reading $\lambda =\Gamma(\rho)\cdot\Delta(\theta)$.
\begin{align}
\mu_O^r(C_{ij}) = \Gamma(\rho)\cdot\triangle(\theta) \cdot f_O(\rho,r),\\
\mu_E^r(C_{ij}) = \Gamma(\rho)\cdot\triangle(\theta) \cdot f_E(\rho,r).
\end{align}

To aggregate several readings in a global map, in \cite{RiboPinz:2001:A_comparison_of_three_uncertainty_calculi_for_building_sonar-based_occupancy_grids} the authors propose to use a t-conorm
$S$, for example $S(a,b) = a +b -a\cdot b$.

\begin{align*}
 \mu_O = \overset{n}{\underset{i=1}{\bigcup}}\mu_O^{r_i} & &
 \mu_E = \overset{n}{\underset{i=1}{\bigcup}}\mu_E^{r_i}
\end{align*}

In fact, fuzzy measures need less axioms than probabilities, so a wider choice
of operators are available for modeling the imprecision, and for aggregating
information from different sources
\cite{PraTriGuaRen:2007:On_fuzzy_set_theories}.

\subsection{Antonyms in Fuzzy Logic}

\label{sec:antonyms}

Nature shows a lot of geometrical symmetries; the mathematical concept of
symmetry is of paramount importance for nature's scientific study. Underlying
the concept of symmetry is the concept of opposite. Humans tend to perceive
and categorize the world by means of opposite concepts and to find different
kind of symmetries. Linguistic expressions that describe these perceptions and
concepts have, therefore, incorporated antonyms to express opposite meanings.

Antonymy is a phenomenon of language based on pairs of opposite words $(P,Q)$
called pairs of antonyms by grammarians.  Antonyms (or opposites) play a key
role in perceptions and knowledge organization
\cite{Cruse:2000:Meaning_in_Language}. Antonyms have not received too much
attention form the point of view of classical logic, perhaps because of the
barely syntactical character of that phenomenon in language. However, fuzzy
logic deals more with semantical aspects of language than classical logic, and
the concept of antonym was early considered in fuzzy logic by Zadeh in
\cite{Zadeh:1975:The_concept_of_Linguistic_Variable}.

The importance of antonyms in linguistics has been studied by Lyons
\cite{Lyons:1977:semantica}, Lehrer \cite{Lehrer:1985:Markedness_and_antonymy}
and many others, and is evidenced by the fact that there are many dictionaries
of antonyms and synonyms. In this context an antonym of a word (or term) $P$
is defined as an opposite word (or term) $aP$, and it can be related
``opposite of meaning" with ``symmetry of use" between $P$ and $aP$. In fact,
to fully understand the meaning of a predicate $P$, one usually need to
understand the meaning of one of its antonyms $aP$; that is, we need to know
both, the use of $P$ and the use of $aP$, because their uses are entwined. In
some cases the antonym $aP$ coincides with the negation of $P$, but this is a
limited case and only occurs when there is not middle term, such as in
{\it(dead, alive)} or {\it(even, odd)}.

In fuzzy logic, one of the most important values of pairs of antonyms is that
they allow us to build linguistic variables
\cite{Zadeh:1975:The_concept_of_Linguistic_Variable}. A linguistic variable is
a set of linguistic labels usually built from a pair of antonyms, the medium
term (when it exists), a set of linguistic modifiers (i.e. not, very,
moderately, quite, ...) and combinations of them by conjunctions (i.e. and,
or). For example, the linguistic variable ``Height'' may be defined by the set of linguistic labels
\{tall, short, medium, very tall, moderately tall, quite tall, not very short,
medium or tall, ...\}. For more details about modifiers and about linguistic
variables, see \cite{Zadeh:1975:The_concept_of_Linguistic_Variable}, and for
more details about conjunctions and disjunctions, see
\cite{AlsTriVal:1983:On_some_logical_connectives_for_fuzzy_sets_theories,GuaRenTri:2006:Some_Fuzzy_Counterparts_of_the_Language_uses_of_And_and_Or}. Different models of antonyms have been studied in the context of fuzzy logic
in \cite{DeSotoTrillas:1999:On_Antonym_and_negate_in_fuzzy_logic, TriCub:2000:On_a_Type_of_Antonymy_in_F([ab])}, which later were compiled
and extended in \cite{TriMorGuaCubCas:2007:Computing_with_Antonyms}.

\tab Language is too complex to admit strictly formal definitions
\cite{TriRenGua:2006:Fuzzy_Sets_vs_Language} as it happens for antonyms.
Nevertheless, there are some properties that different authors attribute to a
pair of antonyms (see \cite{Lehrer:1985:Markedness_and_antonymy}). These
properties were summarized in
\cite{TriMorGuaCubCas:2007:Computing_with_Antonyms}, but we will use here the
following two:

\begin{enumerate}
 \item Involution: In considering a pair $(P,Q)$ of antonyms, $Q$ is an antonym of
    $P$, and $P$ is an antonym of $Q$.
    $$P=a(aP)\mbox{, and }Q=a(aQ)$$
 \item Coherence with the negation: Pairs of antonyms $(P,Q)$ are N-contradictory  \cite{TriAlsJacas:1999:On_contradiction_in_fuzzy_logic,GuaTriRen:2002:Non_Contradiction_and_Excluded_Middle_with_antonyms}, for certain strong negation; it is expressed by:
    $$P\leq not\,Q \mbox{, and }Q \leq not\,P$$(see for example the pair of antonyms $(Far, Near)$ in Figure \ref{fig:distances} or $(None, Some)$ in Figure \ref{fig:times}).

    Therefore they should verify that $P \wedge Q \leq not\,P \wedge not\,Q $, for some conjunction $\wedge$ usually a
    t-norm.
\end{enumerate}

For a more complete study of fuzzy models of antonyms, further examples and proofs
see \cite{TriMorGuaCubCas:2007:Computing_with_Antonyms}. In the next section we
will try to take advantage of these properties to obtain a proper
representation of the pair of antonyms $(Occupied, Empty)$.

\section{Antonyms-based Fuzzy Maps}
\label{sec:approximate_maps}

Let us make a reflection about the nature of the concepts occupied and empty
in the context of occupancy grid maps for robot navigation. We consider that
occupied and empty are antonyms and not complementary. First, because if some
space is not being occupied, we cannot infer that it is empty, it could be
unknown or ambiguous, and second because there is a middle term which
represents the cells that are not occupied and not empty. A cell may be
unknown if the robot perception has not reached it yet, and may be ambiguous
if it is partially occupied and partially empty, so it cannot be assigned to
any of them without introducing errors.

In consequence, we propose to build a pair of fuzzy models, one of obstacles
and another one of empty-spaces, that verify the properties of antonyms. If we
want that occupied and empty maps form a pair of antonyms, they must fulfill
certain conditions: involution and N-Contradiction; they were explained in Section \ref{sec:antonyms}. In the following sections we present how
these maps can be built from observations, and how they can be refined by
properly handling the contradictions that appear from the data during the
process.

First, we defined a set of linguistic rules that allowed us to build obstacles
and empty maps in a coherent way. And second, we defined another set of
linguistic rules to build a contradictions map, that allows the robot to
handle the contradictions that arise in the process, and an integrated map,
that allows the robot to integrate properly the obstacles and empty space maps
by removing the contradictions.

\subsection{Building the obstacles and empty-space maps}

Let us now introduce a linguistic description of how the robot should model
its perceptions of obstacles and empty spaces. By means of this antonyms-based
model we will be able to model properly the imprecision inherent in the robot
perception process. The ways obstacles and empty-spaces are modeled are
different since the information provided by sonar sensors must be interpreted
differently for obstacles than for empty-spaces.

Let us first introduce a linguistic description of the obstacles perception
model from the reading of one ultrasonic sensor. Basically, the model must
assign some degree to the grid cells affected by the sensor reading, that is,
the cells inside the circular sector projected by the sensor cone over the
grid.

\begin{itemize}
\item Confidence rule:
\begin{itemize}
  \item If the measure is small (an obstacle is near), then assign a high confidence to the measure ($\mu_{Near}$).
\end{itemize}
\item Local perception rule:
\begin{itemize}
  \item If the measure is $r$ at angle  $\theta$, then put an obstacle at this position ($\mu_{Occup}$).
\end{itemize}
\item Aggregation rule:
\begin{itemize}
  \item If an obstacle is perceived some times, then increase the confidence on its presence ($\mu_{Some}$).
\end{itemize}
\end{itemize}\vspace{0.2cm}

These rules translate into the following formulas:
\begin{align}
 \mu_{Near}(r) &= \frac{1+tanh(\frac{200-x}{30})}{2}\\
 \mu_{Occup}^k(C_{ij}|(r,\theta)) &= \mu_{Approx}(d,r)\cdot \mu_{Approx}(\alpha,\theta)\\
 \mu_{Some} &= \left\{\begin{array}{cc} 0 & 0\leq x\leq 1 \\ \frac{x-1}{2} & 1<x\leq 3\\
 1 &
 3<x\leq10\end{array}\right.
\end{align}

\begin{figure}[!htb]
\begin{minipage}{.5\linewidth}
\centering\epsfig{file=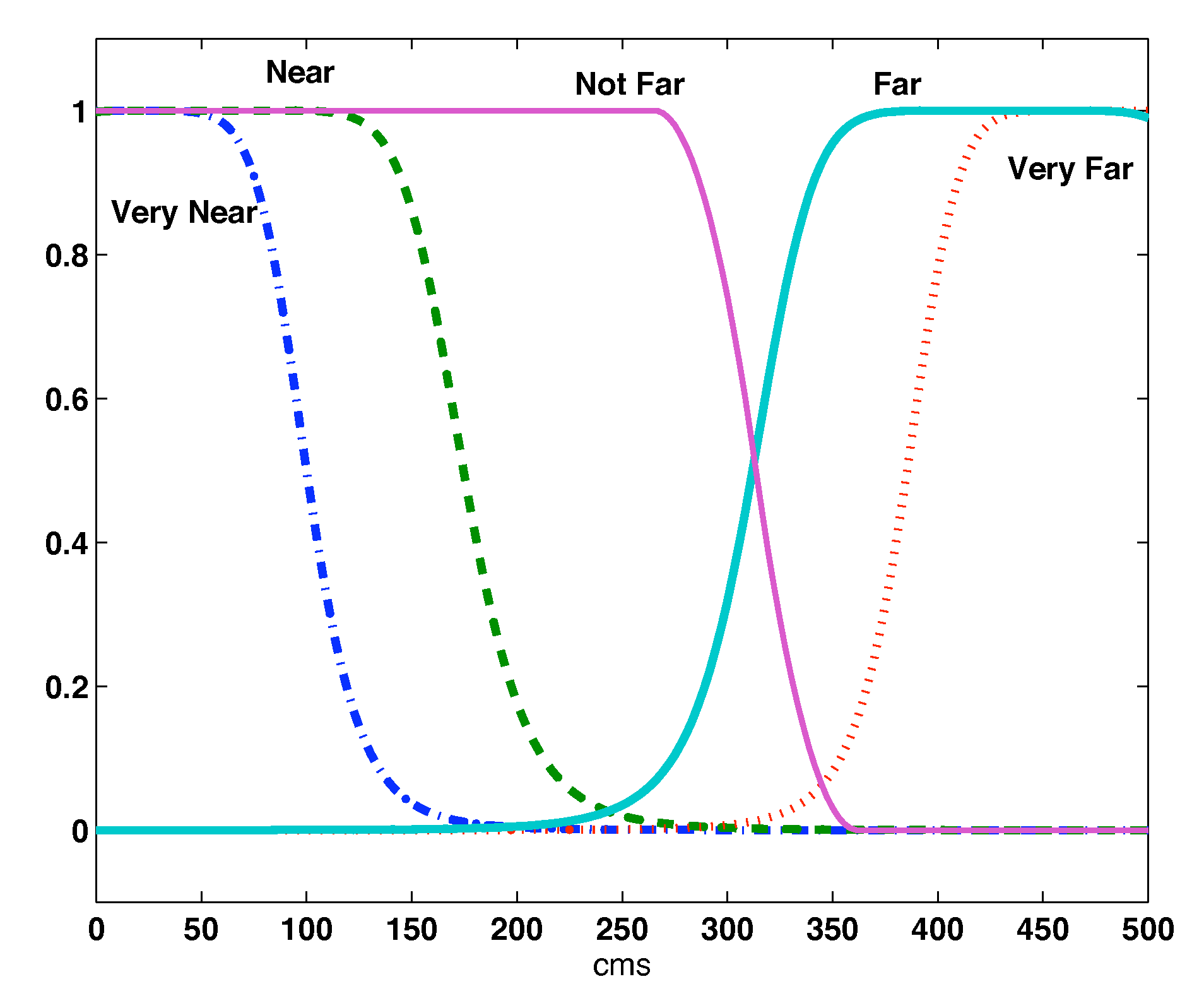,width=\linewidth} \caption{Distance
labels} \label{fig:distances}
\end{minipage}
\begin{minipage}{.5\linewidth}
\centering\epsfig{file=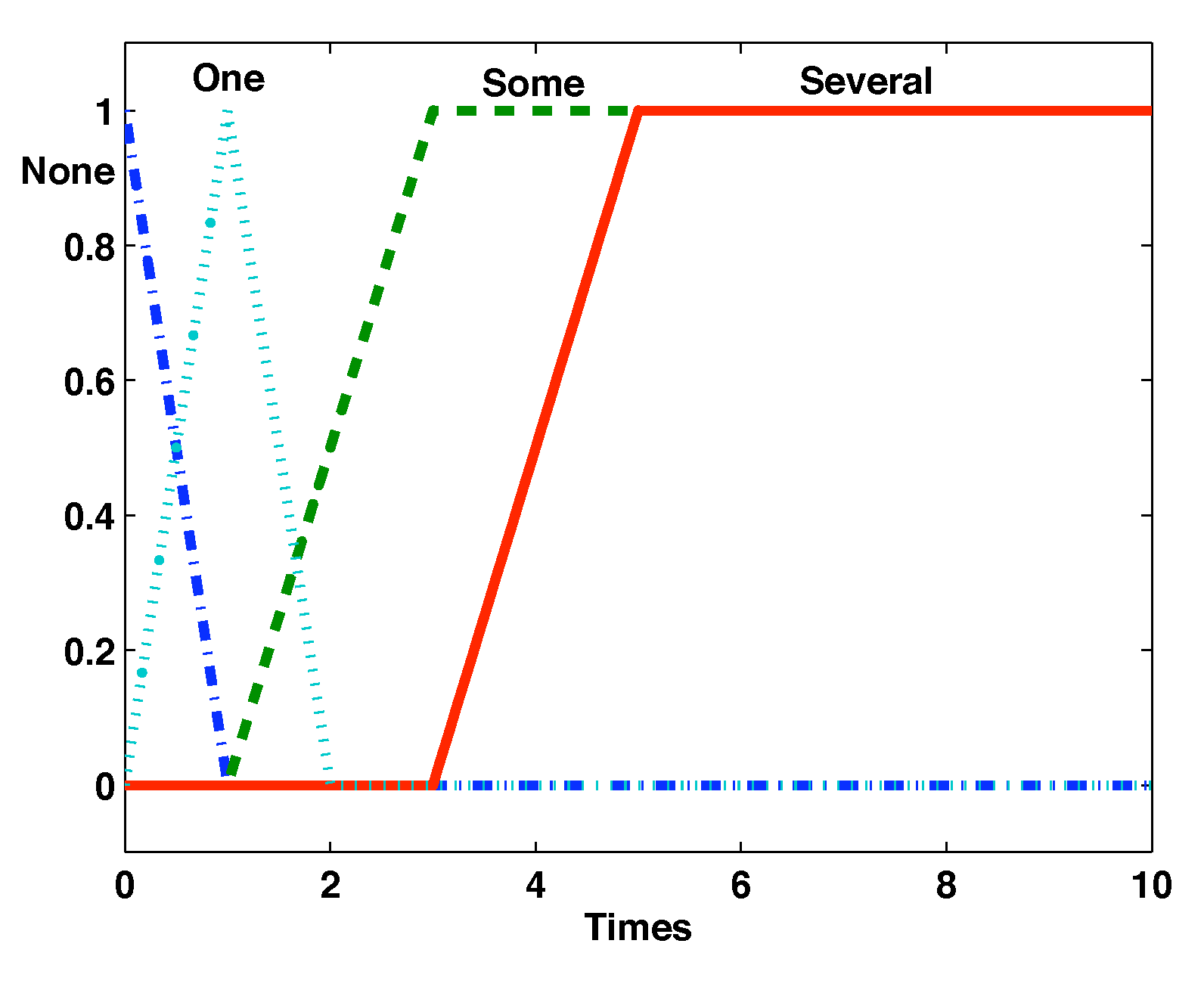,width=\linewidth} \caption{Times labels}
\label{fig:times}
\end{minipage}
\end{figure}

These fuzzy sets (see Figures \ref{fig:distances} and \ref{fig:times}) try to
capture the fact that $not-near$ readings tend to be confusing for the
obstacle map, and that the robot need to see an obstacle more than once
($some$ times) to know that it is there. After analyzing the data coming from
the sensors in different situations, we tuned the parameters of these fuzzy
sets.

Analogously, we established the following linguistic descriptions of how to
model the empty space perception from each ultrasonic sensor:
\begin{itemize}
\item Confidence rule:
\begin{itemize}
  \item If the measure is not big (an obstacle is not far), then assign a high confidence to the measure ($\mu_{Not-Far}$).
\end{itemize}
\item Local perception rule:
\begin{itemize}
  \item If the measure is $r$ at angle $\theta$, then assign empty space inside the circular sector with radius $r$ and aperture $\delta_{\alpha}$ ($\mu_{Empty}$).
\end{itemize}
\item Aggregation rule:
\begin{itemize}
  \item If an empty space is perceived several times, then increase the confidence on its emptiness ($\mu_{Several}$).
\end{itemize}
\end{itemize}

These rules translate into the following formulas:
\begin{align}
 \mu_{Not-Far}(r) &= 1-\frac{1+tanh(\frac{x-300}{30})}{2}\\
 \mu_{Empty}^k(C_{ij}|(r,\theta)) &= \mu_{Smaller}(d,r)\cdot \mu_{Approx}(\alpha,\theta)\\
 \mu_{Several} &= \left\{\begin{array}{cc} 0 & 0\leq x\leq 3 \\ \frac{x-3}{2} & 3<x\leq 5\\
 1 &
 5<x\leq10\end{array}\right.
\end{align}

These fuzzy sets (see Figures \ref{fig:distances} and \ref{fig:times}) try to
capture the fact that $far$ away readings tend to be wrong, and that the robot
needs to see an empty-space $several$ times before knowing that it is empty.
After analyzing the data coming from the sensors in different situations, we
tuned the parameters of these fuzzy sets.

\begin{figure}[!htb]
\begin{minipage}{.5\linewidth}
\centering\epsfig{file=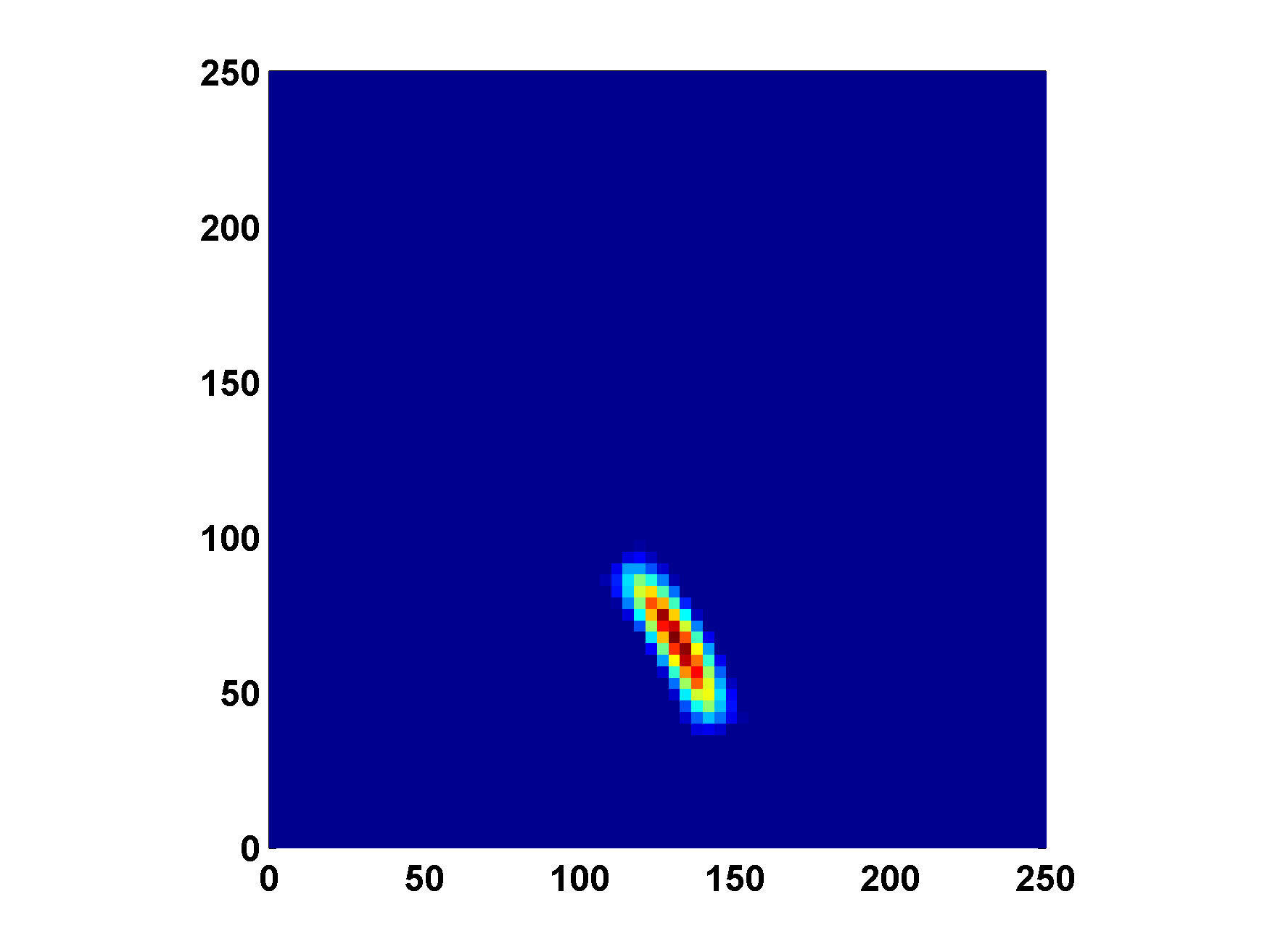,width=\linewidth}
\caption{Obstacle with $(r,\theta)=(150,30^o)$} \label{fig:labelapproxa}
\end{minipage}
\begin{minipage}{.5\linewidth}
 \centering\epsfig{file=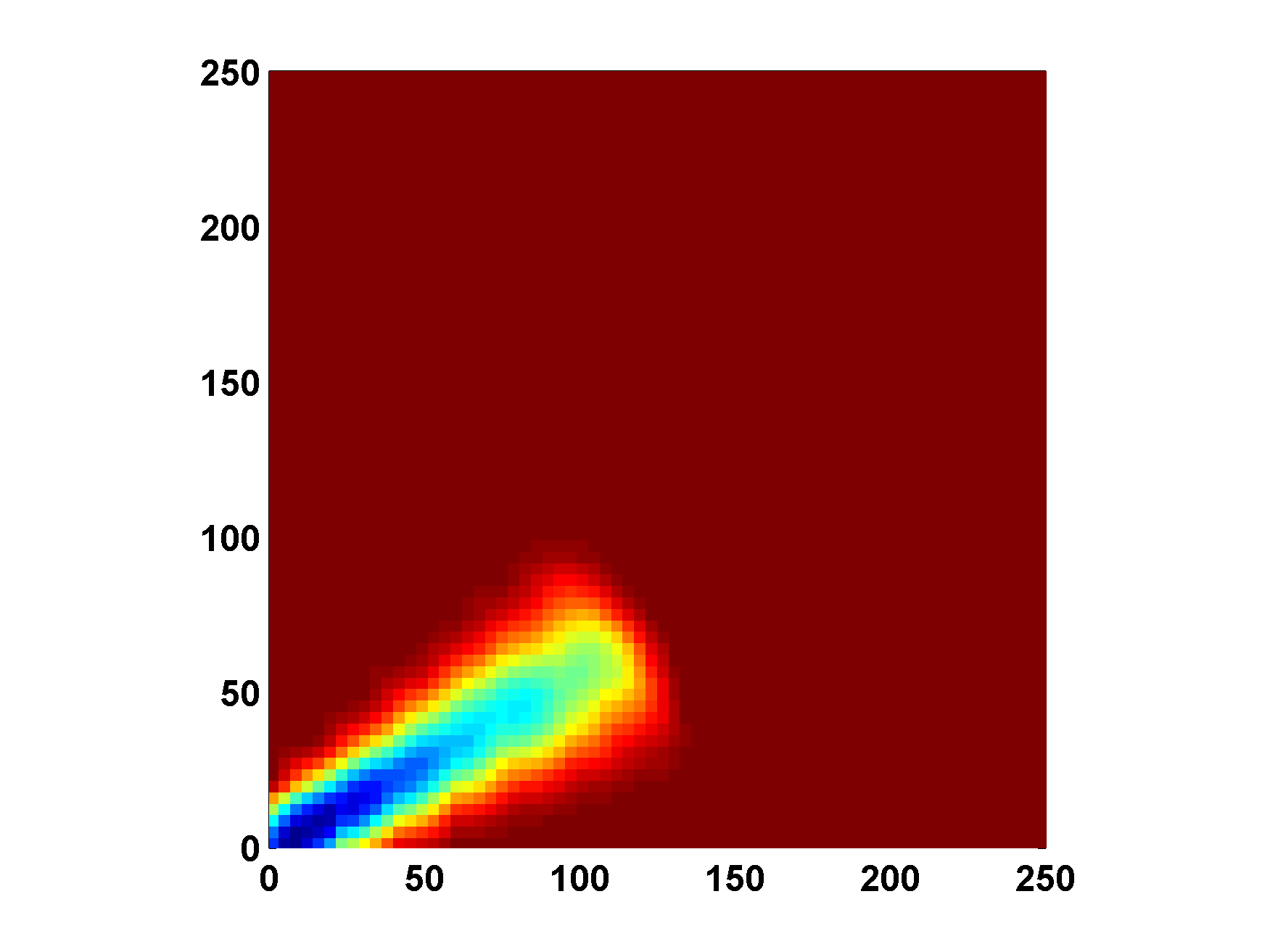,width=\linewidth}
\caption{Empty-space with $(r,\theta)=(150,30^o)$} \label{fig:untild_inisidea}
\end{minipage}
\end{figure}

The linguistic labels used in these rules can be seen in Figures
\ref{fig:distances} and \ref{fig:times}. Examples of the contribution to the
obstacles map and empty-spaces maps produced by a single sonar reading
$\mu_{Occup}^k(C_{ij}|(r,\theta))$ and $\mu_{Empty}^k(C_{ij}|(r,\theta))$ are
presented in Figures \ref{fig:labelapproxa} and \ref{fig:untild_inisidea},
respectively.

These rules try to take into account the problems underlying sonar readings
and allow us to build two approximate maps in such a way that every cell
contains a number between 0 and 1, which represents the degree of the cell
being an obstacle, or the degree of being an empty-space.

In order to define the fuzzy sets $\mu_{Approx}$ an $\mu_{Smaller}$ and to
allow a fair comparison with other approaches, we have used the values of the parameters
$(\delta_r = 15\,cm, \delta_{\alpha}=radians(30^o)=0.2618\, rad)$ that were
used in
\cite{RiboPinz:2001:A_comparison_of_three_uncertainty_calculi_for_building_sonar-based_occupancy_grids}:

\begin{align}
 \mu_{Approx}(d,r) &= 1 - \frac{(d-r)^2}{\delta_r^2} \\
 \mu_{Approx}(\alpha,\theta) &= 1- \frac{|\alpha-\theta|^2}{\delta_{\alpha}^2}\\
 \mu_{Smaller}(d,r) &= 1 - \frac{1+tanh(\frac{(d - r)}{50})}{2}
\end{align}

\noindent $r$ is a given range reading, $d$ is the distance from the
sensor to $C_{ij}$, $\theta$ is the sensor bearing and $\alpha$ is the angle
between the beam axis and $C_{ij}$.

\begin{figure}[!htb]
\begin{minipage}{.5\linewidth}
\centering\epsfig{file=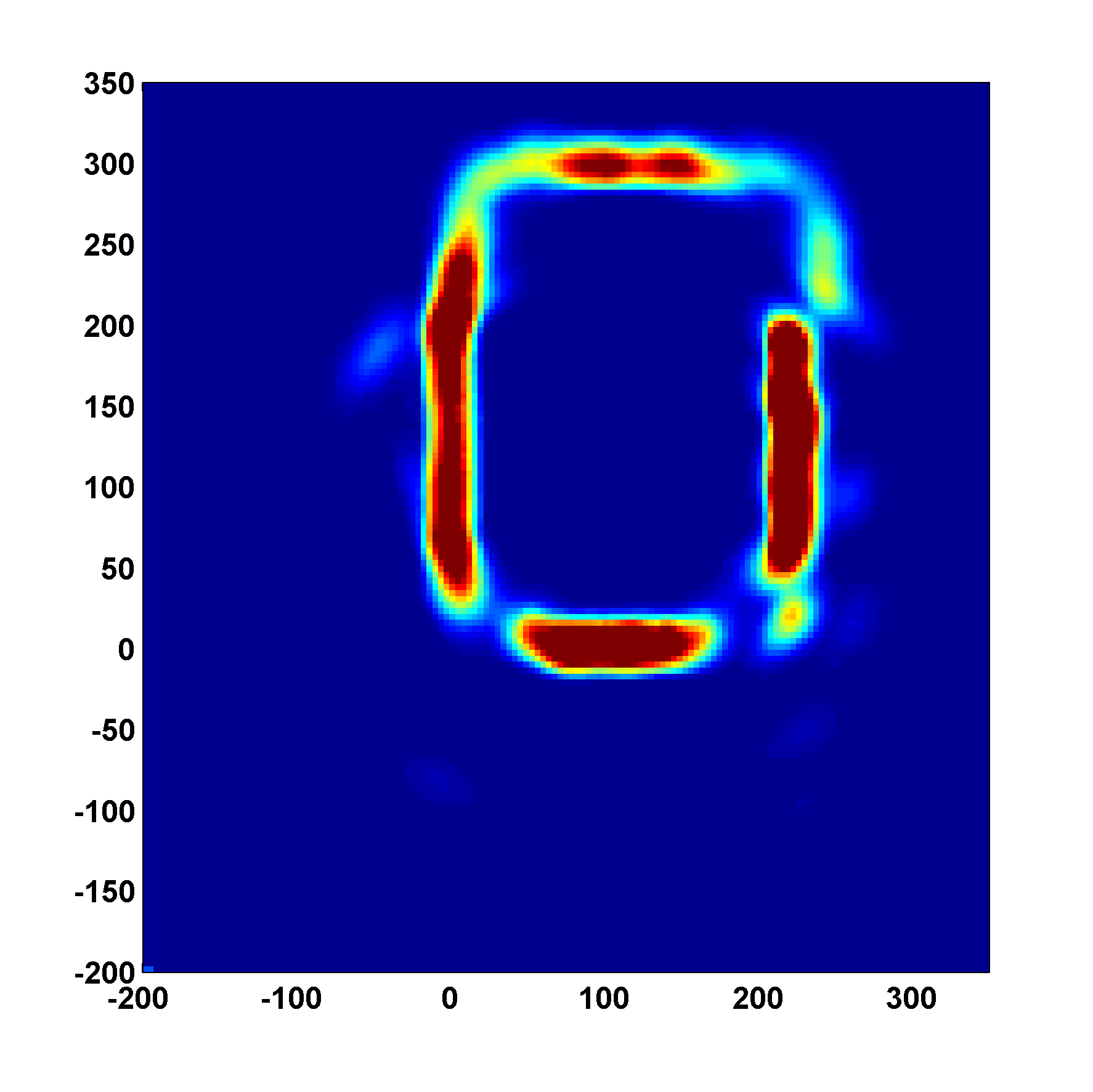,width=\linewidth}
\caption{Obstacles map} \label{fig:obs_room}
\end{minipage}
\begin{minipage}{.5\linewidth}
\centering\epsfig{file=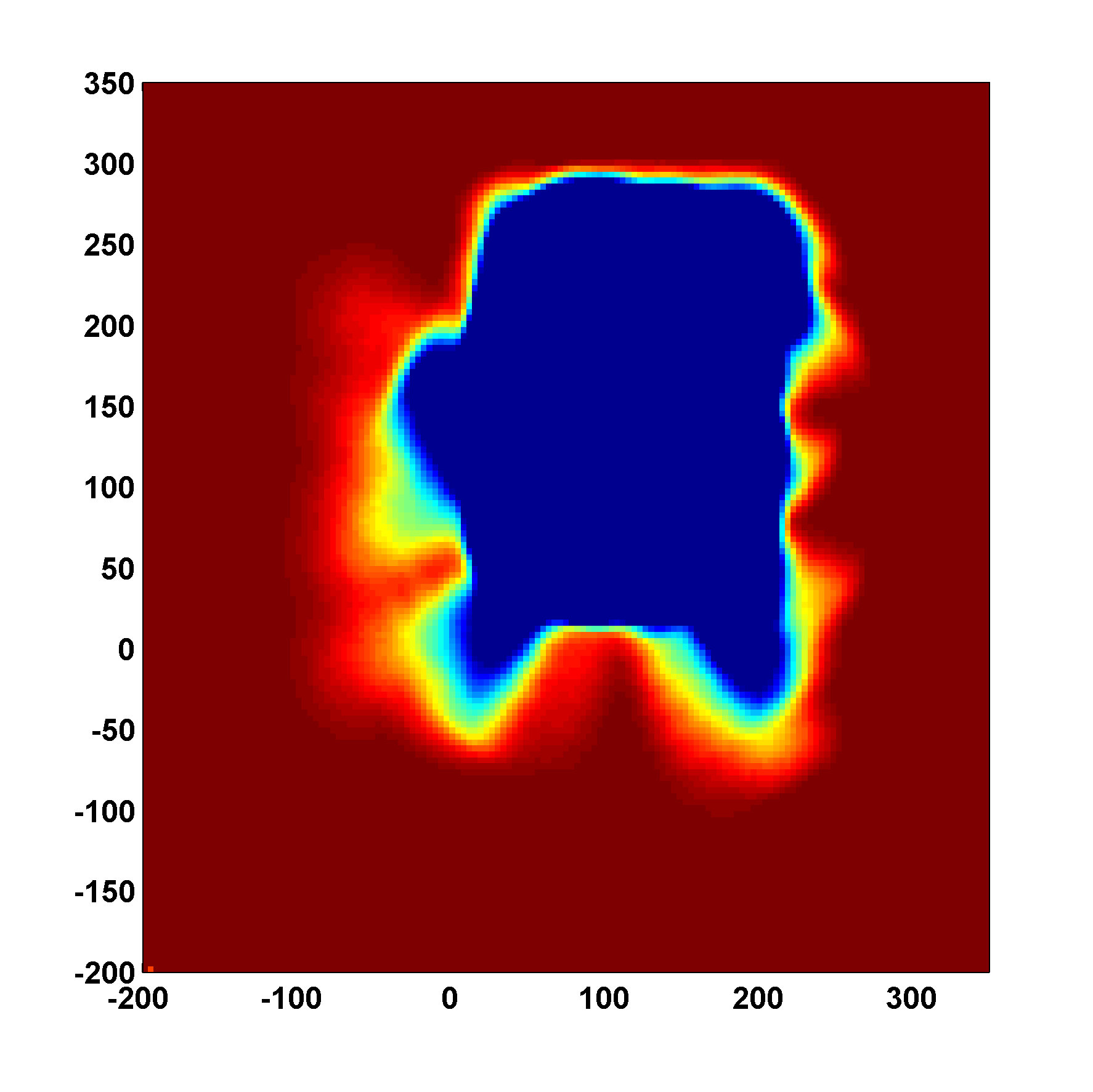,width=\linewidth} \caption{Empty
spaces map} \label{fig:hue_room}
\end{minipage}
\end{figure}

Every cell may receive contributions from several sensor readings. The
approximate maps are built by the aggregation of a sequence of readings
$R=\{(r_1,\theta_1),...,(r_k,\theta_k),...,(r_n,\theta_n)\}$ as follows (see
Figures \ref{fig:obs_room} and \ref{fig:hue_room} for obstacles and
empty-space maps of the Office):

\begin{align}
\mu_{Occup}(C_{ij}|R) = & \mu_{Some}\left(\overset{n}{\underset{k=1}{\sum}}( \mu_{Near}(r_k)\cdot\mu_{Occup}^k(C_{ij}|(r_k,\theta_k)))\right)\\
\mu_{Empty}(C_{ij}|R) =&
\mu_{Several}\left(\overset{n}{\underset{k=1}{\sum}}(\mu_{Not-Far}(r_k)\cdot\mu_{Empty}^k(C_{ij}|(r_k,\theta_k)))\right)
\end{align}

This way of building the aggregated maps by means of linguistic quantifiers
(see \cite{diaz2009fuzzy}) differs greatly from other approaches
\cite{Elfes:1987:Sonar_based_real_world_mapping_and_navigation,OrioloUliviVendittelli:1997:Fuzzy_maps:_A_new_tool_for_mobile_robot_perception_and_planning,Li:2007:A_successful},
and allows us to easily handle the partial contribution of each sonar reading
to the aggregated map.

Figures \ref{fig:obs_room} and \ref{fig:hue_room} show an example of an
obstacles map (obstacles in red) and an empty-space map (empty space in blue),
respectively. In these figures, the limits of the place may easily be
recognized although due to the presence of rebounds and short echoes some
outliers also appear.

\subsection{Building the contradiction map and the integrated map}
\label{sec:con_int}

If one cell is considered simultaneously empty and occupied, then we have a
contradiction in that cell. This situation violates the principle of coherence
with the negation (the value of the cell in one map must be smaller than the
negation of the value in the other map, see Section \ref{sec:antonyms}).

In order to analyze the contradictions we built a contradiction map by the
conjunction of the obstacles and empty space maps. Since all t-norms are
smaller than minimum (see \cite{PraTriGuaRen:2007:On_fuzzy_set_theories}), we
used minimum to obtain the largest degree of contradiction and to work in the
worst case.
\begin{equation}\mu_{Contra} =
min(\mu_{Occup},\mu_{Empty})\end{equation}

The integrated map is built by combining the occupied and empty maps by the
following procedure. In this case, to work in a very restrictive case, we used
the $\L$ukasiewicz t-norm $W(x,y)=max(0,x+y-1)$ (see
\cite{PraTriGuaRen:2007:On_fuzzy_set_theories}).
\begin{enumerate}
 \item Calculate the occupied cells that are not empty:
 \begin{equation}\mu_{Occup^*} = W(\mu_{Occup},1-\mu_{Empty}) =
 max(0,\mu_{Occup}-\mu_{Empty})\end{equation}
 \item Calculate the empty cells that are not occupied:
 \begin{equation}\mu_{Empty^*} =  W(\mu_{Empty},1-\mu_{Occup}) =
 max(0,\mu_{Empty}-\mu_{Occup})\end{equation}
 \item Aggregate both maps in an integrated map $\mu_{Integ}:[cells]\rightarrow[-1,1]$:
 \begin{equation}\mu_{Integ}(C_{ij}) = \left\{
 \begin{array}{ll}
  \mu_{Occup^*}(C_{ij}) & \mbox{ if } \mu_{Occup}(C_{ij}) > \mu_{Empty}(C_{ij})\\
  -\mu_{Empty^*}(C_{ij}) & \mbox{ if } \mu_{Occup}(C_{ij}) \leq \mu_{Empty}(C_{ij})
 \end{array}
 \right.\end{equation}
\end{enumerate}

\begin{figure}[!htb]
\begin{minipage}{.5\linewidth}
\centering\epsfig{file=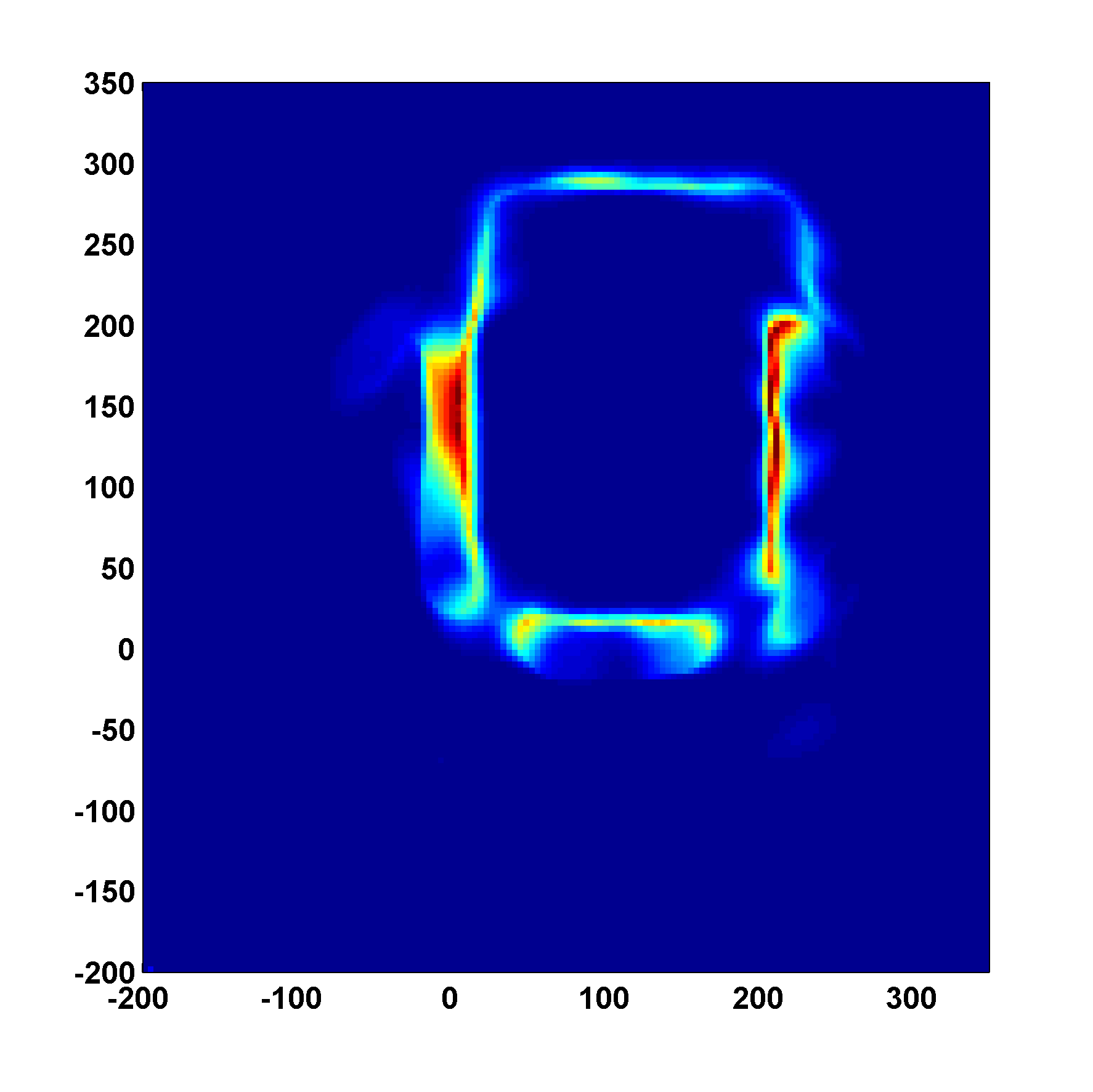,width=\linewidth}
\caption{Contradictions map} \label{fig:con_room}
\end{minipage}
\begin{minipage}{.5\linewidth}
\centering\epsfig{file=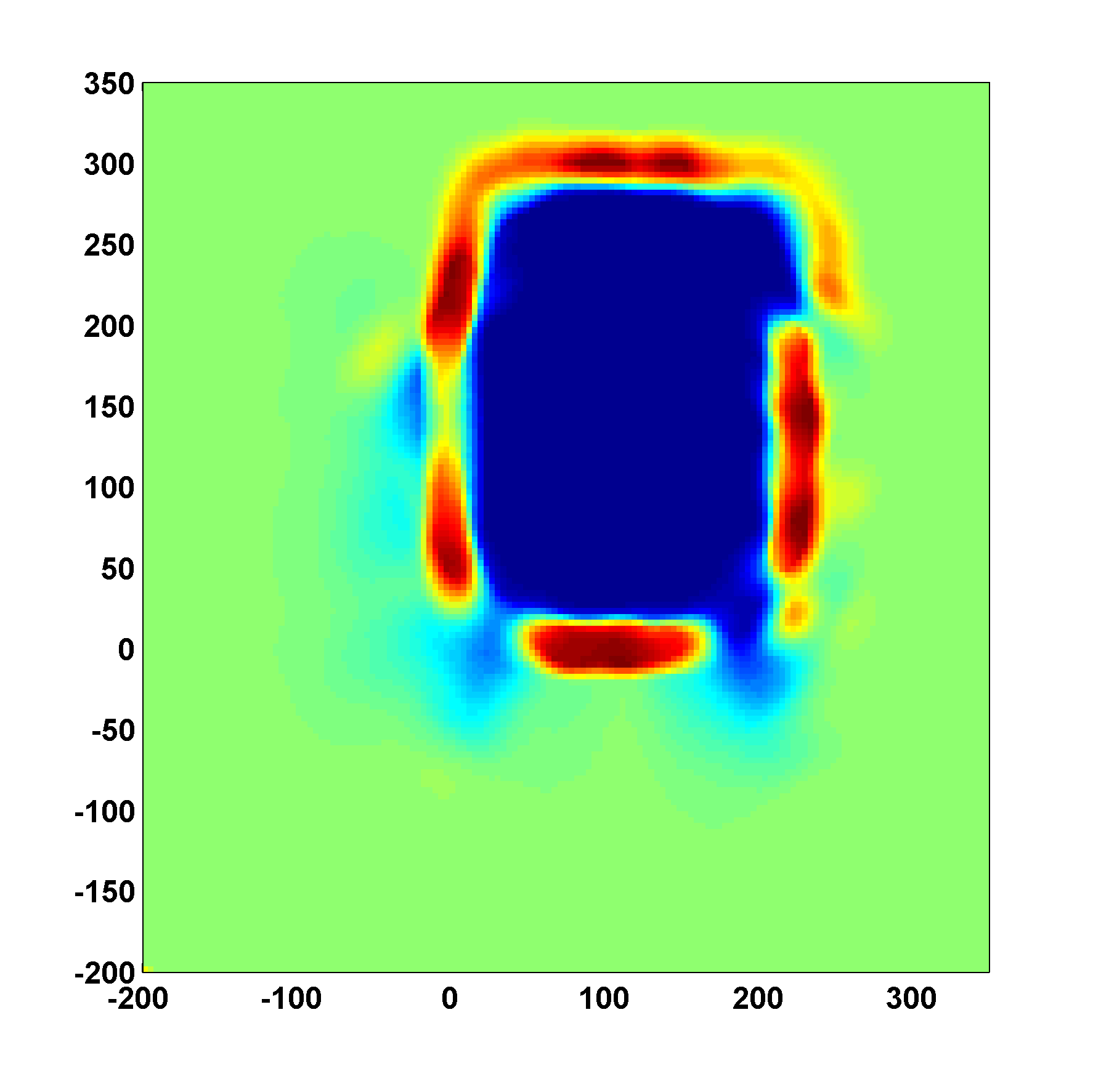,width=\linewidth}
\caption{Integrated map} \label{fig:int_room}
\end{minipage}
\end{figure}

The integrated map (see Figure \ref{fig:int_room}) is defined from -1 to 1, with -1 meaning that the cell is completely empty (represented as blue in the figures), 1 meaning that the cell is completely occupied (represented as red in the figures) and 0 meaning that the cell is unknown (represented as
green in the figures). Cells can be unknown due to two reasons: first because the cell's state is ambiguous (or contradictory) since the cell is occupied and empty simultaneously, and second because the robot has never explored that zone and
the cell's state is neither occupied nor empty.

It can be seen in the contradictions map, Figure \ref{fig:con_room}, that there are many contradictions between obstacles and empty-space maps and, therefore, that the integrated map plotted in Figure \ref{fig:int_room} contains several errors. For instance, there is a hole in the left wall, and there are obstacles around corners.

Analyzing the contradictions map we can see that it reflects the existence of
errors and imprecision. If we could have perfect knowledge and there were no
errors, then there would be no contradictions. These contradictions are usually
solved, in a conservative approach, by marking them as unsafe to avoid
collisions (see
\cite{GasosMartin:1997:Mobile_robot_localization_using_fuzzy_maps} or
\cite{RiboPinz:2001:A_comparison_of_three_uncertainty_calculi_for_building_sonar-based_occupancy_grids}).
One problem of this conservative approach is that when a cell is marked
unsafe, it eventually will remain that way forever.

There are two known situations that cause cell contradictions: rebounds and
short echoes. We can model our knowledge about them in order to correct as many
contradictions as possible.

A rebound is a range larger than the real distance to the obstacle. It
generates some false empty cells inside its circular segment and false
occupied cells in its arc. We have empirically observed that it is very
infrequent to obtain a rebound in a short range reading (bellow more or less
150 centimeters), this explains the formula choused for the confidence on readings $\Gamma(\rho)$.

A short echo is a range shorter than the real distance to the obstacle. Short
echoes are mainly produced by the significant aperture of ultrasound sensors.
Although an echo is usually produced only at one point of the wavefront, several grid
cells (those of the arc) are assigned as occupied. This effect is more
dramatic when the echo is produced at one of the sides of the wavefront. In
general, this phenomenon generates false occupied cells along of the arc. Short
echoes are perceived as false obstacles that habitually disappear if the robot
gets closer to them (bellow more or less
150 centimeters). A typical example are corners, perceived from far as
round shapes.

\begin{figure}[!htb]
\begin{minipage}{.5\linewidth}
\centering\epsfig{file=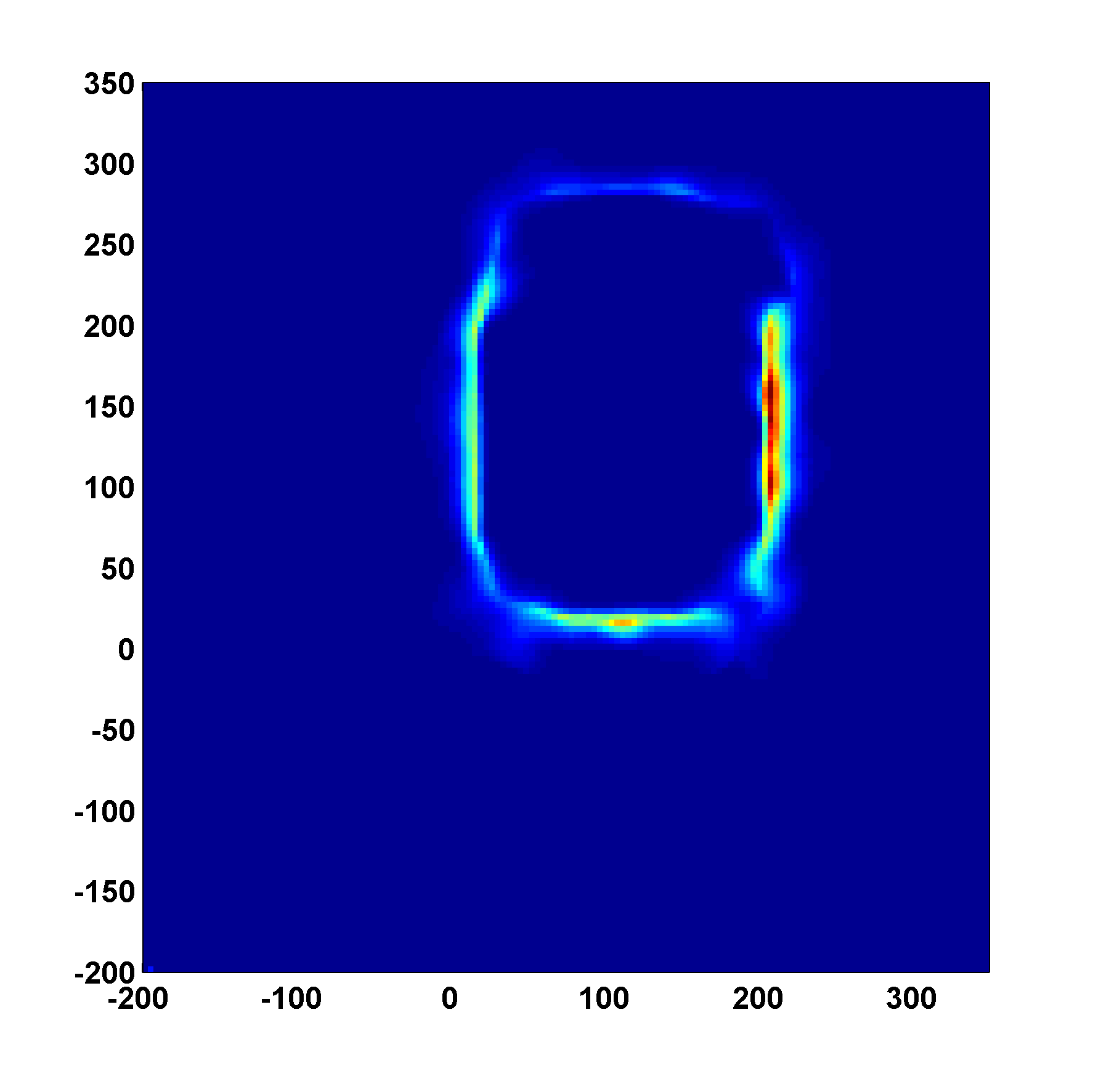,width=1\linewidth} \caption{Errors
due to Short-Echoes} \label{fig:short-echos}
\end{minipage}
\begin{minipage}{.5\linewidth}
\centering\epsfig{file=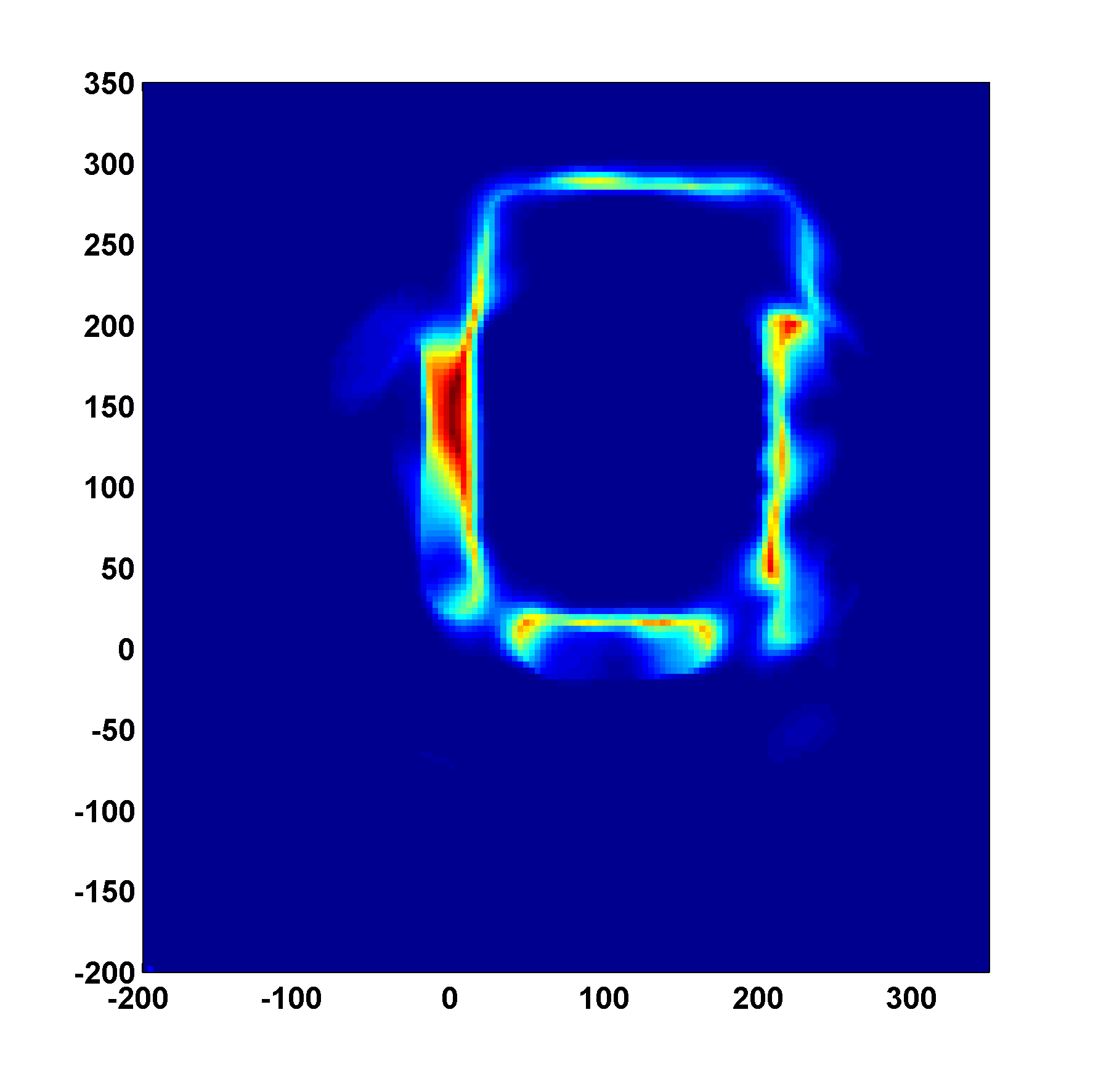,width=1\linewidth} \caption{Errors
due to Rebounds} \label{fig:rebounds}
\end{minipage}\vspace{0.2cm}
\end{figure}

False occupied cells and false empty cells can be detected if they have been
correctly captured by another range. To ensure this correction, we
additionally require that the other range is obtained near the cell.

We model this knowledge about rebounds and short echoes by means of the
following fuzzy rules:

\begin{itemize}
 \item If one cell is occupied and empty, but from near looks empty, then it is a short-echo (a false obstacle) (see Figure \ref{fig:short-echos}).
 \item If one cell is occupied and empty, but from near looks occupied, then it is a rebound (a false empty-space) (see Figure \ref{fig:rebounds}).
\end{itemize}\vspace{0.2cm}

\begin{figure}[!htb]
\begin{minipage}{.5\linewidth}
\centering\epsfig{file=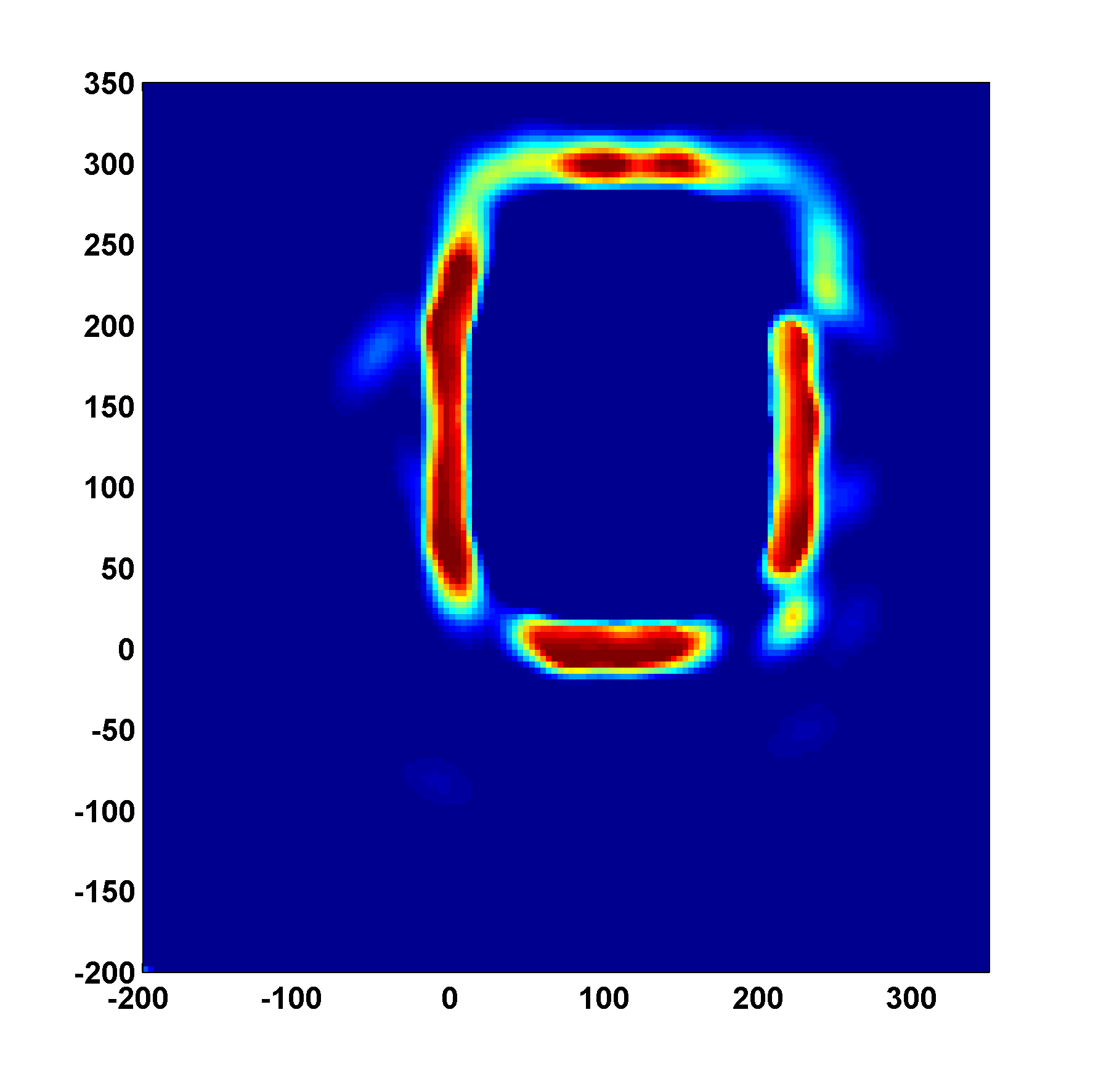,width=\linewidth}
\caption{Corrected obstacles map} \label{fig:obs_rellano2}
\end{minipage}
\begin{minipage}{.5\linewidth}
\centering\epsfig{file=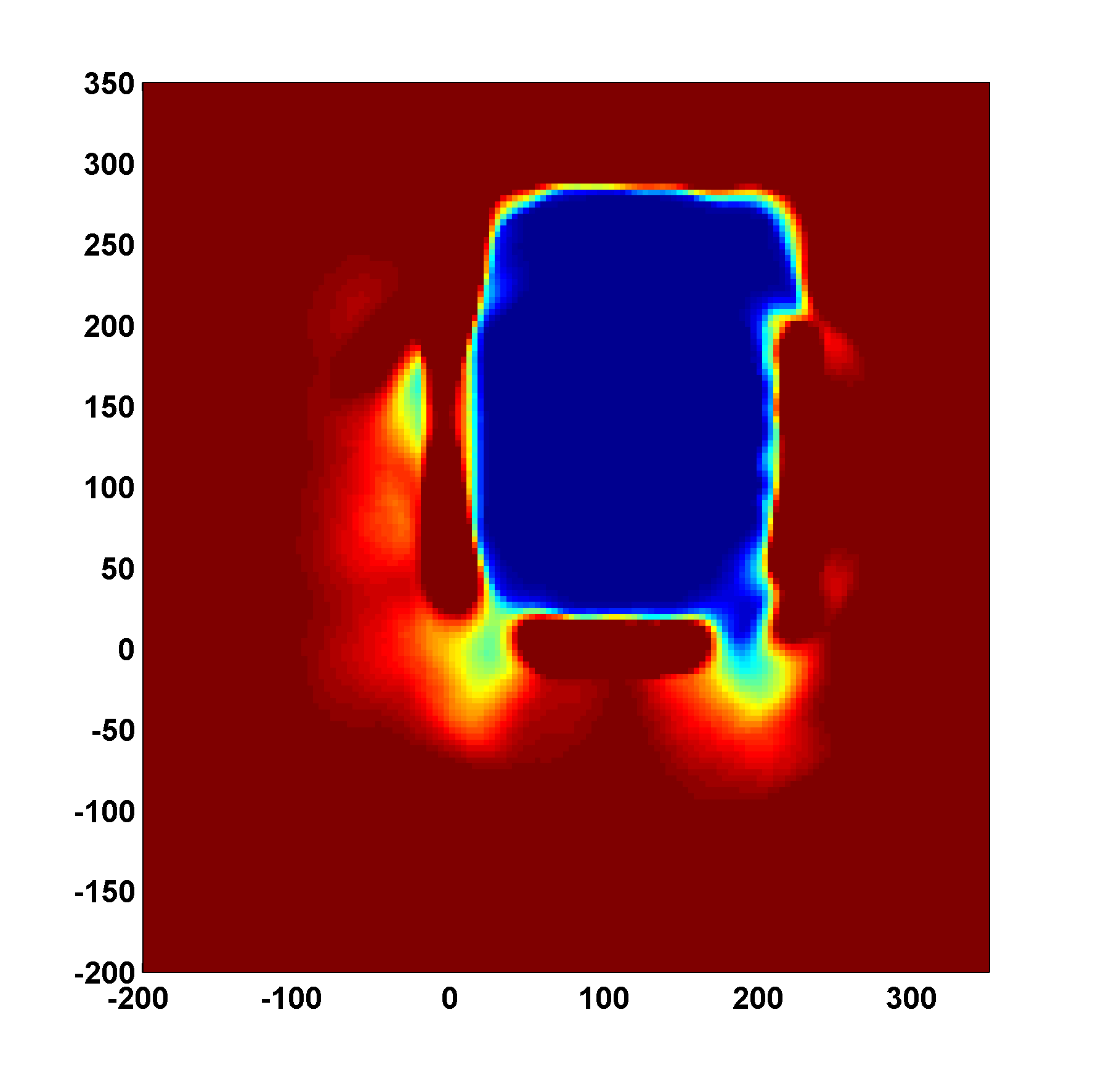,width=\linewidth}
\caption{Corrected empty-spaces map} \label{fig:hue_rellano2}
\end{minipage}\end{figure}

\begin{figure}[htb]
\begin{minipage}{.5\linewidth}
\centering\epsfig{file=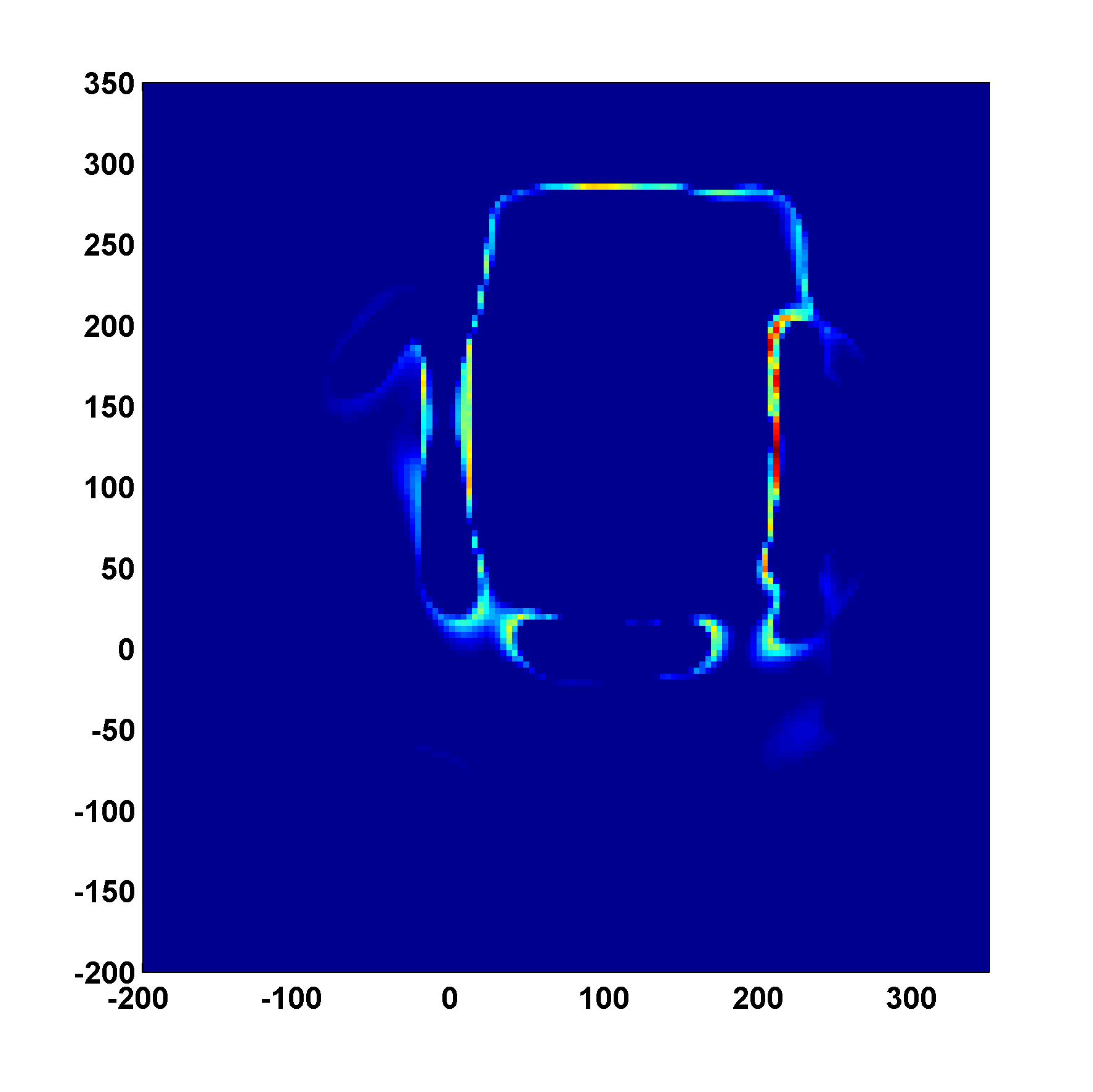,width=\linewidth}
\caption{Contradictions map after corrections} \label{fig:con_rellano2}
\end{minipage}
\begin{minipage}{.5\linewidth}
\centering\epsfig{file=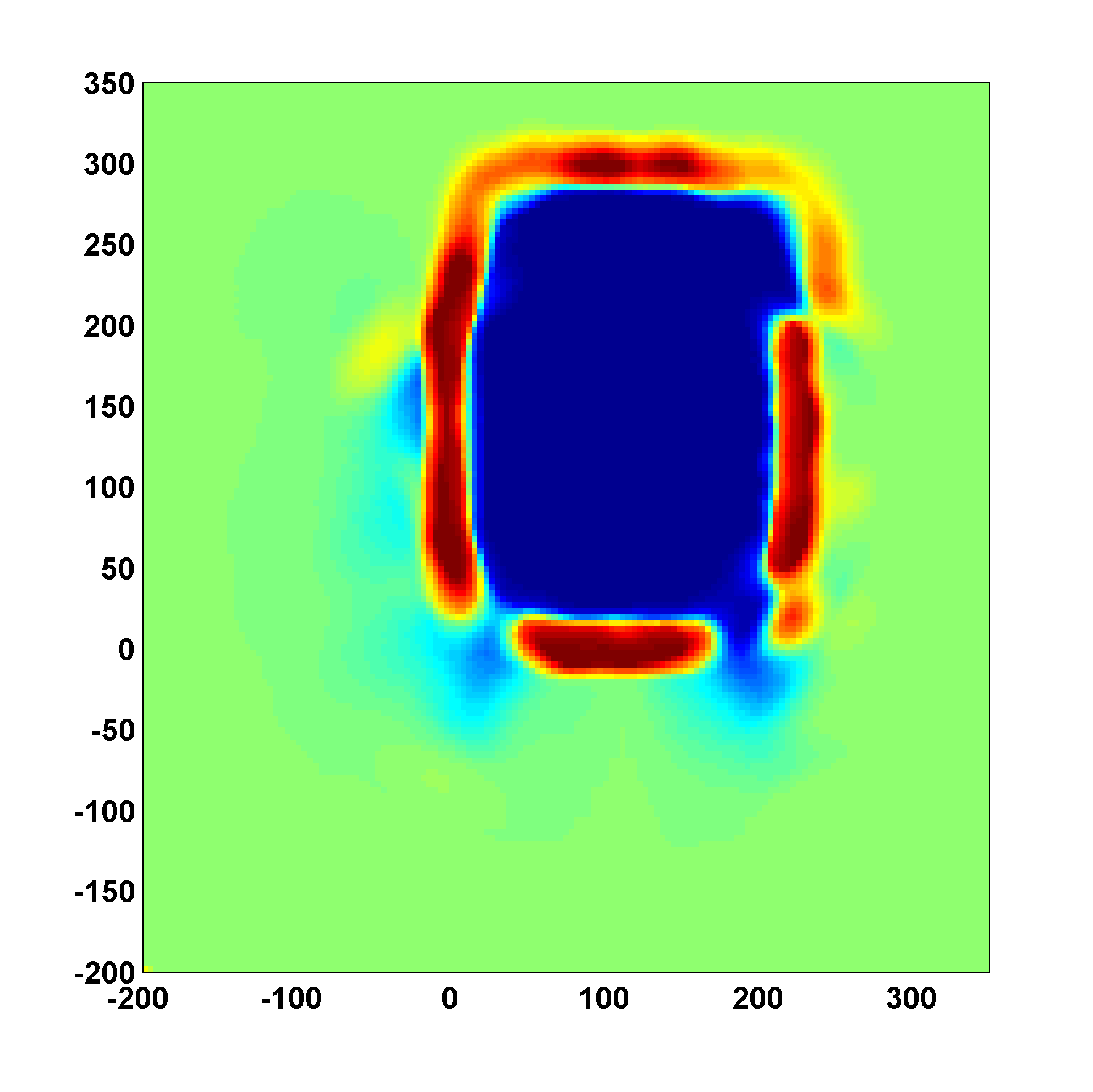,width=\linewidth}
\caption{Integrated map after corrections} \label{fig:fused_rellano2}
\end{minipage}
\end{figure}

Using these rules we can remove false obstacles from the obstacles map and
false empty cells from the empty-space map (see Figures \ref{fig:obs_rellano2}
and \ref{fig:hue_rellano2}). After removing these false obstacles and false
empty cells we obtained a new contradiction map and integrated map shown in
Figures \ref{fig:con_rellano2} and \ref{fig:fused_rellano2}, which show less
contradictions and contains less errors, respectively.

\section{Experiments and results}
\label{sec:experiments}
\subsection{System Description}
\label{sec:system}

\begin{figure}[htb]
 \centering\includegraphics[width=.6\linewidth]{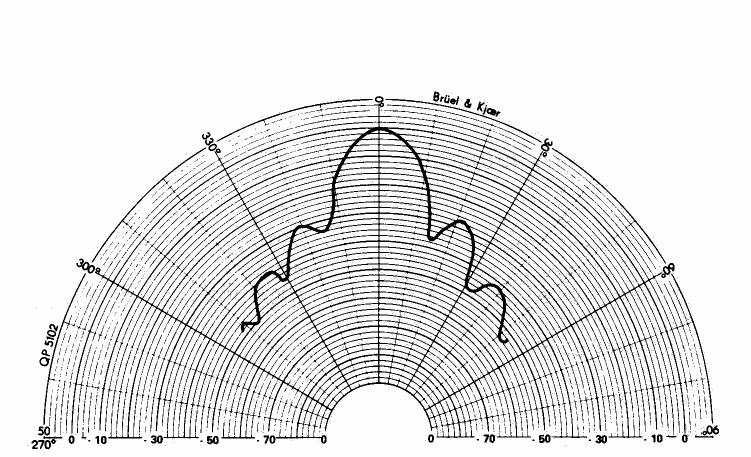}
\caption{Polaroid 6500 sonar beam (extracted from
\cite{CaoBorenstein:2002:Experimental_Characterization_of_Polaroid_Ultrasonic_Sensors})}\label{fig:polaroid}
\end{figure}

The proposed model have been validated by means of experimentation with real
data. In order to obtain data-sets from real environments, the robot
Sancho-2\footnote{Sancho-2 has been designed and built by Gracian Trivi\~{n}o at
the Universidad Polit\'{e}cnica de Madrid \cite{tesisgracian}}, an indoor mobile
robot for research purposes, has been used. It is a 50 cm x 50 cm x 50 cm
platform with motor and sensory capabilities, over which a laptop PC is placed
and connected through a serial connection, for controlling the robot. The
wheels are disposed in tricycle configuration with two driven wheels and one
passive castor wheel. Each driven wheel is controlled in closed-loop by means
of a coupled encoder disk. The resolution of these odometers, projected over
the floor, is 1.2 cm. The sensory capability is disposed by a ring of twelve
ultrasonic, or sonar, Polaroid 6500 sensors distributed at angles of 30$^o$
around it. A detailed description of the Polaroid ultrasonic sensor may be
found in
\cite{CaoBorenstein:2002:Experimental_Characterization_of_Polaroid_Ultrasonic_Sensors}.
The range resolution produced by our sensor control hardware is 4 cm, and the
aperture of the transducer's main lobe (cone) is 30$^o$ at 3dB although in
some cases the second and even the third lobe can be reflected and received
(see Figure \ref{fig:polaroid}). This causes considerable imprecision about
the location of the object that generates the returned echo. This imprecision
grows with the measured range.

\subsection{Experiments setup}

\begin{figure}[!htb]
\begin{minipage}{.55\linewidth}
 \centering\epsfig{file=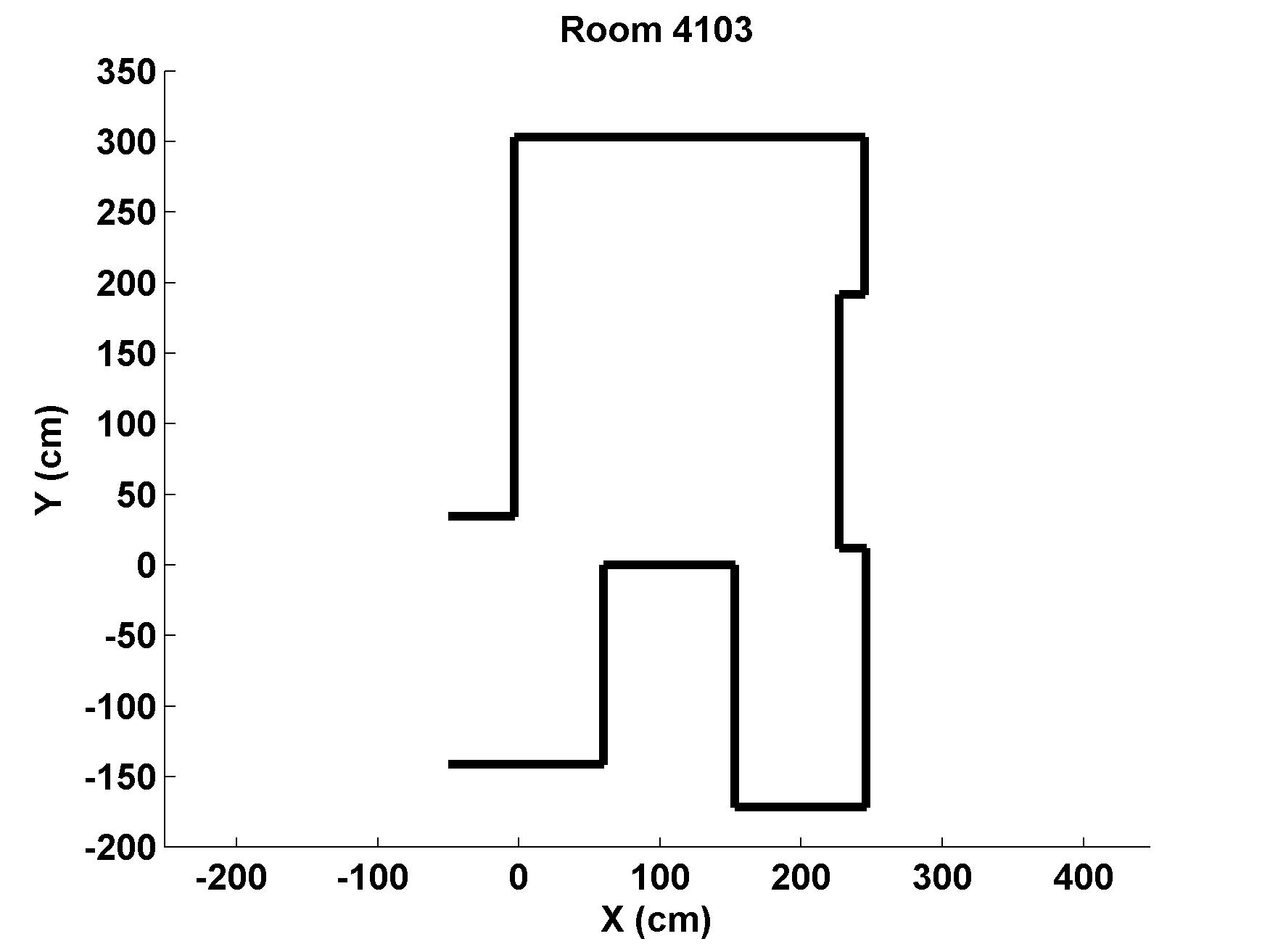,width=\linewidth}
\caption{Office}\label{fig:room4103} \vspace{0.8cm}
 \centering\epsfig{file=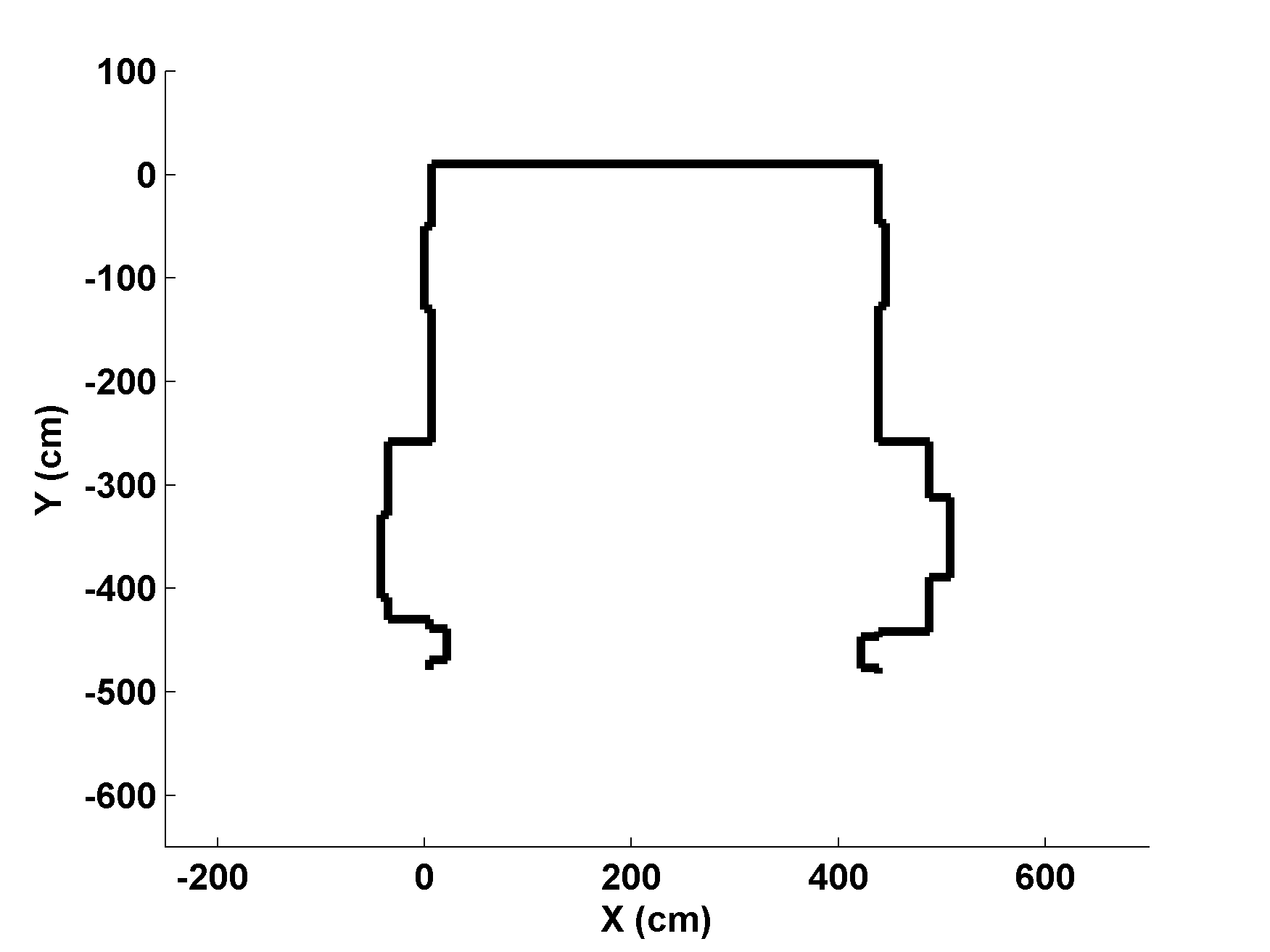,width=1\linewidth}
\caption{Hall}\label{fig:hall}
\end{minipage}\hspace{0.5cm}
\begin{minipage}{.45\linewidth}
 \centering\epsfig{file=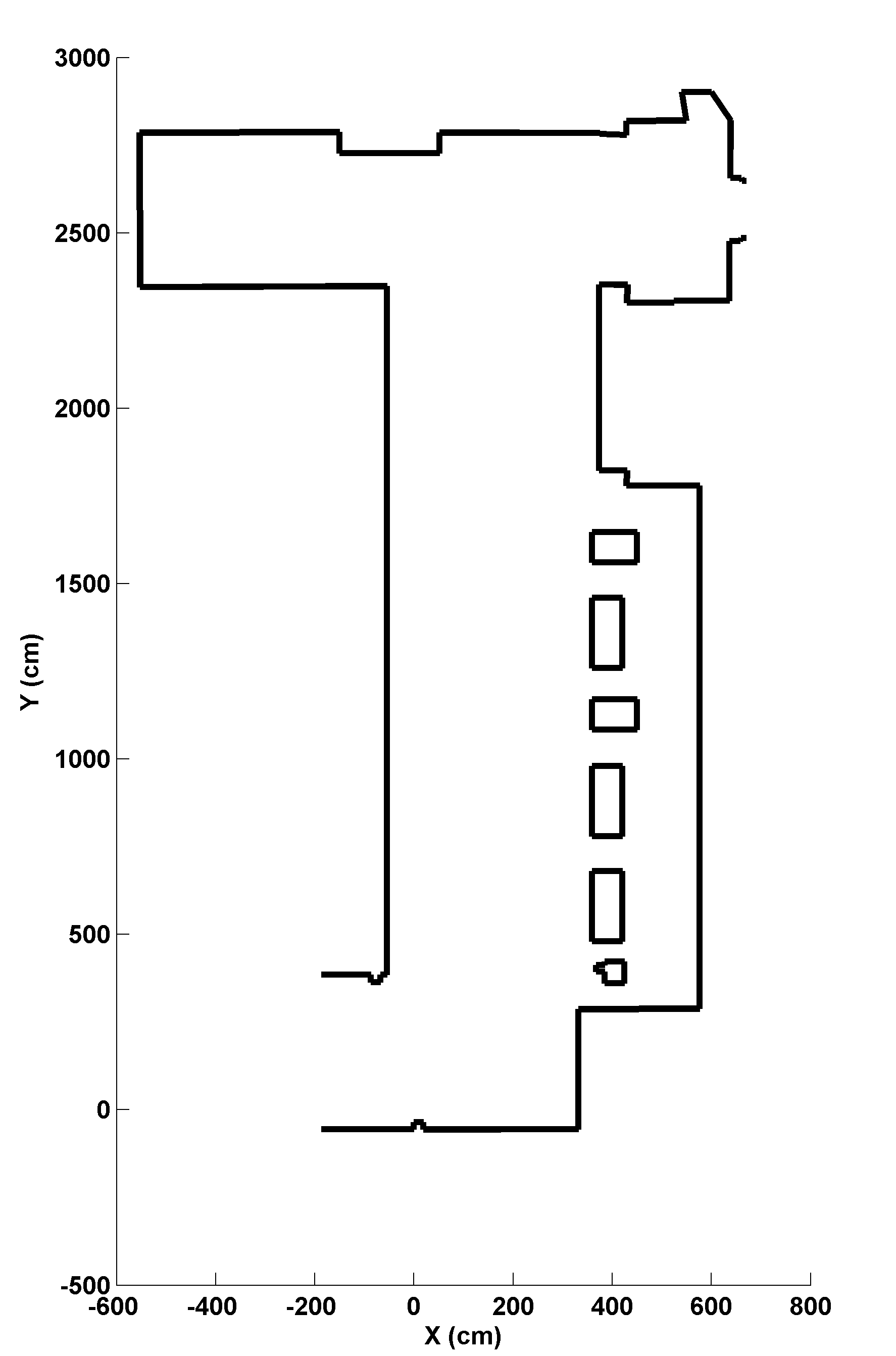,height=13cm,width=\linewidth}
\caption{Corridor}\label{fig:corridor}
\end{minipage}
\end{figure}

Several controlled experiments have been performed to validate the proposed
mapping approach. Three real indoor places: an office (see Figure
\ref{fig:room4103}), a medium sized hall (see Figure \ref{fig:hall}) and a
large corridor (see Figure \ref{fig:corridor}), were selected. Sonar and
position data were collected from real navigation of Sancho-2 around these
environments. Periodically, the robot was stopped to manually measure its pose
(position and orientation). These recorded data were post-processed to obtain
reduced data sets with controlled position ground truth. All these
experimentation steps are detailed as follows.

The first selected place (see Figure \ref{fig:room4103}) was
a 25 square meters office, reduced to a 2 by 2.5 meters navigation area due to
the furniture. The obstacles, excepting the upper line, were made of polished
surfaces susceptible to generate rebounds. The second place was a hall (see
Figure \ref{fig:hall}) of about 6 by 8 meters in its central zone. There was no furniture, walls were covered by smooth plastic panels, and people passed and
sometimes stopped to see the robot movements.
The third place was a large corridor which can be seen in Figure
\ref{fig:corridor}. It had an approximate size of 10 by 30 meters, the walls
were covered by smooth plastic panels, it included several doors (which
remained closed during the experiment) and several benches and columns on the
right side. These three places showed many of the problems that we wanted to deal
with, and are representative of the places that a mobile robot may find in
indoor environments.

The robot trajectories were programmed to visit the main zones of each place
in order to bring the robot near walls and corners, for perceiving them from
close and from far. In the office the task was to visit sequentially a list of
predefined positions. In the hall a first trajectory visited the four extremes
of an imaginary rectangle through its diagonals, describing an 8-like figure
over the floor. In the second group of trajectories, the task was to visit a
random position and return to the starting point. In this case the movement of
people was not restricted, so in the perception data unexpected obstacles
appeared sometimes. During navigation, the robot estimated its pose using only
odometry.

In every experiment, the robot stopped periodically to allow the human
operator to manually measure the robot pose. The recorded position by the
robot was an odometric estimation, periodically corrected by the introduction
of the manual measurement. Every advanced 50 cm, sonar sensors were fired, and
its readings logged, so each recorded position $(x, y,\alpha)$ was associated
with twelve sonar readings $(d1, d2, ..., d12)$. The resulting data set,
called raw trace, is a chronologically ordered set of vectors:
$$(x, y, \alpha, d1, d2, ..., d12)$$

This trace contained some positions with an estimation error too large for
mapping or localization purposes. These positions were those recorded long
after a manual measure was introduced, and its accumulated odometric error was
not tolerable. Therefore, the raw trace was post-processed to obtain a trace
with controlled position error, called controlled trace. When a manual
measurement was introduced the position estimation error of the last odometric
position was calculated. The post-process assumed a linear increase of the
odometric error, starting at 1 cm, 1$^o$ in every manual measurement, and
ending at the calculated estimation error in the next manual measurement. So
an estimation of error was computed for each position of the raw trace, and
positions with error greater than 25 cm, 15$^o$ were rejected.

\subsection{Mapping Results}

Results for the first place, the office, have been explained in Section
\ref{sec:approximate_maps}, so here we present the results for the other
places: the hall and the corridor.

\begin{figure}[!htb]
\begin{minipage}{.47\linewidth}
\centering\includegraphics[width=\linewidth]{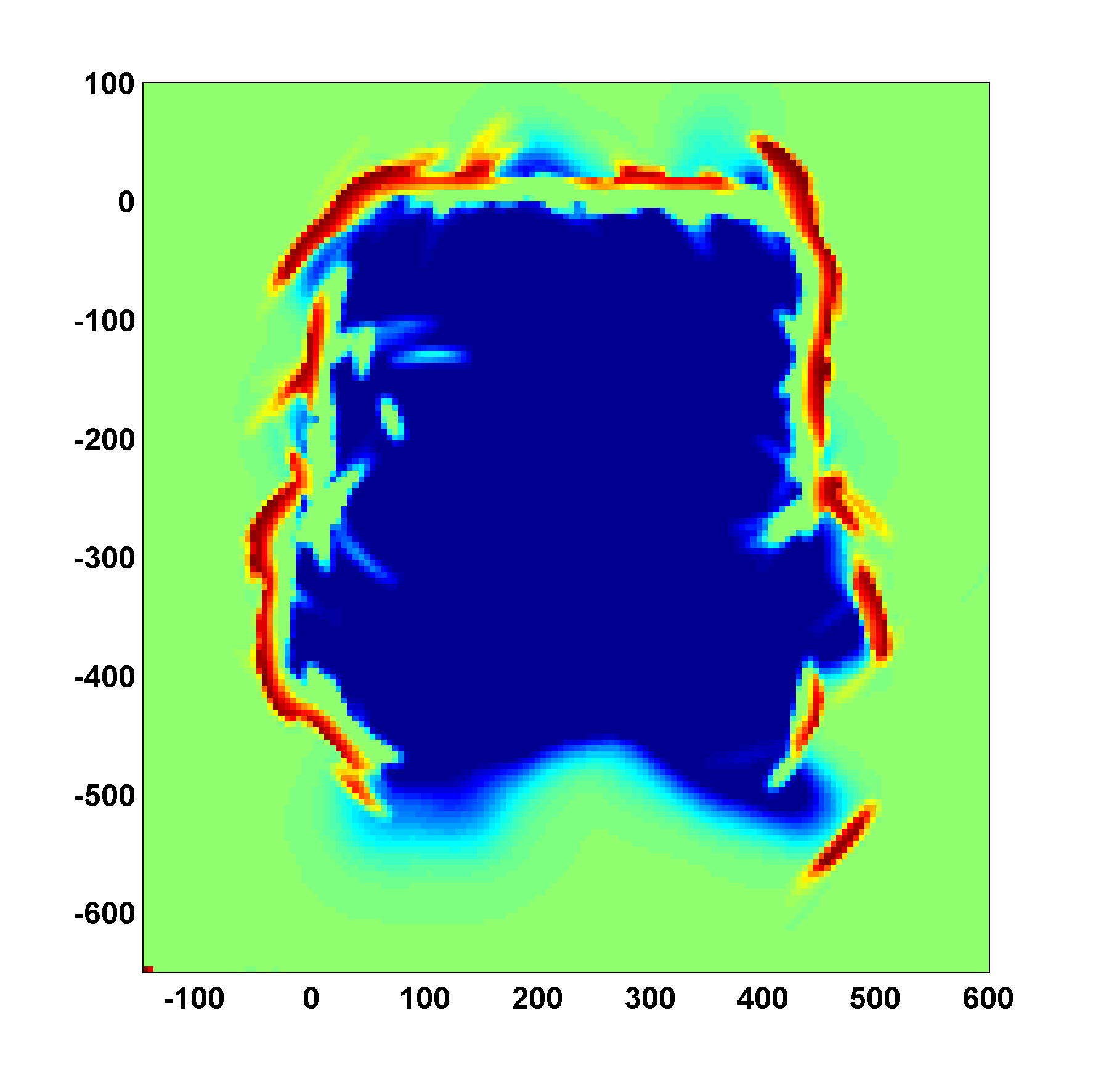}
\caption{Initial integrated map of the hall}\label{fig:fused1-hall}
\end{minipage}
\begin{minipage}{.47\linewidth}
\centering\includegraphics[width=\linewidth]{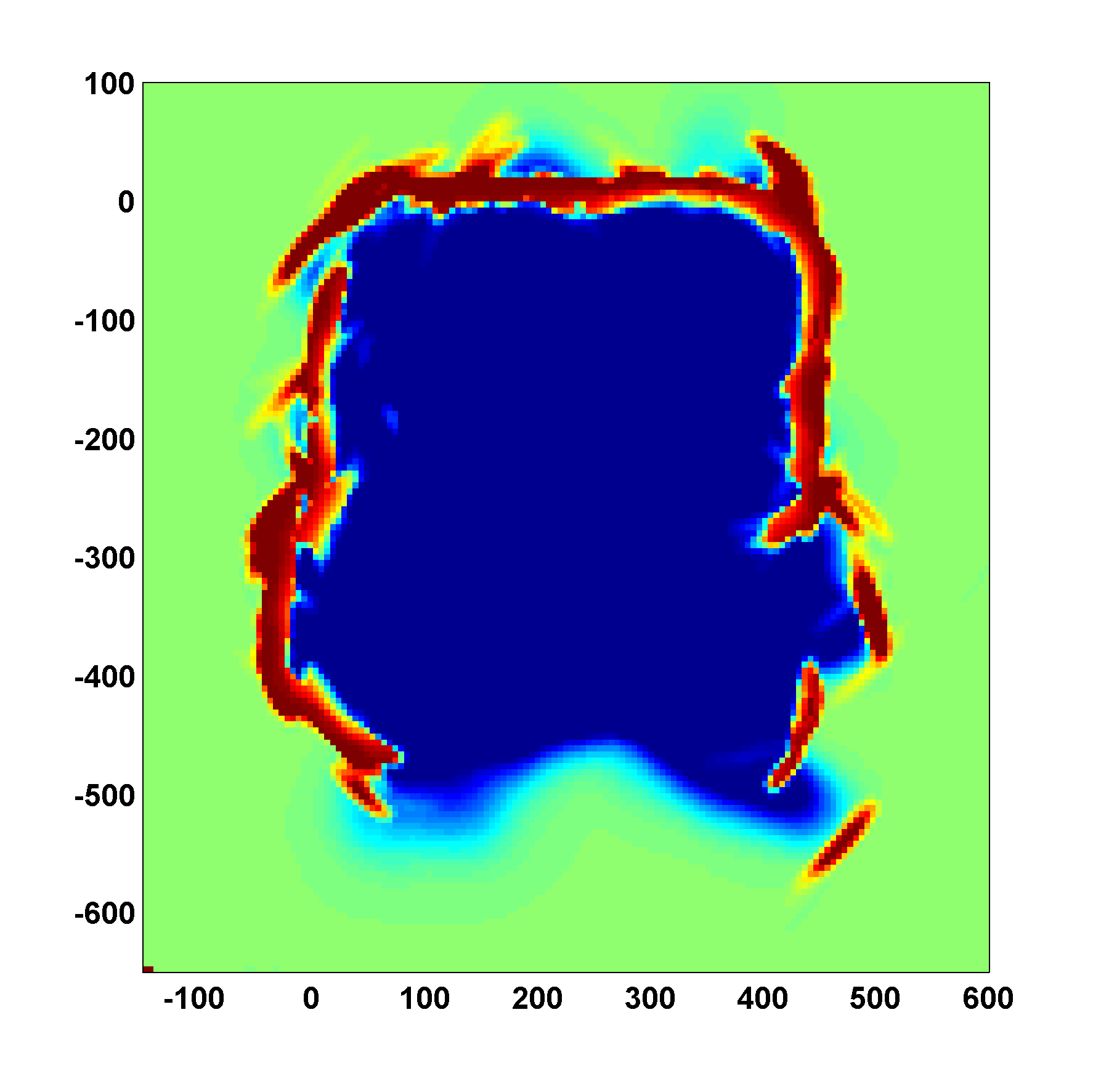}
\caption{Final integrated map of the hall}\label{fig:fused2-hall}
\end{minipage}
\end{figure}

The initial integrated map is shown in Figure \ref{fig:fused1-hall}, where the
false obstacles  at the upper side (the small green areas) correspond to
people. A clear improvement can be seen in the final integrated map in Figure
\ref{fig:fused2-hall}, when contradictions were detected and corrected. Note
that there are fewer outliers, more empty space is detected, and it is nearer
to the walls. When the rebounds were discarded, the locations of the walls
could be appreciated.

\begin{figure}[!htb]
\begin{minipage}{.47\linewidth}
\centering\includegraphics[width=\linewidth]{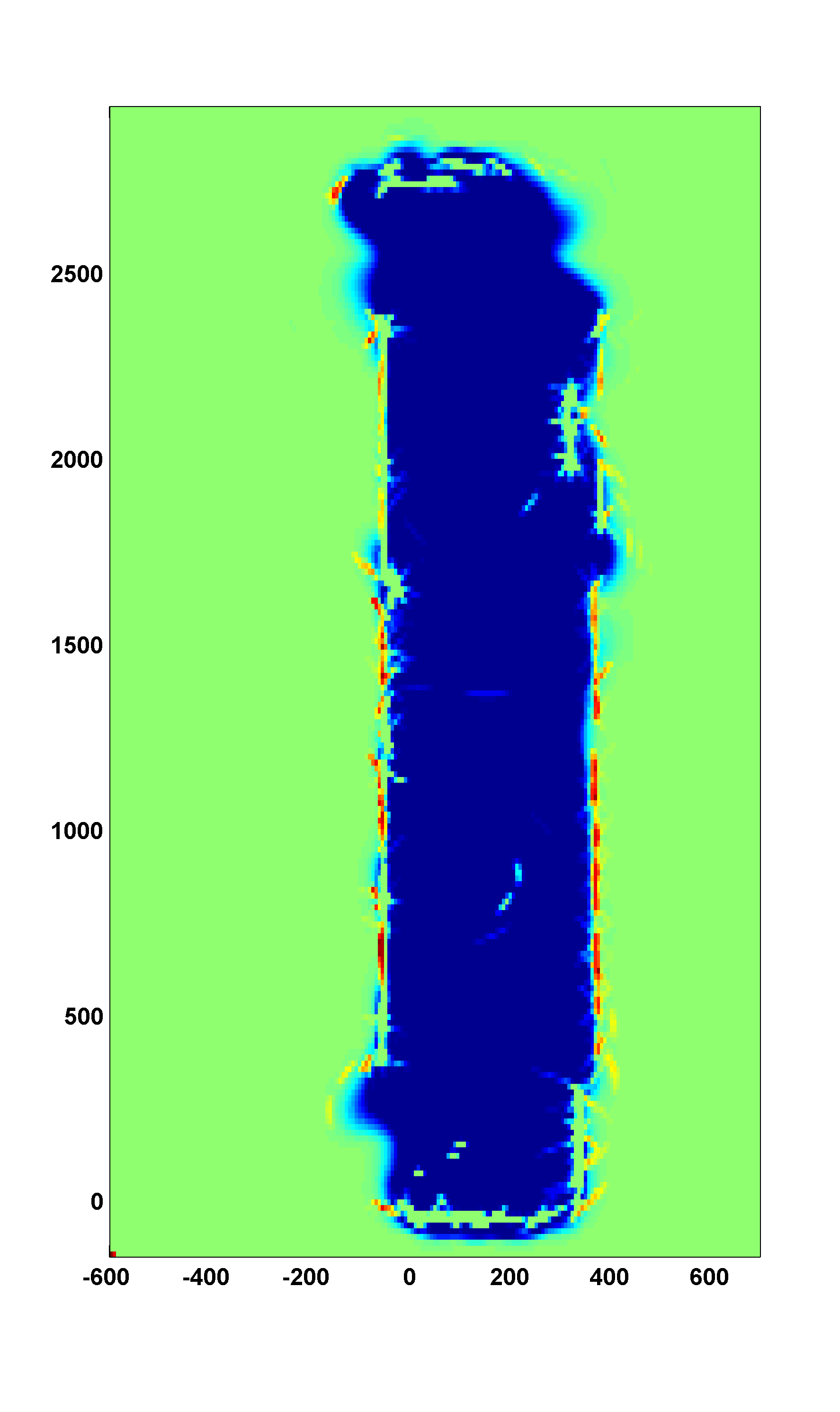}
\caption{Initial Integrated Map of the Corridor}\label{fig:fused1-corridor}
\end{minipage}
\begin{minipage}{.47\linewidth}
\centering\includegraphics[width=\linewidth]{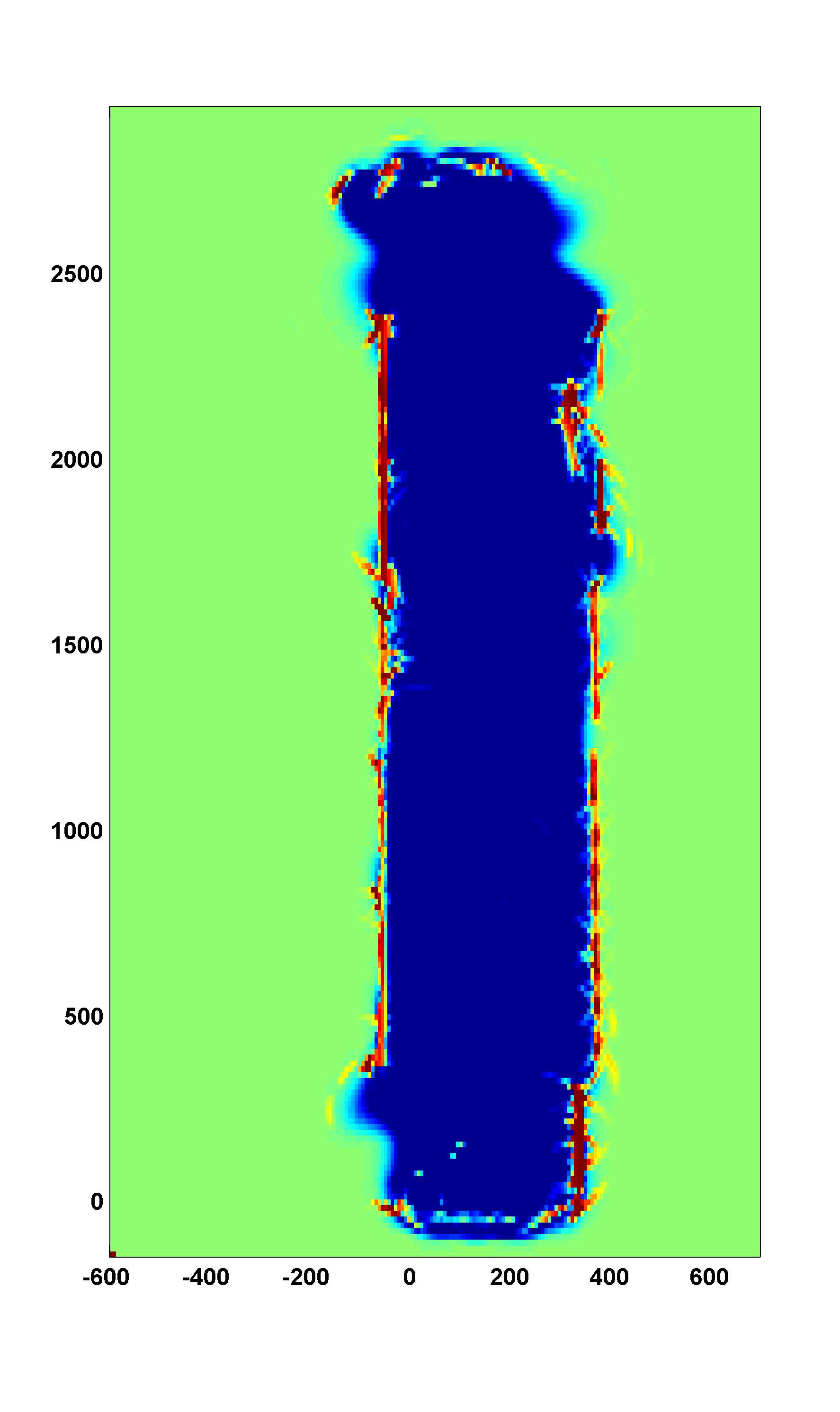}
\caption{Final Integrated Map of the Corridor}\label{fig:fused2-corridor}
\end{minipage}
\end{figure}

A similar effect can be seen in Figures \ref{fig:fused1-corridor} and
\ref{fig:fused2-corridor}, where initially most of the obstacles were
considered unknown but later were recovered and shown in the final integrated
map.

Looking at Figure \ref{fig:fused2-hall} a good map of the hall can be seen,
containing few rebounds and short-echoes. However, looking at Figure
\ref{fig:fused2-corridor} it is harder to see some of the walls of the
corridor although the empty-space represents the place very well.

\subsection{Comparison with other approaches}
\label{sec:comparison}

In order to compare this approach with previous approaches, we ran the
algorithms compared in
\cite{RiboPinz:2001:A_comparison_of_three_uncertainty_calculi_for_building_sonar-based_occupancy_grids}:
a probabilistic approach based on
\cite{Elfes:1987:Sonar_based_real_world_mapping_and_navigation} and a fuzzy
approach  based on
\cite{OrioloUliviVendittelli:1997:Fuzzy_maps:_A_new_tool_for_mobile_robot_perception_and_planning}.
We used the  parameters' values used in the corresponding papers: $\rho_v =
1.2 m$, $\delta_r = 0.15m$, $p_O = 0.6$ for the probability approach, and
$k_E=0.45$ and $k_O=0.65$ for the fuzzy approach.

In Appendix A, the complete set of maps obtained by the three approaches are
presented (see Figures \ref{fig:prob-office} to \ref{fig:fuzz-corridor}). It
can be seen that, due to rebounds and short-echoes, many obstacles (i.e.
walls) are missing in the probabilistic and previous fuzzy methods.

\begin{figure}[htb]
\begin{minipage}{.47\linewidth}
\centering\epsfig{file=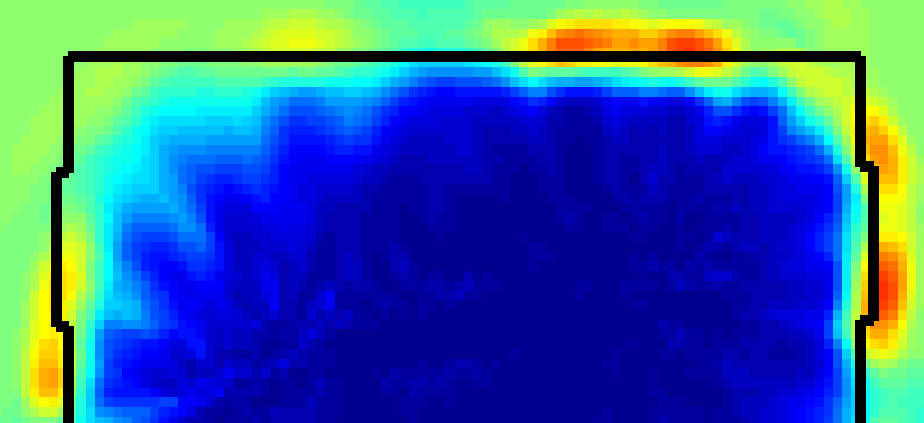,width=\linewidth} \caption{Zoom into the
Probabilistic map} \label{fig:zoom-prob}
\end{minipage}
\begin{minipage}{.47\linewidth}
\centering\epsfig{file=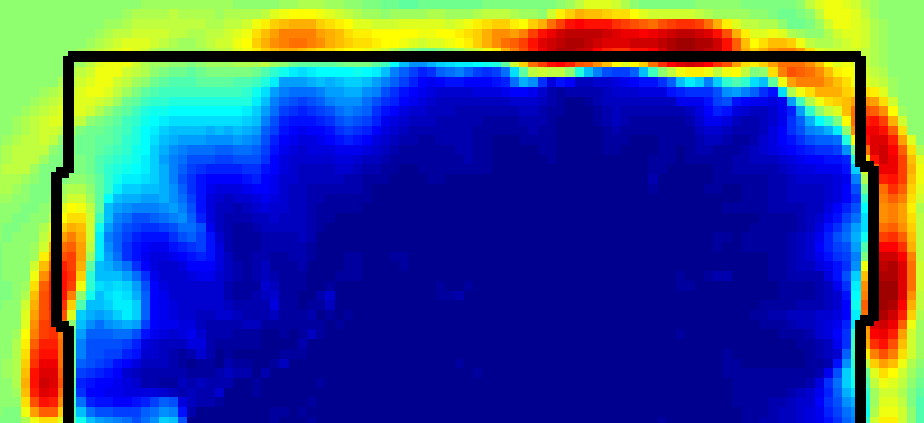,width=\linewidth} \caption{Zoom into the
Fuzzy Map} \label{fig:zoom-poss}
\end{minipage}
\centering\epsfig{file=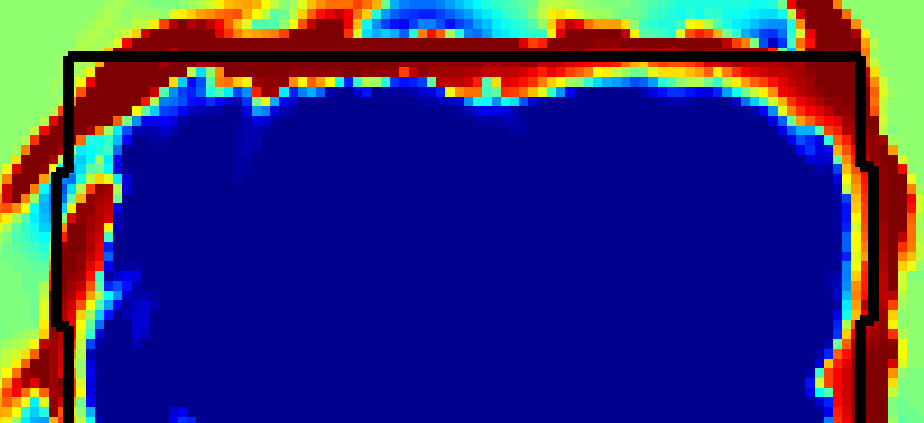,width=.5\linewidth} \caption{Zoom
into the Antonyms-Based Map} \label{fig:zoom-fuzz}
\end{figure}

If one magnifies into the up side of Figure \ref{fig:prob-hall}, it can be
seen that although empty spaces are more or less well captured, most obstacles
are missed by the probabilistic approach (see details in Figure
\ref{fig:zoom-prob}). Something similar occurs to the previous fuzzy method
which can be seen in figure \ref{fig:zoom-poss}. Nevertheless, we can see by
magnifying Figure \ref{fig:fuzz-hall} that, although obstacles are in some
zones imprecise, the map built by the antonyms-based method follows the wall
very well (see details in Figure \ref{fig:zoom-fuzz}).

In the case of the fuzzy approach, we can also compare obstacles maps and empty space
maps directly, and compare the integrated map obtained using the antonyms-based approach with the safe map obtained using the fuzzy approach proposed in \cite{RiboPinz:2001:A_comparison_of_three_uncertainty_calculi_for_building_sonar-based_occupancy_grids}, a map that directly excludes contradictory cells and indeterminate cells.

It can be seen in this comparison with the other approaches that the
antonyms-based method has created better maps with more defined walls and
clear empty-spaces, in spite of imprecision inherent to the sensors. If
necessary, we could introduce thresholds to decide more precisely the limits
of obstacles and of empty spaces, but in that case we would be forcing the
model by introducing more constraints. However, antonyms-based maps recall
most of the obstacles and empty spaces (see quantitative comparison bellow for
details), since with this approach, rebounds and short echoes can be detected,
differentiated, and removed from the integrated map.

\subsection{Quantitative comparison}

In order to perform a quantitative comparison, we needed to define a set of
measures that allowed us to compare the results of the different solutions.
First, we defined a reference occupancy map obtained from the manual measurement
of the walls' positions. In the reference map obstacles, empty-spaces and unknown
cells were represented by 1, -1 and 0 values respectively (as a ternary
classification problem). Then, we quantified the accuracy of the maps, estimated
by comparing the resulting map of the three approaches with the reference map, by means
of the precision, recall, f-measure and mean absolute error (MAE) measures.

To make the comparison and discretization easier we  re-scaled all the
obtained maps to [-1,1] with -1 meaning completely empty, 1 meaning completely
occupied and 0 meaning unknown.

We discretized the integrated maps obtained by assigning to obstacles those
cells that had a value between $\alpha$ and 1, assigning to unknowns those
cells that had a value between $-\alpha$ and $+\alpha$ and assigning to
empty-spaces those cells that had a value between $-1$ and $-\alpha$. That
means that a cell needed to have a degree greater than $\alpha$ to be
considered an obstacle (this resembles an $\alpha$-cut of the fuzzy set of
obstacles), smaller than $-\alpha$ to be considered an empty-space (this
resembles an $\alpha$-cut of the fuzzy set of empty-spaces), and considered
unknown in other cases. From that we built the following confusion matrix:.

\begin{center}
\begin{tabular}{|rl|c|c|c|}
\cline{1-5}& & \multicolumn{3}{|c|}{Actual}\\
\cline{3-5} & & Obstacles & Empty & Unknown\\
\cline{1-5} \multirow{3}{*}{Predicted} \vline & Obstacles & oto & efo & ufo\\
\cline{2-5} \vline & Empty & ofe & ete & ufe\\
\cline{2-5} \vline & Unknow & ofu & efu & utu\\\cline{1-5}
\end{tabular}
\end{center}

\noindent where $oto$ stands for Obstacle-True-Obstacle, which means that the
method predicted an obstacle when it was obstacle, $efo$ stands for
Empty-False-Obstacle, which means that the method predicted an empty-space
when it was an obstacle, $ofe$ stands for Obstacle-False-Empty, which means
that the method predicted an obstacle when it was an empty-space, and so on.

From the confusion matrix, Precision ($P$) and Recall ($R$) for obstacles and
empty-spaces are defined as:
\begin{align}
\mbox{P}_{O}=\frac{oto}{oto+efo+ufo}\hspace{0.5cm}
\mbox{R}_{O}=\frac{oto}{oto+ofe+ofu}\hspace{0.5cm}\\
\mbox{P}_{E}=\frac{ete}{ete+ofe+ufe}\hspace{0.5cm}
\mbox{R}_{E}=\frac{ete}{efo+ete+efu}\hspace{0.5cm}
\end{align}

We didn't calculate the precision and recall of unknown space since it was
complementary to the other two and didn't add further information.

From precision and recall the f-measure is defined using the weighted
harmonic mean
\begin{equation}
F_\beta= \frac{(1 + \beta)}{ \frac{1}{\mathrm{Precision}} +
\frac{\beta}{\mathrm{Recall}}}\,
\end{equation}

In this case, since recall is more important than precision (to avoid collisions with obstacles), we  used the $F_2$ measure, whose weights recall twice as much as precision.

To aggregate and obtain a Total Combined Rate (TCR) from the results of
obstacles and empty-spaces, we used the arithmetic mean of f-measures from
the obstacles $F_{O}$ and from the empty-spaces $F_{E}$ (the results can be
see in tables \ref{tab:pre_rec_small}, \ref{tab:pre_rec_hall} and
\ref{tab:pre_rec_corridor}).

\begin{equation}
TCR = \frac{F_{O}+F_{E}}{2}
\end{equation}

Different elections of the confidence threshold $\alpha$ returned different
solutions (see Figures \ref{fig:tcr_office} to \ref{fig:tcr_corridor}), but to
show a detailed comparison, a reasonable choice was to split the range $[-1,1]$
in three parts $([-1,-\alpha],[-\alpha,\alpha],[\alpha,1])$ and take $\alpha = \frac{1}{3}$. The numerical results of this
choice can be seen in the tables \ref{tab:pre_rec_small},
\ref{tab:pre_rec_hall} and \ref{tab:pre_rec_corridor}.

To perform an exhaustive comparison, we calculated the Total Combined Rate
(TCR) for 30 different $\alpha\in(0,1)$, for each method and for each place.
The results of these comparisions can be seen in Figures \ref{fig:tcr_office},
\ref{fig:tcr_hall}, and \ref{fig:tcr_corridor}.

\begin{table}[!htb]
\begin{center}
\caption{Precision, Recall and $F_2$  measures for the obstacles and
empty-spaces maps of the office, and the Total Combined Rate (TCR) for each
method}\label{tab:pre_rec_small}
\begin{tabular}{|c|c|c|c|c|c|c|c|c|}
  \hline\hline
  Method & $P_{O}$ & $R_{O}$ & $F_{O}$ & $P_{E}$ & $R_{E}$ & $F_{E}$ & TCR \\\hline
  Probabilistic & \textbf{40}\%  &  25\% &  {29}\% & 99\% & {55}\% & {64}\% & {47}\% \\\hline
  Fuzzy &  30\%  &  {50}\% &  \textbf{41}\% & \textbf{99}\% & 56\% & 65\% &  \textbf{53}\%  \\\hline
  \textbf{Antonyms} & 22\%  &  \textbf{54}\% &  36\% & 91\% & \textbf{62}\% & \textbf{69}\% & \textbf{53}\% \\\hline
  \hline
\end{tabular}
\end{center}
\end{table}

\begin{table}[!htb]
\begin{center}
\caption{Precision, Recall and $F_2$  measures for the obstacles and
empty-spaces maps of the hall, and the Total Combined Rate (TCR) for each
method}\label{tab:pre_rec_hall}
\begin{tabular}{|c|c|c|c|c|c|c|c|}
  \hline\hline
  Method & $P_{O}$ & $R_{O}$ & $F_{O}$ & $P_{E}$ & $R_{E}$ & $F_{E}$ & TCR \\\hline
  Probabilistic & \textbf{52}\%  &  14\% &  {18}\% & \textbf{99}\% & {77}\% & {83}\% & {51}\% \\\hline
  Fuzzy &  39\%  &  {39}\% &  39\% & \textbf{99}\% & {78}\% & {84}\% & 61\%  \\\hline
  \textbf{Antonyms} & {33}\% & \textbf{75}\% & \textbf{53}\% & 97\% & \textbf{80}\% & \textbf{85}\% & \textbf{69}\% \\\hline
  \hline
\end{tabular}
\end{center}
\end{table}

\begin{table}[!htb]
\begin{center}
\caption{Precision, Recall and $F_2$  measures for the obstacles and
empty-spaces maps of the corridor, and the Total Combined Rate (TCR) for each
method}\label{tab:pre_rec_corridor}
\begin{tabular}{|c|c|c|c|c|c|c|c|}
  \hline\hline
  Method & $P_{O}$ & $R_{O}$ & $F_{O}$ & $P_{E}$ & $R_{E}$ & $F_{E}$ & TCR \\\hline
  Probabilistic & \textbf{78}\%  & 2\% &  {3}\% & \textbf{99}\% & {75}\% & {81}\% & {42}\% \\\hline
  Fuzzy &  43\%  &  {8}\% &  10\% & \textbf{98}\% & 74\% & {81}\% & 46\%  \\\hline
  \textbf{Antonyms} & {40}\%  &  \textbf{39}\% &  \textbf{40}\% & {94}\% & \textbf{91}\% & \textbf{92}\% & \textbf{66}\% \\\hline
  \hline
\end{tabular}
\end{center}
\end{table}

To obtain a more global evaluation of the obtained maps $Obt$ we also
calculated the Mean Absolute Error (MAE) with respect to the reference map
$Ref$ (the results can be seen in table \ref{tab:mae}).

\begin{equation}
MAE = \frac{\sum_{i=1}^n\sum_{j=1}^m|Ref(C_{ij})-Obt(C_{ij})|}{n\cdot m}
\end{equation}

\begin{table}[!htb]
\begin{center}
\caption{Mean Absolute Error (MAE) for each method}\label{tab:mae}
\begin{tabular}{|c|c|c|c|}
  \hline\hline
  & \multicolumn{3}{|c|}{MAE} \\\hline
  Method & Office & Hall & Corridor \\\hline
  Probabilistic & 0.2115 & 0.1985 & 0.1512 \\\hline
  Fuzzy &  0.2184 &  0.1848& 0.1400 \\\hline
  \textbf{Antonyms} & \textbf{0.1688} &  \textbf{0.1395} & \textbf{0.0767} \\\hline
  \hline
\end{tabular}
\end{center}
\end{table}

\subsection{Discussion}

As it can be seen in Appendix A the maps obtained by the antonyms-based method
are better defined, capture better the shape of the walls and of the
empty-spaces, and contain less errors due to rebounds and short-echoes. While
in the probabilistic and in the previous fuzzy method many obstacles (i.e.
walls) are missing, and many empty-spaces not clear enough. The quantitative
measures of performance (see Tables \ref{tab:pre_rec_small} to \ref{tab:mae})
of the three different methods in the three places come to confirm these
qualitative evaluation.

For instance, in Tables \ref{tab:pre_rec_small}, \ref{tab:pre_rec_hall} and
\ref{tab:pre_rec_corridor} it can be seen that although the probabilistic
method has the highest precision it has the lowest recall; while the
antonyms-based method has the highest recall, and the best balance between
precision and recall. This means that antonyms-based method
detects more obstacles and more empty spaces, and have a sound compromise between
precision and recall.

In addition, the antonyms-based method has the highest f-measure $F_2$ and the
highest Total Combined Rate $TCR$, which means that more obstacles and
empty-spaces are captured, while less empty-spaces are considered as obstacles
and less obstacles are considered as empty-spaces.

\begin{figure}[!htb]
\centering\includegraphics[width=.65\linewidth]{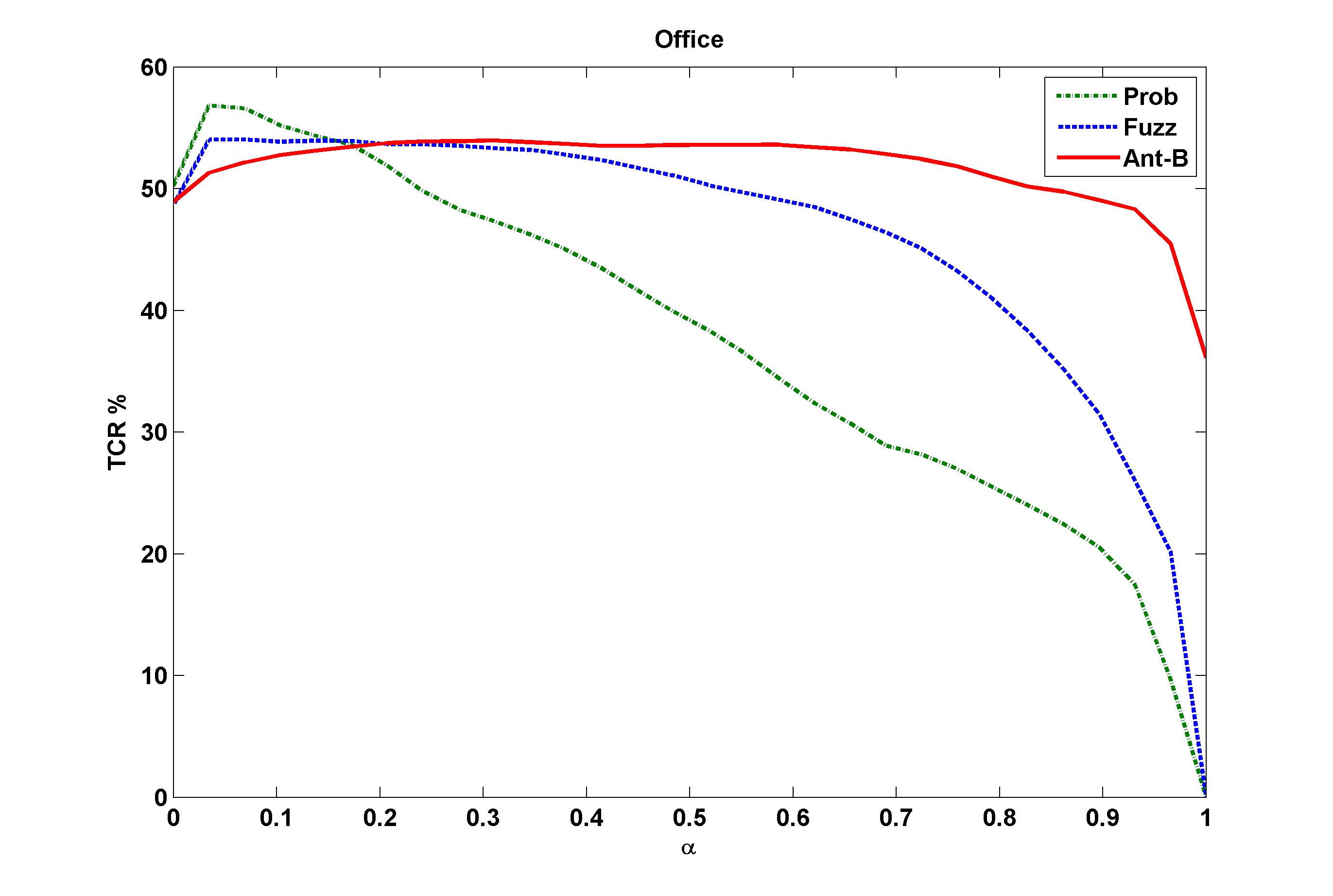}
\caption{Comparison of TCR for different $\alpha$ for the
office}\label{fig:tcr_office}
\end{figure}

\begin{figure}[!htb]
\centering\includegraphics[width=.65\linewidth]{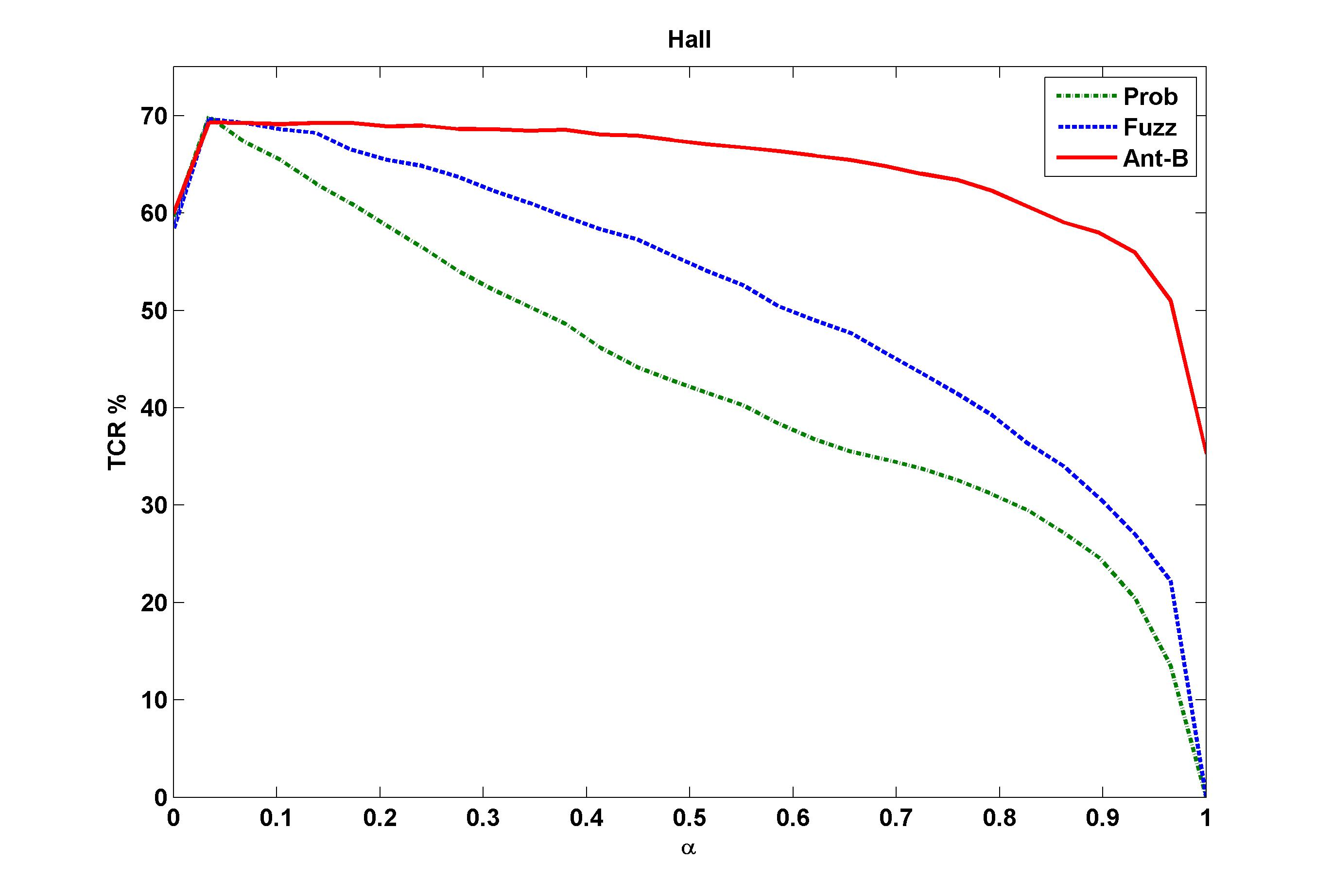}
\caption{Comparison of TCR for different $\alpha$ for the
hall}\label{fig:tcr_hall}
\end{figure}

\begin{figure}[!htb]
\centering\includegraphics[width=.65\linewidth]{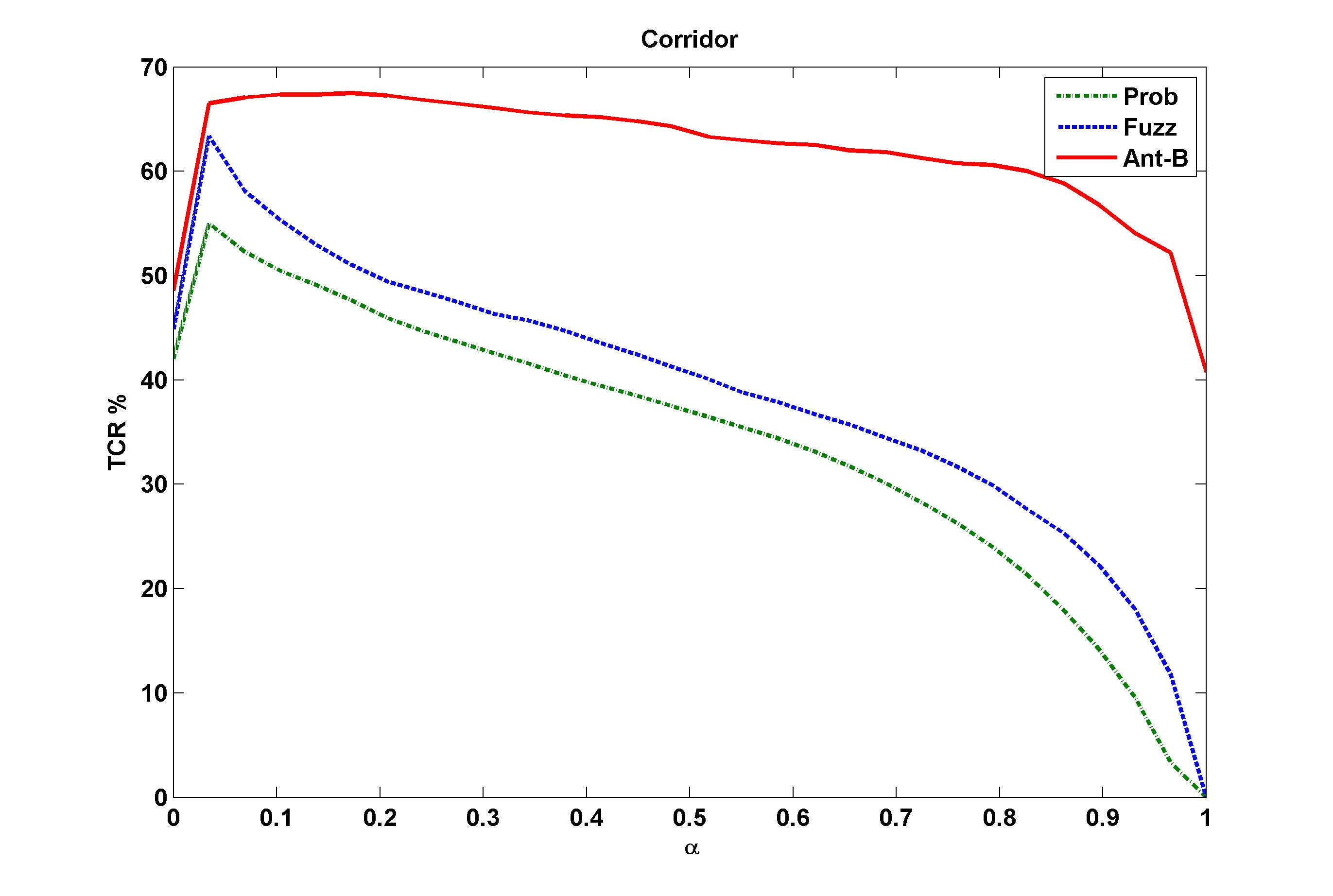}
\caption{Comparison of TCR for different $\alpha$ for the
corridor}\label{fig:tcr_corridor}
\end{figure}

Looking at Figures \ref{fig:tcr_office}, \ref{fig:tcr_hall} and
\ref{fig:tcr_corridor}, it can be seen that antonyms-based method obtains
higher and more stable $TCR$ for different thresholds $\alpha$, while the
others degrade faster when confidence threshold $\alpha$ grows. This means
that antonyms-based method obtains better maps with higher confidences, and
therefore is more robust to noise and to imprecision of sonar-sensors.

When the environment becomes bigger, precision of obstacles and empty-spaces
increases while recall of obstacles decreases and recall of empty-spaces
increases. This is due to the fact that there are many more empty-spaces than
obstacles and also that empty-spaces are easier to  identify than obstacles.
One question that goes beyond the scope of this work is how these methods could
be applied to a dynamic environments where many obstacles are moved, removed
or added (see \cite{Bibby-RSS-07,diaz2009fuzzy} for solutions to this
problem).

For a more global comparison we also calculated the Mean Absolute Error (MAE),
obtaining as result that antonyms-based method has the smallest MAE in the
three environments, as can be seen in table \ref{tab:mae}, and therefore it can
be considered more accurate than the other methods, since the map obtained is
closer to the reference map.



\section{Conclusions}
\label{sec:conclusion}

In this paper we have proposed new ways of representing the perceptions that
an Autonomous Mobile Robot can obtain of its environment, emphasizing that
environment and sensor characteristics must be considered as a whole. The aim
has been to model these perceptions properly by taking into account the
imprecision inherent in the environment and the robot sensors. For that
purpose we have built a robust model based on antonyms that can properly
handle the imprecision and contradictions that arise in the process of
building these maps, which, in addition, represents a new way of building
antonyms from data.

Fuzzy logic theory has proved to be adequate and useful for representing the
inherent imprecision in this robotic mapping problem. This theory has allowed
us to successfully integrate the information obtained from different
observations, and to build accurate maps for exploration tasks. In particular,
antonyms abstraction has made possible to model and handle the concepts
``occupied space'' and ``empty space'' when reasoning about the integration of
past and actual perceptions during the navigation of a mobile robot. From that
we have built a set of fuzzy grid maps that represent the robot's perception of
``obstacles'', of ``empty spaces'', of ``short-echoes'', and of ``rebounds'',
respectively. These maps were later fused to obtain an integrated map.

A controlled experimentation with a real robot equipped with sonar
exteroception was performed in three representative real indoor places.
Obtained maps show high fidelity with respect to the real architectural walls
map, demonstrating the suitability and robustness of this approach to robotic
mapping. Furthermore, it should be noted that these results were obtained
using low cost sensors of limited accuracy.

Based on the qualitative and quantitative comparison performed, we can
conclude that the antonyms-based method performs better than the probabilistic
method and the previous fuzzy method, obtaining a better recall of obstacles
and empty-spaces, a good balance between precision and recall, a higher $TCR$
and a smaller $MAE$ in the three experiments performed.

The inclusion of antonyms also allowed us to discard rebounds and short-echoes
and reduce greatly the contradictions and errors. We dealt with rebounds,
short echoes, and other noises by defining a set of fuzzy rules that capture
our knowledge about the problem. Also, the way of building the aggregated maps
by means of linguistic quantifiers differs greatly from previous approaches,
and it allowed us to handle the partial contribution of each sonar reading to
the aggregated map.

Thanks to maintaining obstacles and empty maps as a pair of antonyms, we were
able to deal with contradictions, and build an accurate and robust integrated
map. The use of approximate maps allowed us to synthesize the accumulated
information from samples in a way that kept the data structure constant while
the accuracy of the representation achieved increased with the number of
samples.

An important point is that the obtained integrated map can be traced back to
the contradictions map and from it to the obstacles and empty spaces maps,
adding an explicative feature to the results. Also, by dealing explicitly with
contradictions, we were able to distinguish between two kinds of unknown cells,
the ones that are unknown due to contradictions and the ones that are unknown
because are unexplored. This allowed to the robot to recognize which zones needed
to be navigated with care and which zones needed to be explored later on.

The main difference of this contribution from other published works is that
our approach is based on the manipulation of perceptions by means of antonyms.
In addition, the obtained results are more accurate, more robust and more
understandable than the previous ones.

The approach to model perceptions by means of linguistic descriptions and
antonyms, and its applicability to real problems can be seen as an initial
step in the development of Computing with Words and Computational Theory of Perceptions proposed by Zadeh in \cite{Zadeh:2001:A_new_direction_in_AI,Zadeh:2006:A_new_frontier_in_computation,Zadeh:2008:Is_there_a_need_for_fuzzy_logic,Zadeh:2008:Toward_human_level_machine_intelligence}.

\subsection*{Acknowledgments}
The authors want to thank Claudio Moraga, Enric Trillas and David P\'{e}rez for
their comments, corrections and fruitful discussion about the topic of the
paper. Also, the authors want to thank the reviewers and editor for their helpful comments to improve the quality of the paper.





\bibliography{E:/pubs/bib/xbib_paper,E:/pubs/bib/xbib_proc,E:/pubs/bib/xbib_book,E:/pubs/bib/xbib_otros}

\begin{thebibliography}{62}
\expandafter\ifx\csname natexlab\endcsname\relax\def\natexlab#1{#1}\fi
\expandafter\ifx\csname url\endcsname\relax
  \def\url#1{\texttt{#1}}\fi
\expandafter\ifx\csname urlprefix\endcsname\relax\def\urlprefix{URL }\fi

\bibitem[{Aguirre and
  Gonz{\'a}lez(2003)}]{AguirreGonzalez:2003:A_Fuzzy_Perceptual_Model_for_Ultra%
_sound_Sensors_Applied_to_Intelligent_Navigation_of_Mobile_Robots}
Aguirre, E., Gonz{\'a}lez, A., 2003. A fuzzy perceptual model for ultrasound
  sensors applied to intelligent navigation of mobile robots. Applied
  Intelligence 19~(3), 171--187.

\bibitem[{Alsina et~al.(1983)Alsina, Trillas, and
  Valverde}]{AlsTriVal:1983:On_some_logical_connectives_for_fuzzy_sets_theorie%
s}
Alsina, C., Trillas, E., Valverde, L., 1983. On some logical connectives for
  fuzzy sets theories. J. of Math. Anal. \& Appl. 93, 15--26.

\bibitem[{Bailey and Durrant-Whyte(2006)}]{bailey2006simultaneous}
Bailey, T., Durrant-Whyte, H., 2006. {Simultaneous localization and mapping
  (SLAM): part II}. IEEE Robotics \& Automation Magazine 13~(3), 108--117.

\bibitem[{Bibby and Reid(2007)}]{Bibby-RSS-07}
Bibby, C., Reid, I., 2007. Simultaneous localisation and mapping in dynamic
  environments {(SLAMIDE)} with reversible data association. In: Proceedings of
  Robotics: Science and Systems. Atlanta, GA, USA, pp. 105--112.

\bibitem[{Borenstein et~al.(1996)Borenstein, Everett, and
  Feng}]{BorensteinEverett:96:Navigating_Mobile_Robots}
Borenstein, J., Everett, H., Feng, L., 1996. Navigating Mobile Robots. AK
  Peters, New York.

\bibitem[{Cao and
  Borenstein(2002)}]{CaoBorenstein:2002:Experimental_Characterization_of_Polar%
oid_Ultrasonic_Sensors}
Cao, A., Borenstein, J., 2002. {Experimental Characterization of Polaroid
  Ultrasonic Sensors in Single and Phased Array Configuration}. In: UGV
  Technology Conference at the 2002 SPIE AeroSense Symposium. Orlando, FL, pp.
  95--99.

\bibitem[{Chen et~al.(2009)Chen, Li, and Yeh}]{chen2009ep}
Chen, C., Li, T., Yeh, Y., 2009. {EP-based kinematic control and adaptive fuzzy
  sliding-mode dynamic control for wheeled mobile robots}. Information Sciences
  179~(1-2), 180--195.

\bibitem[{Chow et~al.(2002)Chow, Rad, and
  Ip}]{ChowRadIP:2002:Enhancement_of_Probabilistic_Grid-based_Map_for_Mobile_R%
obot_Applications}
Chow, K., Rad, A., Ip, Y., 2002. {Enhancement of Probabilistic Grid-based Map
  for Mobile Robot Applications}. Journal of Intelligent and Robotic Systems
  34~(2), 155--174.

\bibitem[{Cohen et~al.(2006)Cohen, Edan, and
  Schechtman}]{Cohen:2006:Statistical_evaluation_method}
Cohen, O., Edan, Y., Schechtman, E., 2006. {Statistical evaluation method for
  comparing grid map based sensor fusion algorithms}. The International Journal
  of Robotics Research 25~(2), 117--133.

\bibitem[{Cruse(2000)}]{Cruse:2000:Meaning_in_Language}
Cruse, A., 2000. Meaning in Language: An Introduction to Semantics and
  Pragmatics. Oxford University Press.

\bibitem[{{de Soto} and
  Trillas(1999)}]{DeSotoTrillas:1999:On_Antonym_and_negate_in_fuzzy_logic}
{de Soto}, A.~R., Trillas, E., 1999. On antonym and negate in fuzzy logic. Int.
  Jour. of Intelligent Systems 14, 295--303.

\bibitem[{Diaz-Hermida et~al.(2009)Diaz-Hermida, Bugarin, Carinena, Mucientes,
  and Losada}]{diaz2009fuzzy}
Diaz-Hermida, F., Bugarin, A., Carinena, P., Mucientes, M., Losada, D., 2009.
  {Fuzzy quantification in two real scenarios: Information retrieval and mobile
  robotics}. International Journal of Intelligent Systems 24~(6), 572--586.

\bibitem[{Dudek and
  Jenkin(2000)}]{DudekJenkin:2000:Computational_Principles_of_Mobile_Robotics}
Dudek, G., Jenkin, M., 2000. Computational Principles of Mobile Robotics.
  Cambridge University Press.

\bibitem[{Durrant-Whyte and Bailey(2006)}]{durrant2006simultaneous}
Durrant-Whyte, H., Bailey, T., 2006. {Simultaneous localization and mapping:
  part I}. IEEE Robotics \& Automation Magazine 13~(2), 99--110.

\bibitem[{Elfes(1987)}]{Elfes:1987:Sonar_based_real_world_mapping_and_navigati%
on}
Elfes, A., 1987. Sonar based real world mapping and navigation. IEEE Journal of
  robotics and automation 3~(3), 249--265.

\bibitem[{Elfes(1989)}]{elfes:1989:Using_Occupancy_Grids_for_Mobile_Robot}
Elfes, A., 1989. {Using occupancy grids for mobile robot perception and
  navigation}. Computer 22~(6), 46--57.

\bibitem[{Eustice et~al.(2006)Eustice, Singh, Leonard, and
  Walter}]{Eustice:2006:Visually_mapping}
Eustice, R., Singh, H., Leonard, J., Walter, M., 2006. {Visually mapping the
  RMS Titanic: Conservative covariance estimates for SLAM information filters}.
  The International Journal of Robotics Research 25~(12), 1223--1242.

\bibitem[{Fazli and Kleeman(2006)}]{fazli2006simultaneous}
Fazli, S., Kleeman, L., 2006. {Simultaneous landmark classification,
  localization and map building for an advanced sonar ring}. Robotica 25~(3),
  283--296.

\bibitem[{Fox et~al.(1999)Fox, Burgard, Dellaert, and
  Thrun}]{Foxetal:99:Montecarlo_Localization}
Fox, D., Burgard, W., Dellaert, F., Thrun, S., 1999. Montecarlo localization:
  Efficient position estimation for mobile robots. In: Proc. of the Sixteenth
  National Conference on Artificial Intelligence (AAAI-99). Orlando, pp.
  343--349.

\bibitem[{Gasos and
  Mart{\'\i}n(1997)}]{GasosMartin:1997:Mobile_robot_localization_using_fuzzy_m%
aps}
Gasos, J., Mart{\'\i}n, A., 1997. {Mobile robot localization using fuzzy maps}.
  Fuzzy Logic in Artificial Intelligence 1188, 207--224.

\bibitem[{Grisetti et~al.(2007)Grisetti, Stachniss, Grzonka, and
  Burgard}]{Grisetti-RSS-07}
Grisetti, G., Stachniss, C., Grzonka, S., Burgard, W., 2007. A tree
  parameterization for efficiently computing maximum likelihood maps using
  gradient descent. In: Proceedings of Robotics: Science and Systems. Atlanta,
  GA, USA, pp. 65--72.

\bibitem[{Guadarrama(2007)}]{Gua:2007:A_Contribution_to_Computing_with_Words_P%
erceptions}
Guadarrama, S., April 2007. A contribution to computing with words and
  perceptions. Ph.D. thesis, Technical University of Madrid, Madrid, (Available
  online at http://oa.upm.es/448/).

\bibitem[{Guadarrama et~al.(2004)Guadarrama, Mu{\~n}oz, and
  Vaucheret}]{GuaMunozVaucheret:2004:Fuzzy_Prolog:_a_new_approach_using_soft_c%
onstraints_propagation}
Guadarrama, S., Mu{\~n}oz, S., Vaucheret, C., 2004. Fuzzy prolog: a new
  approach using soft constraints propagation. Fuzzy Sets and Systems 144~(1),
  127--150.

\bibitem[{Guadarrama et~al.(2006)Guadarrama, Renedo, and
  Trillas}]{GuaRenTri:2006:Some_Fuzzy_Counterparts_of_the_Language_uses_of_And%
_and_Or}
Guadarrama, S., Renedo, E., Trillas, E., 2006. {Some Fuzzy Counterparts of the
  Language uses of And and Or}. In: Reusch, B. (Ed.), Computational
  Intelligence: Theory and Practice. Vol. 164 of Studies in fuzziness and soft
  computing. Springer, pp. 335--352.

\bibitem[{Guadarrama et~al.(2002)Guadarrama, Trillas, and
  Renedo}]{GuaTriRen:2002:Non_Contradiction_and_Excluded_Middle_with_antonyms}
Guadarrama, S., Trillas, E., Renedo, E., 2002. Non-contradiction and
  excluded-middle with antonyms. In: Proceeding of ESTYLF'2002. Le{\'o}n, pp.
  385--389.

\bibitem[{Gutmann and Schlegel(1996)}]{GutmannSchlegel:96:AMOS}
Gutmann, J., Schlegel, C., 1996. Amos: Comparison of scan matching approaches
  for self-localization in indoor environments. In: Proc. 1st Euromicro
  Workshop on Advanced Mobile Robots (EUROBOT'96). Kaiserslautern, Germany, pp.
  61--67.

\bibitem[{Karaman and Temelta(2005)}]{karaman2005navigation}
Karaman, O., Temelta, H., 2005. {Navigation of Mobile Robots in Unstructured
  Environment Using Grid Based Fuzzy Maps}. Lecture notes in computer science
  3614, 925.

\bibitem[{Kortenkamp and
  Weymouth(1994)}]{kortenkamp:1994:Topological_Mapping_for_Mobile_Robots}
Kortenkamp, D., Weymouth, T., 1994. {Topological mapping for mobile robots
  using a combination of sonar and vision sensing}. In: Proceedings of the
  twelfth national conference on {A}rtificial {I}ntelligence (vol. 2). American
  Association for Artificial Intelligence Menlo Park, CA, USA, pp. 979--984.

\bibitem[{Lee and
  Chung(2009)}]{LeeChung:2009:Effective_Maximum_Likelihood_Grid_Map}
Lee, K., Chung, W.~K., Aug. 2009. Effective maximum likelihood grid map with
  conflict evaluation filter using sonar sensors. IEEE Transactions on Robotics
  25~(4), 887--901.

\bibitem[{Lehrer(1985)}]{Lehrer:1985:Markedness_and_antonymy}
Lehrer, A., 1985. Markedness and antonymy. Journal of Linguistics 21, 397--429.

\bibitem[{Leonard et~al.(1992)Leonard, Durrant-Whyte, and
  Cox}]{Leonardetal:1992:Dynamic_map_building_for_AMR}
Leonard, J., Durrant-Whyte, H., Cox, I., 1992. {Dynamic Map Building for an
  Autonomous Mobile Robot}. The International Journal of Robotics Research
  11~(4), 286--298.

\bibitem[{Li et~al.(2007)Li, Huang, Dezert, Duan, and
  Wang}]{Li:2007:A_successful}
Li, X., Huang, X., Dezert, J., Duan, L., Wang, M., 2007. {A successful
  application of DSmT in sonar grid map building and comparison with DST-based
  approach}. International Journal of Innovative Computing, Information and
  Control 3~(3), 539--549.

\bibitem[{Li et~al.(2006)Li, Huang, and Wang}]{Li:2006:Robot_map_building}
Li, X., Huang, X., Wang, M., 2006. {Robot map building from sonar and laser
  information using DSmT with discounting theory}. International Journal of
  Information Technology 3~(2), 78--85.

\bibitem[{Lyons(1977)}]{Lyons:1977:semantica}
Lyons, J., 1977. Semantica. Teide, Barcelona.

\bibitem[{Nehmzov(2000)}]{Nehmzov:2000:Mobile_Robotics}
Nehmzov, U., 2000. Mobile Robotics: A Practical Introduction. Springer-Verlag.

\bibitem[{Nguyen et~al.(2007)Nguyen, G{\"a}chter, Martinelli, Tomatis, and
  Siegwart}]{nguyen2007comparison}
Nguyen, V., G{\"a}chter, S., Martinelli, A., Tomatis, N., Siegwart, R., 2007.
  {A comparison of line extraction algorithms using 2D range data for indoor
  mobile robotics}. Autonomous Robots 23~(2), 97--111.

\bibitem[{Nguyen et~al.(2005)Nguyen, Martinelli, Tomatis, and
  Siegwart}]{nguyen2005comparison}
Nguyen, V., Martinelli, A., Tomatis, N., Siegwart, R., 2005. {A comparison of
  line extraction algorithms using 2D laser rangefinder for indoor mobile
  robotics}. In: 2005 IEEE/RSJ International Conference on Intelligent Robots
  and Systems, 2005.(IROS 2005). pp. 1929--1934.

\bibitem[{Noykov and Roumenin(2007)}]{noykov2007calibration}
Noykov, S., Roumenin, C., 2007. {Calibration and interface of a polaroid
  ultrasonic sensor for mobile robots}. Sensors \& Actuators: A. Physical
  135~(1), 169--178.

\bibitem[{Oriolo et~al.(1997)Oriolo, Ulivi, and
  Vendittelli}]{OrioloUliviVendittelli:1997:Fuzzy_maps:_A_new_tool_for_mobile_%
robot_perception_and_planning}
Oriolo, G., Ulivi, G., Vendittelli, M., 1997. Fuzzy maps: A new tool for mobile
  robot perception and planning. Journal of Robotic Systems 14~(3), 179--197.

\bibitem[{Pradera et~al.(2007)Pradera, Trillas, Guadarrama, and
  Renedo}]{PraTriGuaRen:2007:On_fuzzy_set_theories}
Pradera, A., Trillas, E., Guadarrama, S., Renedo, E., 2007. On fuzzy set
  theories. In: Wang, P., Ruan, D., Kerre, E. (Eds.), Fuzzy Logic. A spectrum
  of Theoretical and Practical Issues. Vol. 215 of Studies in Fuzziness and
  Soft Computing. Springer, pp. 15--47.

\bibitem[{Ribo and
  Pinz(2001)}]{RiboPinz:2001:A_comparison_of_three_uncertainty_calculi_for_bui%
lding_sonar-based_occupancy_grids}
Ribo, M., Pinz, A., 2001. {A comparison of three uncertainty calculi for
  building sonar-based occupancy grids}. Robotics and Autonomous Systems
  35~(3-4), 201--209.

\bibitem[{Ruspini(1990)}]{Ruspini:1990:Fuzzy_Logic_in_the_Flakey}
Ruspini, E., 1990. {Fuzzy logic in the flakey robot}. In: Procs. of the Int.
  Conf on Fuzzy Logic and Neural Networks. Iizuka, Japan, pp. 767--770.

\bibitem[{Saffiotti(1997)}]{Saffiotti:1997:The_uses_of_fuzzy_logic_in_autonomo%
us_robot_navigation}
Saffiotti, A., 1997. {The uses of fuzzy logic in autonomous robot navigation}.
  Soft Computing-A Fusion of Foundations, Methodologies and Applications 1~(4),
  180--197.

\bibitem[{Se et~al.(2005)Se, Lowe, and Little}]{se2005vision}
Se, S., Lowe, D., Little, J., 2005. {Vision-based global localization and
  mapping for mobile robots}. IEEE Transactions on Robotics 21~(3), 364--375.

\bibitem[{Thrun(1998)}]{Thrun:1998:Learning_metric-topological_maps}
Thrun, S., 1998. {Learning metric-topological maps for indoor mobile robot
  navigation}. Artificial Intelligence 99~(1), 21--71.

\bibitem[{Thrun(2002)}]{Thrun:2002:Robotic_Mapping_A_Survey}
Thrun, S., 2002. {Robotic Mapping: A Survey}. In: Lakemeyer, G., Nebel, B.
  (Eds.), Exploring Artificial Intelligence in the New Millennium. Morgan
  Kaufmann, pp. 1--36.

\bibitem[{Thrun et~al.(2005)Thrun, Burgard, and
  Fox}]{Thrun:2005:Probabilistic_Robotics}
Thrun, S., Burgard, W., Fox, D., 2005. {Probabilistic Robotics (Intelligent
  Robotics and Autonomous Agents)}. MIT press, Cambridge, Massachusetts, USA.

\bibitem[{Trillas et~al.(1999)Trillas, Alsina, and
  Jacas}]{TriAlsJacas:1999:On_contradiction_in_fuzzy_logic}
Trillas, E., Alsina, C., Jacas, J., 1999. On contradiction in fuzzy logic. Soft
  Computing 3, 197--199.

\bibitem[{Trillas and
  Cubillo(2000)}]{TriCub:2000:On_a_Type_of_Antonymy_in_F([ab])}
Trillas, E., Cubillo, S., 2000. On a type of antonymy in {F([a,b])}. In:
  Proceedings IPMU 2000. Vol. III. Madrid, pp. 1728--1734.

\bibitem[{Trillas et~al.(2007)Trillas, Moraga, Guadarrama, Cubillo, and
  Casti{\~n}eira}]{TriMorGuaCubCas:2007:Computing_with_Antonyms}
Trillas, E., Moraga, C., Guadarrama, S., Cubillo, S., Casti{\~n}eira, E., 2007.
  {Computing with Antonyms}. In: Forging New Frontiers: Fuzzy Pioneers I. Vol.
  217 of Studies in fuzziness and soft computing. Springer, pp. 133--153.

\bibitem[{Trillas et~al.(2006)Trillas, Renedo, and
  Guadarrama}]{TriRenGua:2006:Fuzzy_Sets_vs_Language}
Trillas, E., Renedo, E., Guadarrama, S., 2006. {Fuzzy Sets vs Language}. In:
  Reusch, B. (Ed.), Computational Intelligence: Theory and Practice. Vol. 164
  of Studies in fuzziness and soft computing. Springer, pp. 353--366.

\bibitem[{Trivi{\~n}o(2000)}]{tesisgracian}
Trivi{\~n}o, G., 2000. Un modelo de arquitectura cognitiva aplicaci{\'o}n en
  rob{\'o}tica m{\'o}vil. Ph.D. thesis, Universidad Polit{\'e}cnica de Madrid,
  Facultad de Inform\'{a}tica, Departamento de Tecnolog{\'\i}a Fotonica,
  electronic version in http://oa.upm.es/641/.

\bibitem[{Yan et~al.(2006)Yan, Lee, Lee, and Tian}]{yan2006building}
Yan, L., Lee, C., Lee, S., Tian, Y., 2006. {Building Map Using Peak Amplitude
  of Sonar Echoes}. In: SICE-ICASE, 2006. International Joint Conference. pp.
  1692--1696.

\bibitem[{Yguel et~al.(2007)Yguel, Keat, Braillon, Laugier, and
  Aycard}]{Yguel-RSS-07}
Yguel, M., Keat, C. T.~M., Braillon, C., Laugier, C., Aycard, O., 2007. Dense
  mapping for range sensors: Efficient algorithms and sparse representations.
  In: Proceedings of Robotics: Science and Systems. Atlanta, GA, USA, pp.
  129--137.

\bibitem[{Zadeh(1975)}]{Zadeh:1975:The_concept_of_Linguistic_Variable}
Zadeh, L.~A., 1975. The concept of linguistic variable and its application to
  aproximate reasoning, parts i,ii,iii. Information Sciences 8,8,9,
  199--249,301--357,43--80.

\bibitem[{Zadeh(2001{\natexlab{a}})}]{Zadeh:2001:A_new_direction_in_AI}
Zadeh, L.~A., 2001{\natexlab{a}}. {A New Direction in AI: Toward a
  Computational Theory of Perceptions}. AI Magazine 22~(1), 73.

\bibitem[{Zadeh(2001{\natexlab{b}})}]{Zadeh:2001:From_computing_with_numbers_t%
o_computing_with_words}
Zadeh, L.~A., 2001{\natexlab{b}}. From computing with numbers to computing with
  words.from manipulation of measurements to manipulation of perceptions. In:
  Wang, P. (Ed.), Computing with words. John Wiley \& Sons, pp. 35--68.

\bibitem[{Zadeh(2005)}]{zadeh:2005:Toward_a_generalized_theory_of_uncertainty_%
GTU_an_outline}
Zadeh, L.~A., 2005. {Toward a generalized theory of uncertainty (GTU)--an
  outline}. Information Sciences 172~(1-2), 1--40.

\bibitem[{Zadeh(2006)}]{Zadeh:2006:A_new_frontier_in_computation}
Zadeh, L.~A., 2006. {A New Frontier in Computation? Computation with
  Information Described in Natural Language}. In: Proceedings Symposium on
  Fuzzy Systems in Computer Science FSCS. pp. 1--2.

\bibitem[{Zadeh(2008{\natexlab{a}})}]{Zadeh:2008:Is_there_a_need_for_fuzzy_log%
ic}
Zadeh, L.~A., 2008{\natexlab{a}}. {Is there a need for fuzzy logic?}
  Information Sciences 178~(13), 2751--2779.

\bibitem[{Zadeh(2008{\natexlab{b}})}]{Zadeh:2008:Toward_human_level_machine_in%
telligence}
Zadeh, L.~A., 2008{\natexlab{b}}. {Toward Human Level Machine Intelligence-Is
  It Achievable? The Need for a Paradigm Shift}. IEEE Computational
  Intelligence Magazine 3~(3), 11--22.

\bibitem[{Zhang et~al.(2007)Zhang, Rad, Wong, Huang, Ip, and
  Chow}]{zhang2007comparative}
Zhang, X., Rad, A., Wong, Y., Huang, G., Ip, Y., Chow, K., 2007. {A Comparative
  Study of Three Mapping Methodologies}. Journal of Intelligent and Robotic
  Systems 49~(4), 385--395.

\end{thebibliography}
\bibliographystyle{elsarticle-harv}

\newpage
\appendix
\section{Comparison of maps}

Several maps have already appeared in the text of this paper
to facilitate its explanations. For ease of comparison, we have
repeated some of them here.

\begin{figure}[!htb]
\begin{minipage}{.47\linewidth}
\centering\includegraphics[width=\linewidth]{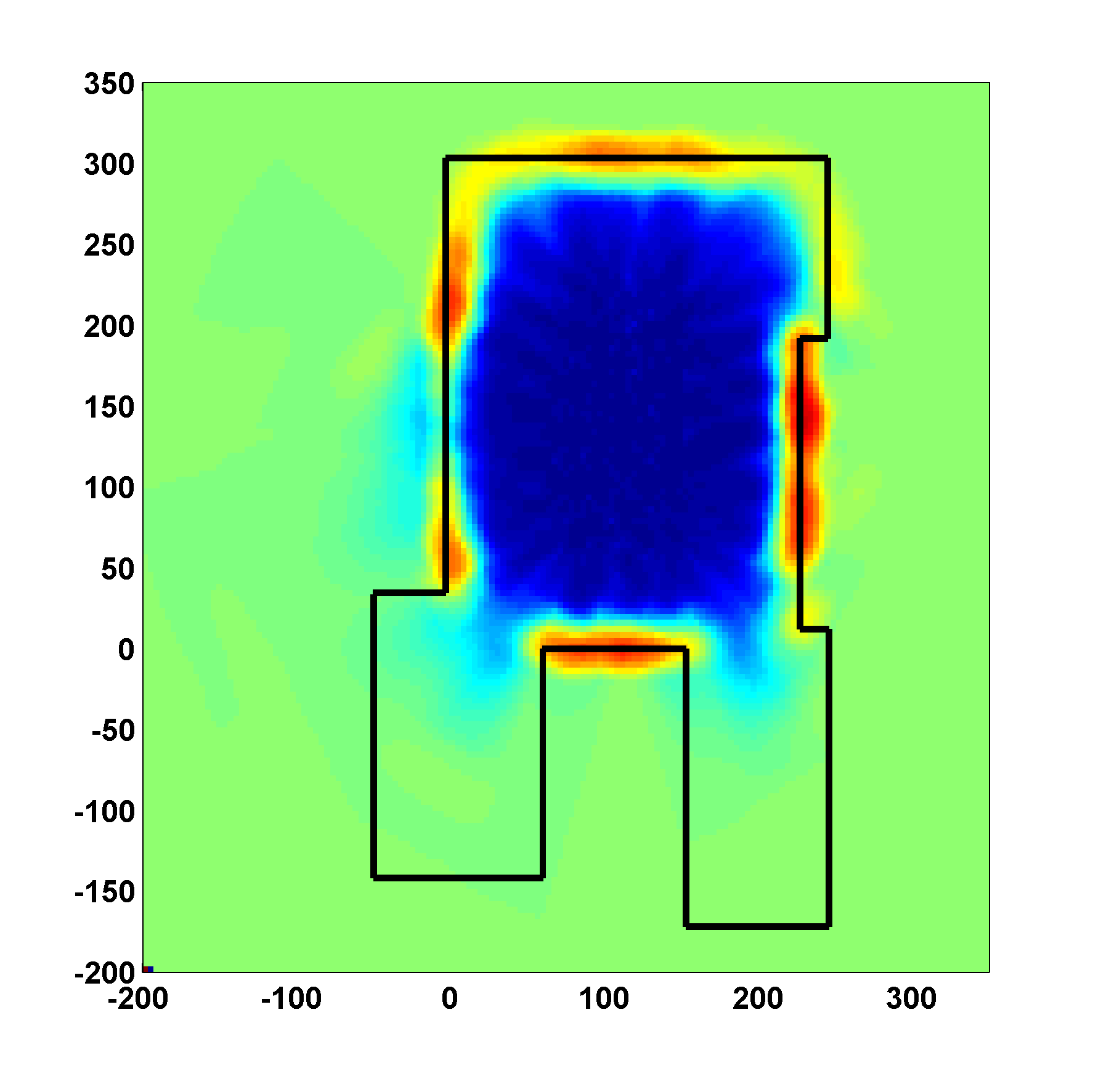}
\caption{Map of the office by the probabilistic method}\label{fig:prob-office}
\end{minipage}
\begin{minipage}{.47\linewidth}
\centering\includegraphics[width=\linewidth]{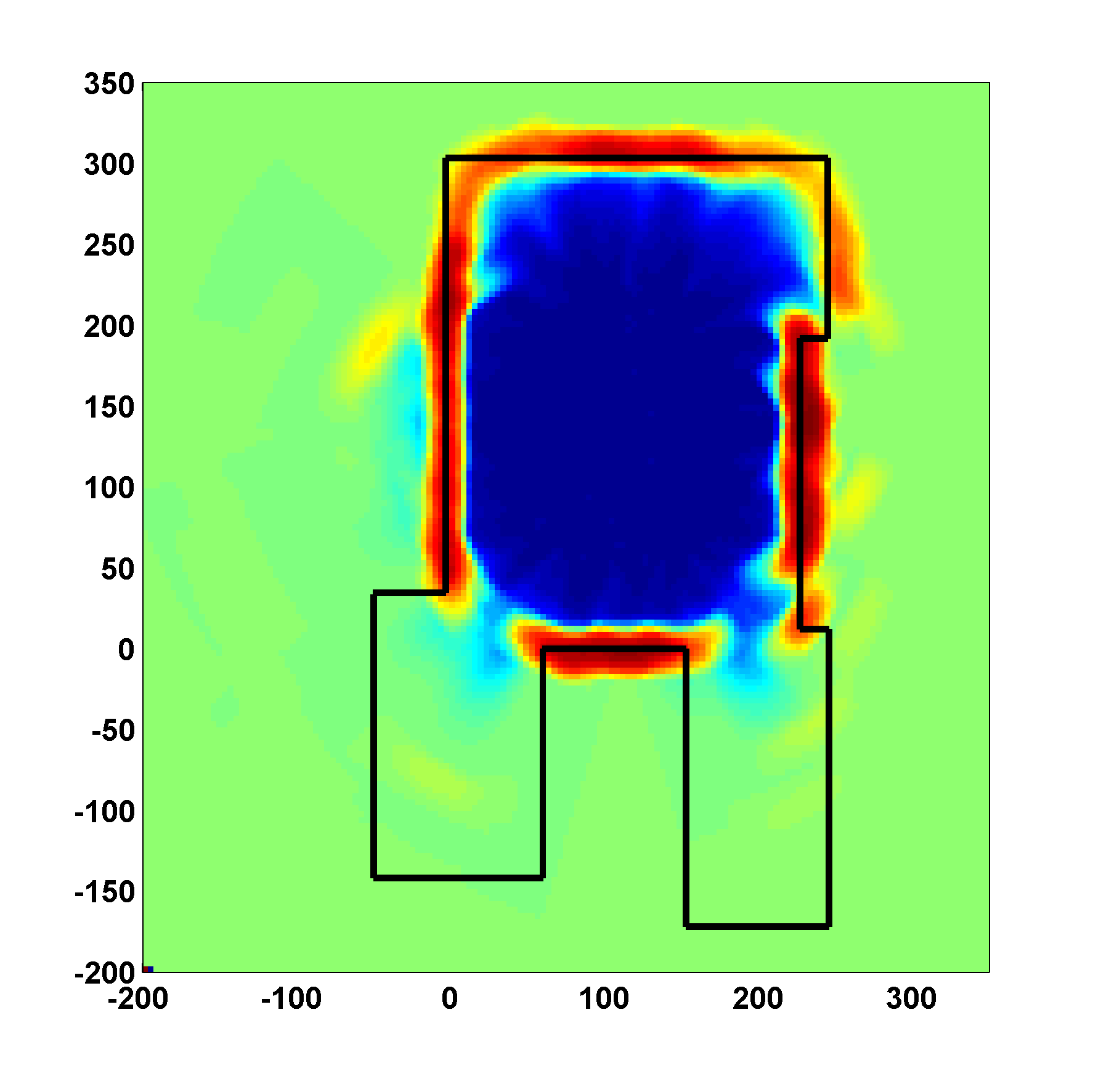}
\caption{Map of the office by the fuzzy method}\label{fig:poss-office}
\end{minipage}
\centering\includegraphics[width=.5\linewidth]{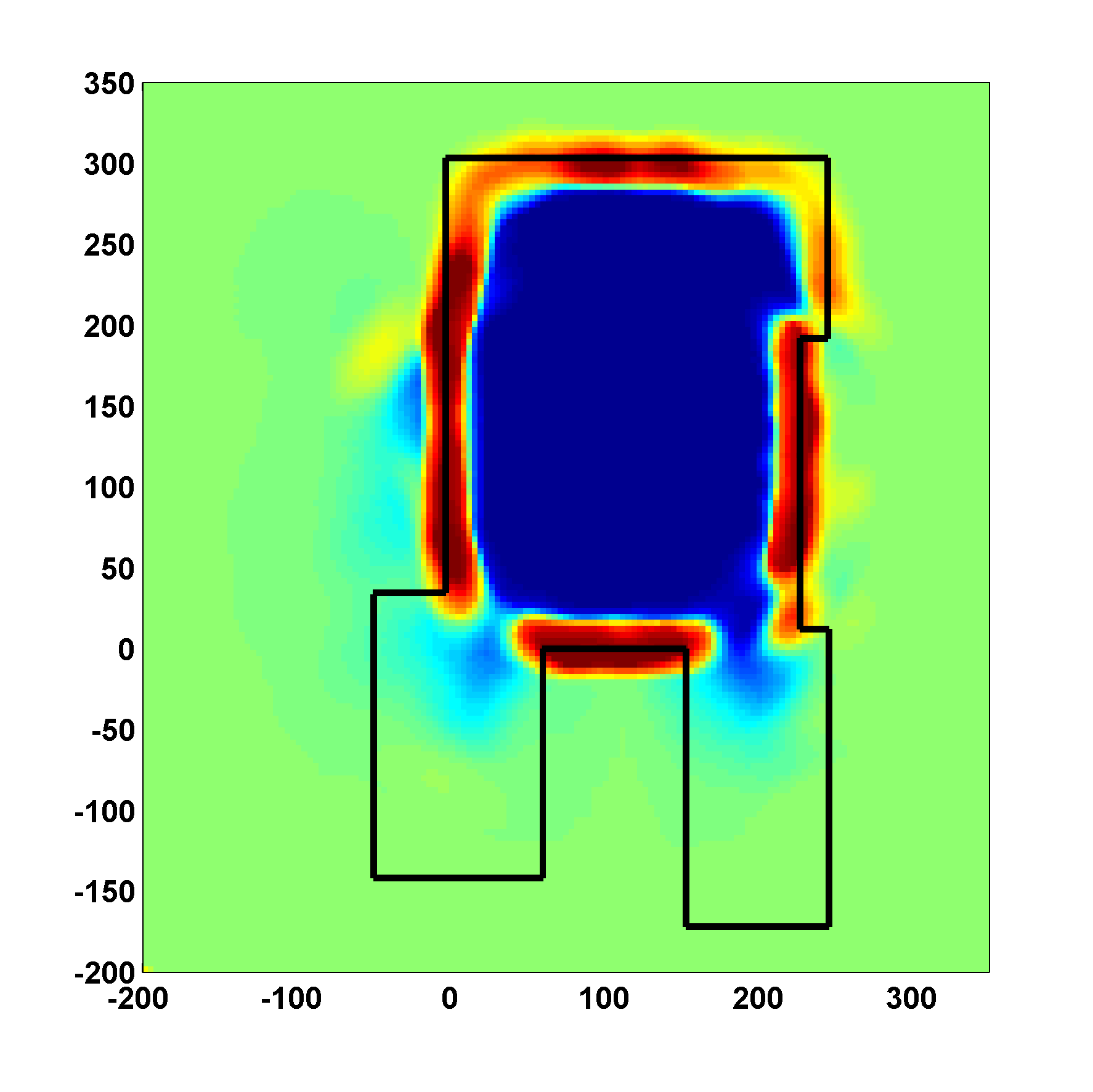}
\caption{Map of the office by the antonyms-Based
method}\label{fig:fuzz-office}
\end{figure}

\begin{figure}[b]
\begin{minipage}{.47\linewidth}
\centering\includegraphics[width=\linewidth]{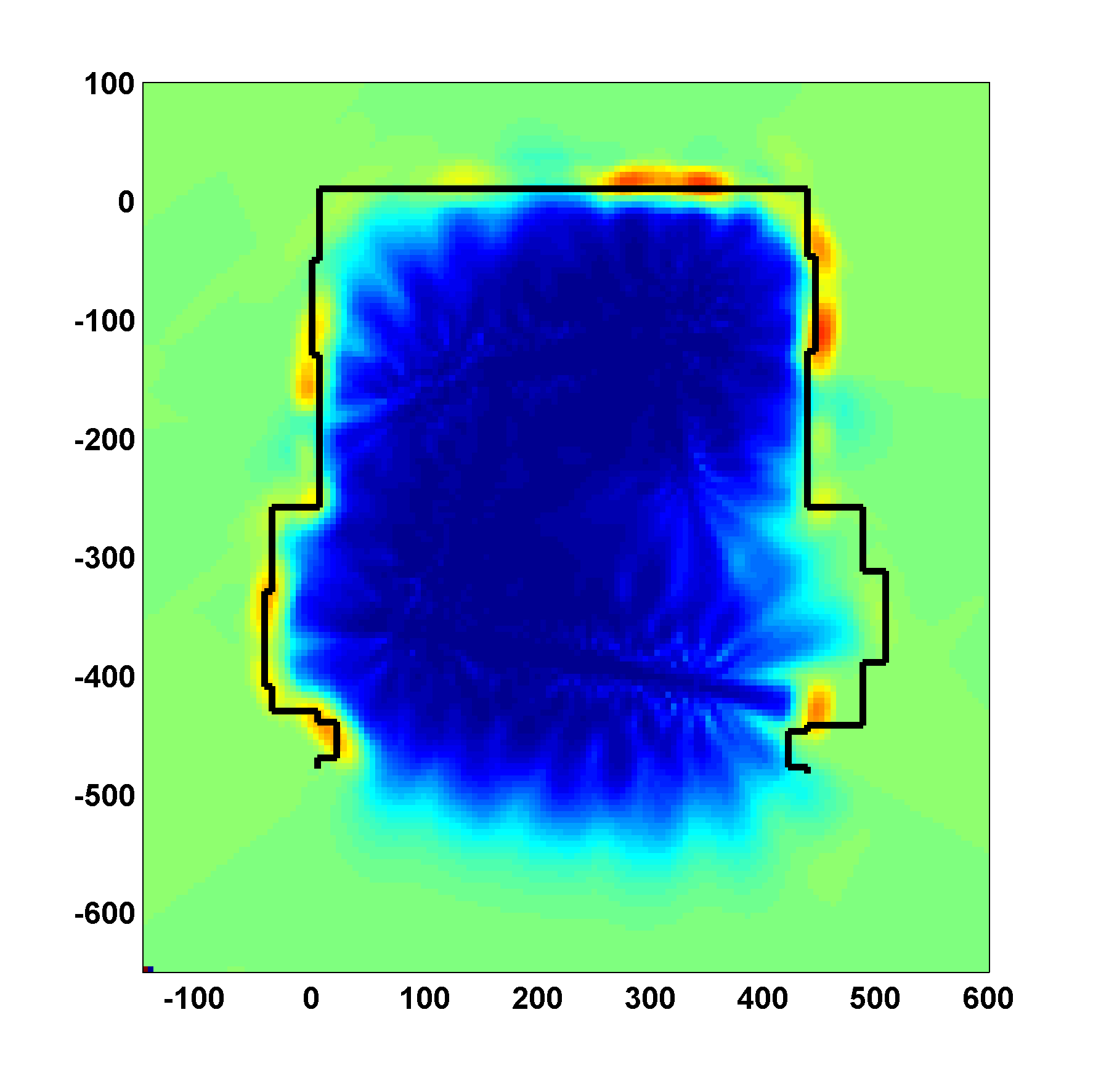} \caption{Map
of the hall by the probabilistic method}\label{fig:prob-hall}
\end{minipage}
\begin{minipage}{.47\linewidth}
\centering\includegraphics[width=\linewidth]{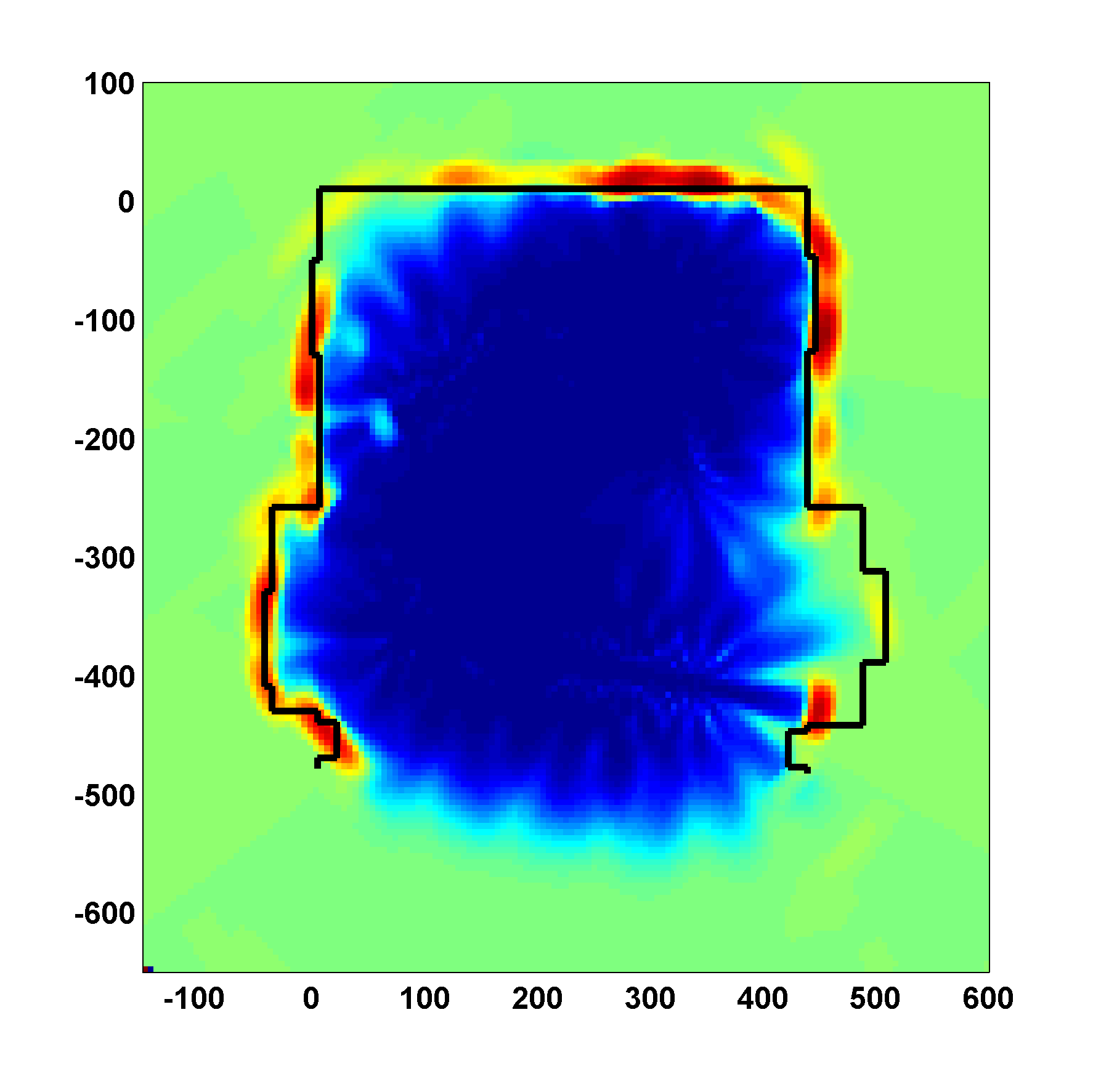} \caption{Map
of the hall by the fuzzy method}\label{fig:poss-hall}
\end{minipage}
\centering\includegraphics[width=.5\linewidth]{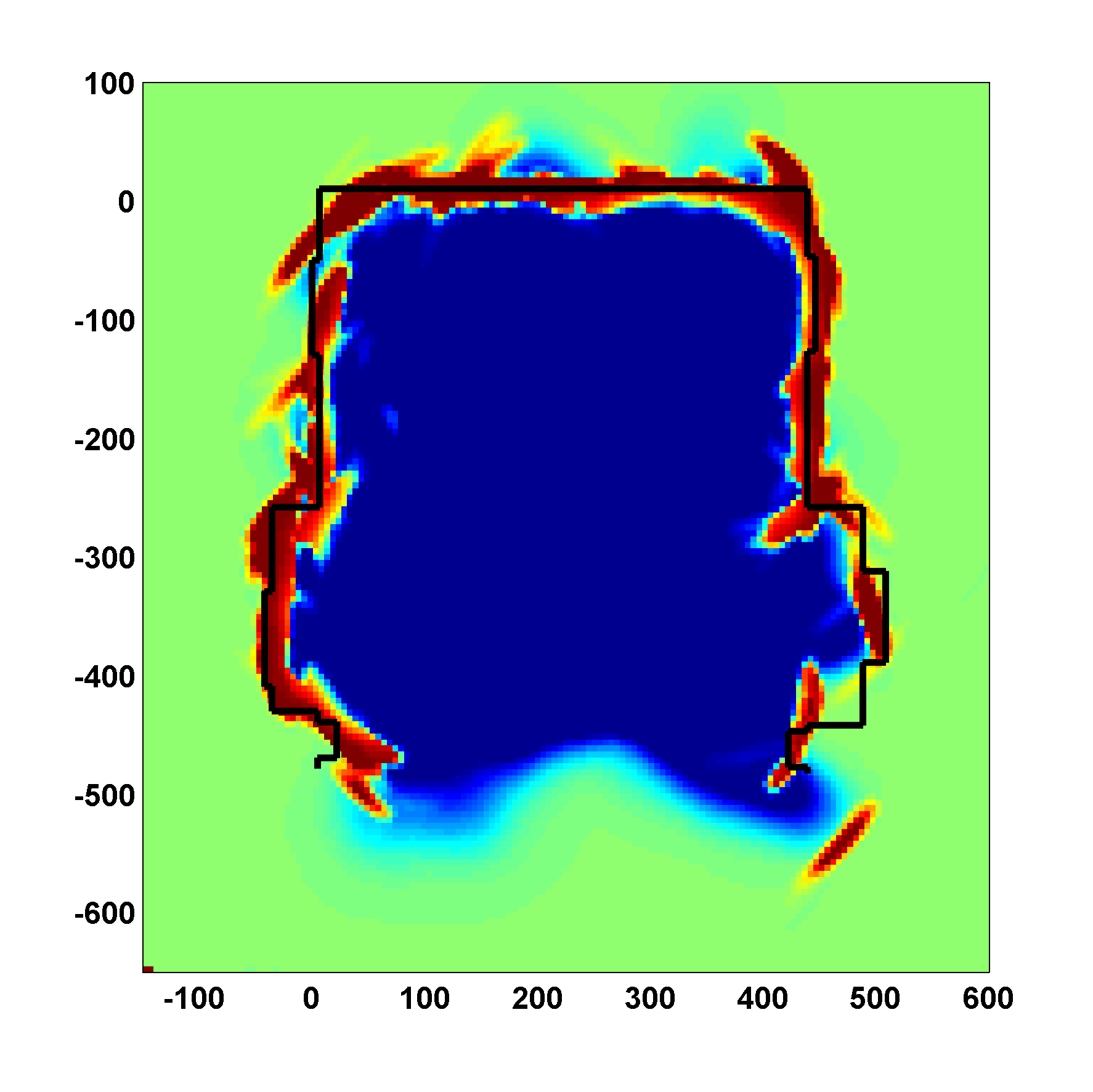}
\caption{Map of the hall by the antonyms-based method}\label{fig:fuzz-hall}
\end{figure}

\begin{figure}[b]
\begin{minipage}{.47\linewidth}
\centering\includegraphics[width=\linewidth]{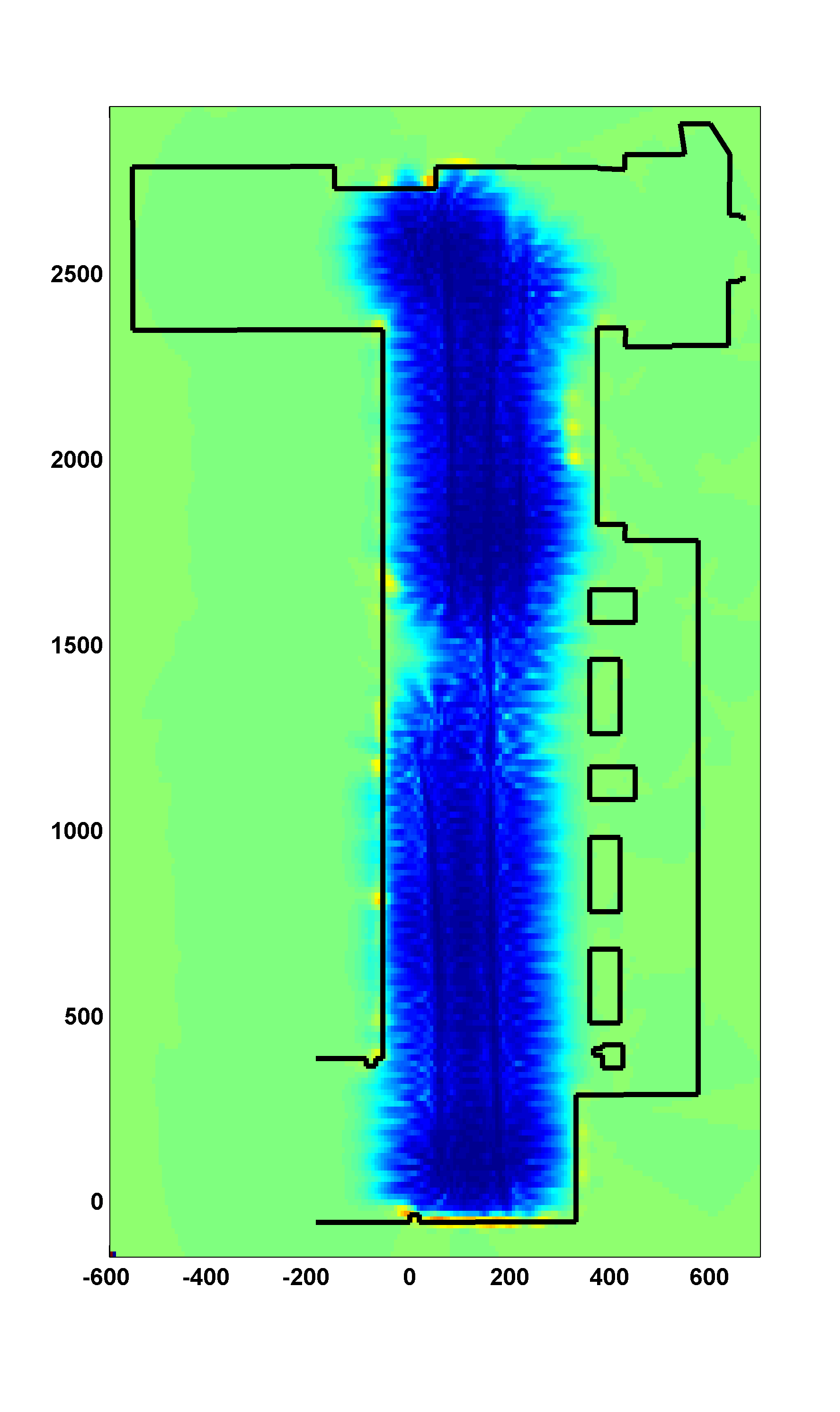}
\caption{Map of the corridor by the probabilistic
method}\label{fig:prob-corridor}
\end{minipage}
\begin{minipage}{.47\linewidth}
\centering\includegraphics[width=\linewidth]{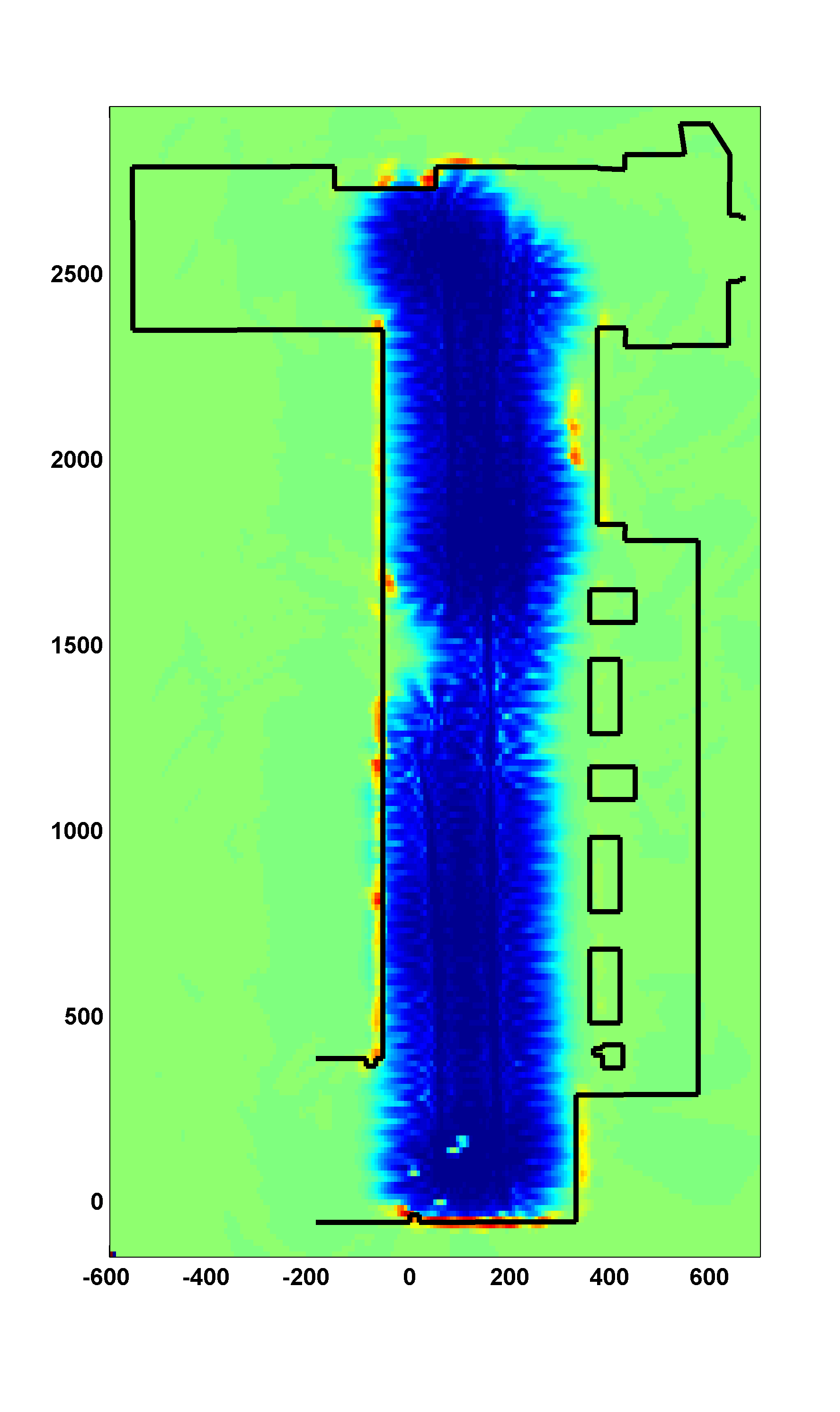}
\caption{Map of the corridor by the fuzzy method}\label{fig:poss-corridor}
\end{minipage}
\centering\includegraphics[width=.5\linewidth]{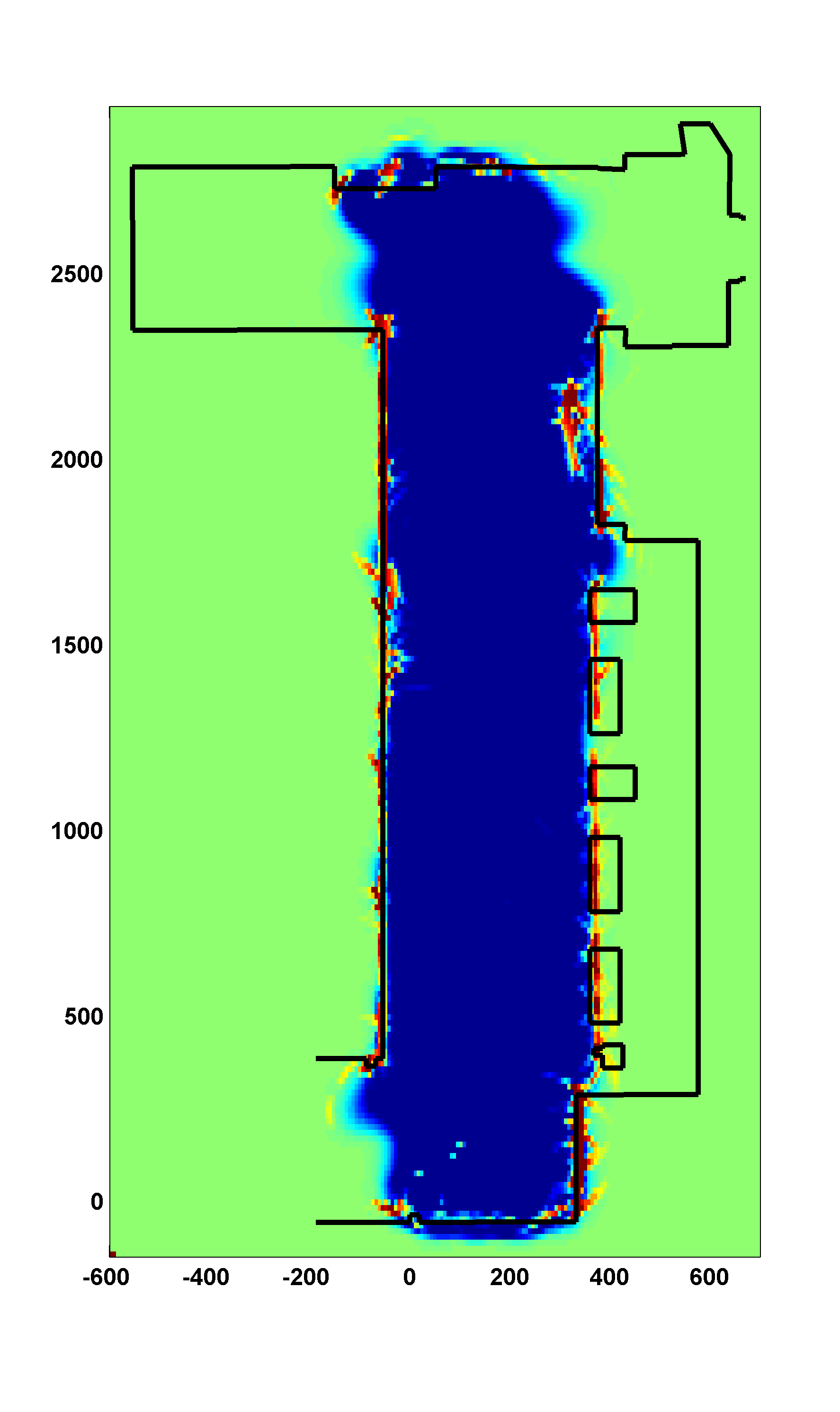}
\caption{Map of the corridor by the antonyms-based
method}\label{fig:fuzz-corridor}
\end{figure}

\end{document}